\documentclass[Afour,sageh,times]{sagej}

\usepackage{moreverb,url}

\usepackage[colorlinks,bookmarksopen,bookmarksnumbered,citecolor=blue,urlcolor=blue]{hyperref}

\usepackage{algorithm}
\usepackage{algpseudocode}
\usepackage[title]{appendix}
\usepackage{enumitem}

\newcommand\BibTeX{{\rmfamily B\kern-.05em \textsc{i\kern-.025em b}\kern-.08em
T\kern-.1667em\lower.7ex\hbox{E}\kern-.125emX}}

\def\volumeyear{2023}
\def\journalname{\mbox{The International Journal of} Robotics~Research}

\begin{document}

\runninghead{Lloyd and Lepora}

\title{Pose and shear-based tactile servoing}

\author{John Lloyd\affilnum{1,2} and Nathan F. Lepora\affilnum{1,2}}

\affiliation{\affilnum{1} School of Engineering Mathematics and Technology, University of Bristol, UK\\ \affilnum{2} Bristol Robotics Laboratory, University of Bristol, UK}

\corrauth{Nathan Lepora, University of Bristol, UK}

\email{n.lepora@bristol.ac.uk}

\begin{abstract}
Tactile servoing is an important technique because it enables robots to manipulate objects with precision and accuracy while adapting to changes in their environments in real-time. One approach for tactile servo control with high-resolution soft tactile sensors is to estimate the contact pose relative to an object surface using a convolutional neural network (CNN) for use as a feedback signal. In this paper, we investigate how the surface pose estimation model can be extended to include shear, and utilize these combined pose-and-shear models to develop a tactile robotic system that can be programmed for diverse non-prehensile manipulation tasks, such as object tracking, surface following, single-arm object pushing and dual-arm object pushing. In doing this, two technical challenges had to be overcome. Firstly, the use of tactile data that includes shear-induced slippage can lead to error-prone estimates unsuitable for accurate control, and so we {\color{black}modified the CNN into} a Gaussian-density neural network and used a discriminative Bayesian filter to improve the predictions with a state dynamics model that utilizes the robot kinematics. Secondly, to achieve smooth robot motion in 3D space while interacting with objects, we used $SE(3)$ velocity-based servo control, which required re-deriving the Bayesian filter update equations using Lie group theory, as many standard assumptions do not hold for state variables defined on non-Euclidean manifolds. In future, we believe that pose and shear-based tactile servoing will enable many object manipulation tasks and the fully-dexterous utilization of multi-fingered tactile robot hands. 
\end{abstract}

\keywords{Tactile sensing, deep learning, Bayesian filtering, feedback control, tactile servoing, robotic pushing, object manipulation\\
\vspace{-1.5em}}

\maketitle

\section{Introduction}
\label{sec:introduction}

\begin{figure*}[t]
	\centering
	\includegraphics[width=1\columnwidth,trim=85 40 85 50,clip]{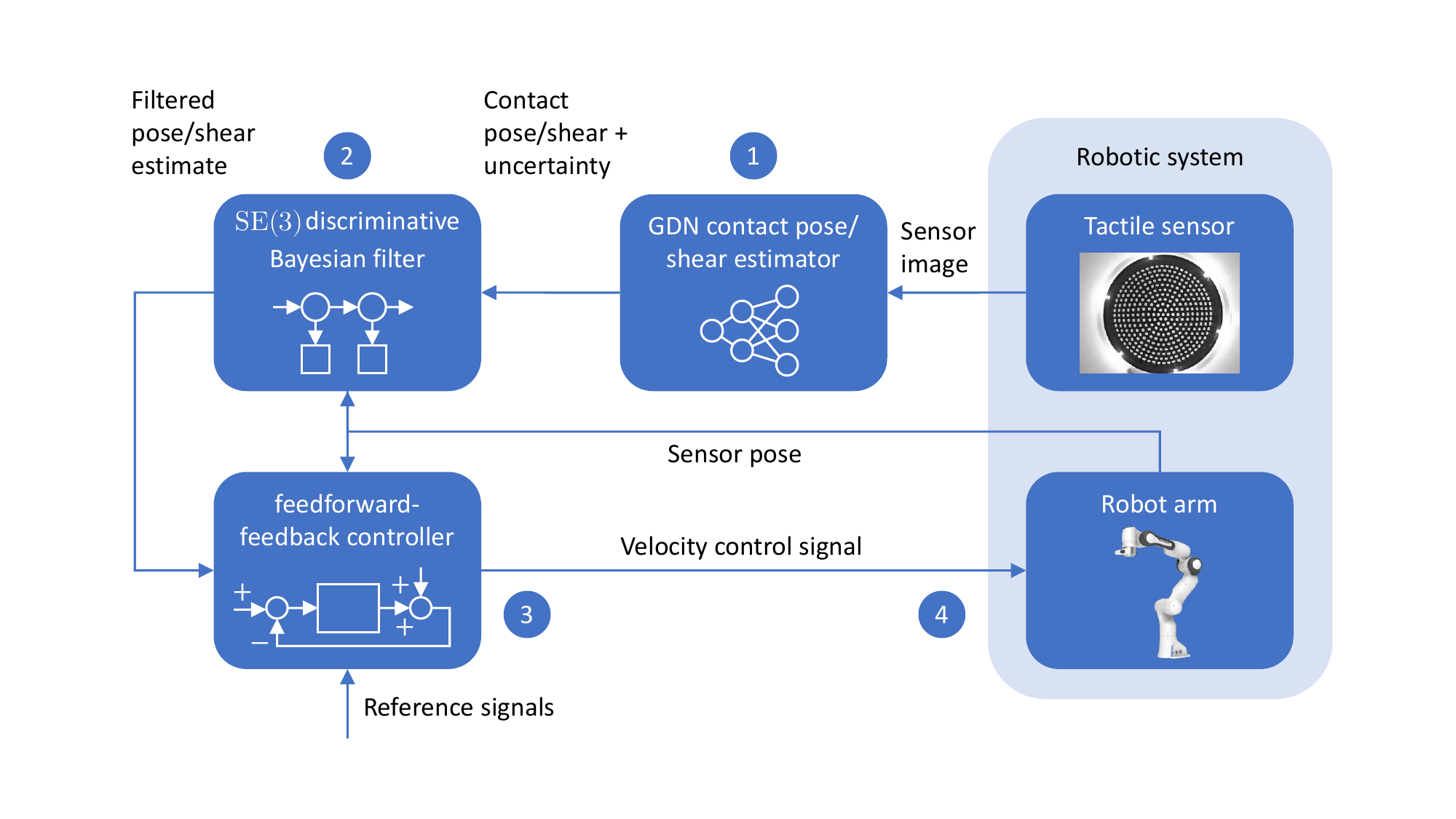}
	\includegraphics[width=1\columnwidth,trim=-30 100 230 0,clip]{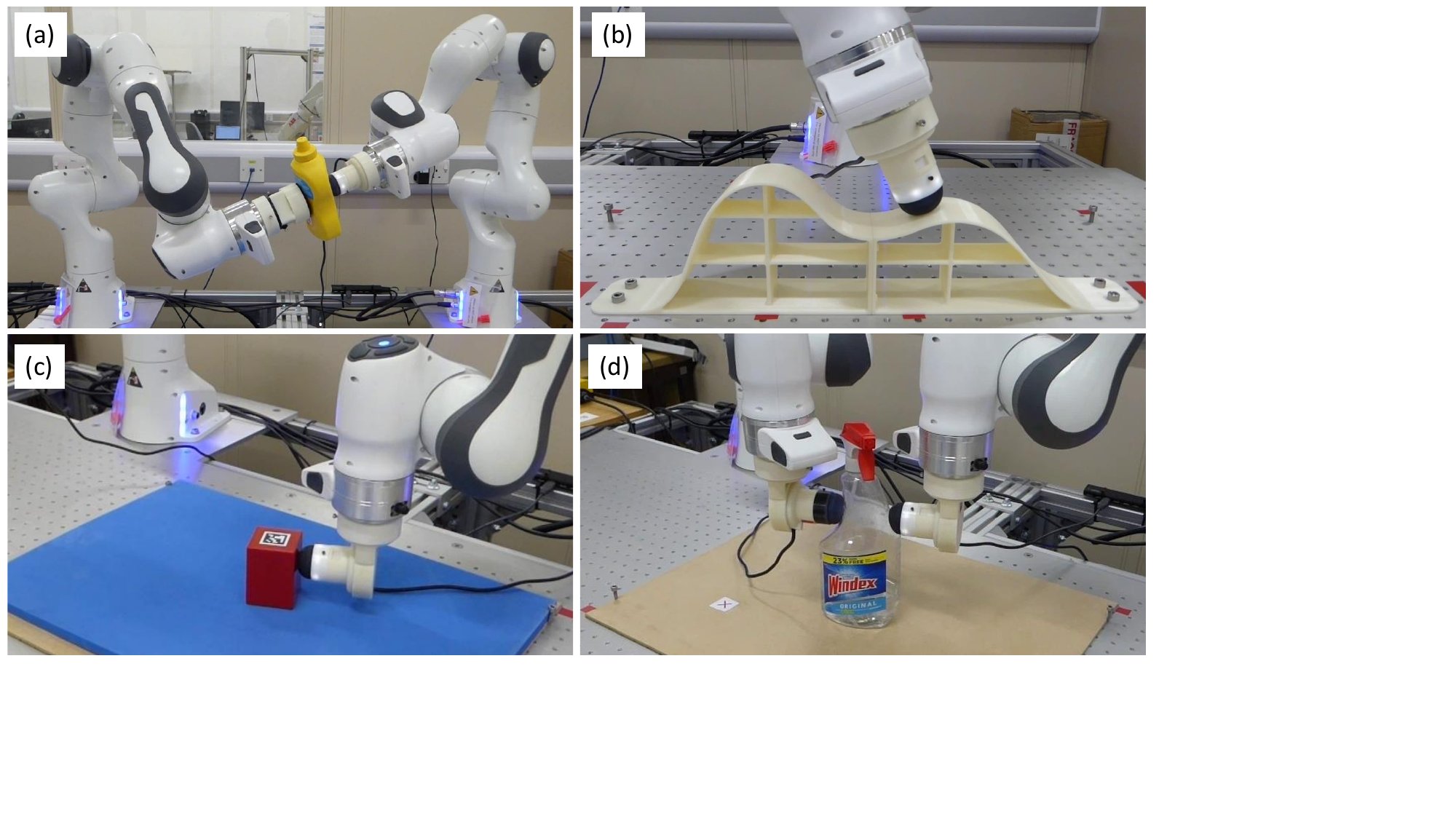}
	\caption{Pose and shear-based tactile servo control (left) applied to four tasks (right): (a) object tracking; (b) surface following; (c)~single-arm object pushing; (d) dual-arm object pushing. Here the servo control loop for each tactile robot has: (1) a Gaussian density network (GDN) model for predicting the contact pose and post-contact shear with uncertainty from a tactile image; (2) an $SE(3)$ discriminative Bayesian filter for reducing the error and uncertainty of pose/shear estimates; (3) a feedforward-feedback controller that outputs a velocity control signal to (4) a robot arm fitted with a vision-based tactile sensor as an end-effector. {\color{black}Examples of tactile data under different contact poses and post-contact shears are given later in Figure 11.}
	\label{fig:system_architecture}}
\end{figure*}

Tactile sensing is an essential component for enabling controlled physical interaction between a robot and its surroundings. A technique known as \emph{tactile servoing} (\cite{li2013control, lepora2021pose}) uses tactile feedback to adjust the position or velocity of a robot's end-effector, such as a gripper, hand or tool. Tactile servoing is an important technique in robotics because it enables robots to manipulate objects with precision and accuracy while adapting to changes in their environments in real-time. {\color{black}Tactile servo control is based on its more established counterpart: visual servo control~({\em e.g.} \cite{espiau1992new,hutchinson1996tutorial,chaumette2006visual}), which since the 1990s has led to many applications in vision-based robotics. However, tactile servoing cannot yet be widely deployed because it is far more difficult to perceive relevant state information from a soft tactile sensor interacting physically with its environment than from a non-contact sense such as vision. For example, the recent time history of the combined normal and shear motion will affect the deformation of a soft tactile sensor. This paper aims to close that gap by showing how tactile pose-and-shear state information can enable many robot~manipulation tasks.} 

One approach for tactile servo control with high-resolution soft tactile sensors is to estimate the contact pose relative to an object surface using a convolutional neural network (CNN) and use this as a feedback signal: the tactile sensor can then slide over unknown curved surfaces and push unknown objects to goal locations (\cite{lepora2020optimal, lepora2021pose, lloyd2021goal}). However, soft tactile sensors shear after contact, which our previous work treated as an undesirable ``nuisance variable" that interferes with the primary goal of contact pose estimation. The new viewpoint taken in this paper is that post-contact shear is a useful attribute that can aid servoing and manipulation tasks. For example, when tracking a moving object that rotates and translates relative to a contact region, it is essential that a robot can detect and respond to both shear and normal contact interactions with the tactile sensor. 

In this paper, we investigate how the surface pose estimation model can be extended to include shear effects, and we utilize these combined pose-and-shear models to develop a tactile robotic system that can be programmed for diverse non-prehensile manipulation tasks, such as object tracking, surface following, single-arm object pushing and dual-arm ({\em i.e.} stabilized) object pushing. While we have previously demonstrated tactile surface following and single-arm pushing, our new system is capable of performing these tasks with a continuous smooth motion instead of using discrete, position-controlled movements. Significant new capabilities such as tactile object tracking and dual-arm pushing are only possible because the tactile robotic system can respond to shearing motion.

Two technical challenges had to be overcome to realise a pose and shear-based tactile servoing system with a soft high-resolution tactile sensor. Firstly, slippage under shear can cause {\em tactile aliasing} (\cite{lloyd-RSS-21}) whereby very similar sensor images may become associated with significantly different poses and shears, which can lead to error-prone pose and shear estimation that is not suitable for accurate control. To address this, we {\color{black}modified the CNN into} a Gaussian-density neural network (GDN) that predicts both the values and the associated uncertainties of the pose and shear components, which we then feed through a Bayesian filter to accurately predict pose and shear. The second technical challenge relates to controlling smooth continuous robot motion in 3D space while interacting with an object or surface. This challenge necessitated the use of $SE(3)$ velocity-based servo control methods, which introduced numerous difficulties into the formalism. For example, the standard assumption underlying many Bayesian filters ({\em e.g.} Kalman filters) that the normalized product of two Gaussian probability distributions is itself Gaussian does not hold on non-Euclidean manifolds such as $SE(3)$. Hence, the techniques in this paper rely on technical derivations using Lie group theory {\color{black}within the context of robot manipulation control in $SE(3)$}, which we present in the Appendices. 

Our main novel contributions are summarised below and pictured in Figure~\ref{fig:system_architecture}:
\begin{enumerate}[leftmargin=0.5cm]
    \item A Gaussian density neural network model that predicts contact pose and post-contact shear with uncertainty from tactile images, and which represents the pose and shear as a single, unified vector that transforms under $SE(3)$.
    \item A discriminative Bayesian filter that reduces the error and uncertainty of the combined pose-and-shear predictions in $SE(3)$, which enables the use of accurate, noise-reduced estimates for tactile servo control.
    \item Feedforward-feedback control methods using velocity control that are driven by tactile pose-and-shear estimation for tactile servo control, supplemented with controllers for goal-based tasks such as object pushing.  
    \item The application and assessment of these techniques for smoothly and accurately controlling single- and dual-armed tactile robotic systems for object tracking, surface following, single-arm pushing and dual-arm pushing.
\end{enumerate}

This paper is organized as follows. Section~\ref{sec:background} gives an overview of tactile pose and shear estimation followed by a survey of tactile servoing and object pushing. In the {\em methodology}, Section~\ref{sec:nn_pose_shear_estimate} introduces our definition of surface pose and shear in $SE(3)$, then goes on to describe the data collection and modelling procedures for {\color{black}CNNs with a regression head, and a} GDN architecture. Section~\ref{sec:se3_bayes_filter} derives our algorithms for discriminative Bayesian filtering over $SE(3)$ random variables with uncertainty, referring to results derived in Appendices~\ref{sec:math_prelim}-\ref{sec:se3_probabilistic_fusion}. Section~\ref{sec:se3_feedforward_feedback_control} details our methods for feedforward-feedback control of pose and shear in $SE(3)$, considering a tactile servoing controller and a combined tactile servoing/pushing controller applied to single- and dual-arm robot systems. Section~\ref{sec:platform}  describes our tactile robot experimental platform and software, comprising a dual robot arm system with soft high-resolution tactile sensors and various test objects. In the {\em experimental results}, Section~\ref{sec:nn_pose_shear_information} begins with examining pose-and-shear information in tactile images. Section~\ref{sec:nn_pose_shear_estimate_exp} quantifies and compares the CNN {\color{black}regression} and GDN model predictions of pose and shear with uncertainty. Then Section~\ref{sec:se3_bayes_filter_exp} assesses the error and uncertainty reduction with $SE(3)$ Bayesian filtering. The tactile controller performance is then assessed on four tasks: (1) object pose tracking (Section~\ref{sec:obj_pose_tracking_exp}) for single and multiple pose components; (2) surface following (Section~\ref{sec:surface_follow_exp}) on a curved ramp and hemisphere; (3) single-arm object pushing (Section~\ref{sec:single_arm_object_pushing_exp}) of geometric objects; and (4) dual-arm object pushing (Section~\ref{sec:dual_arm_object_pushing_exp}) with the same geometric objects and taller versions, along with some tall household objects. Finally, Section~\ref{sec:discussion} discusses these results and their limitations, focusing on the tactile pose-and-shear estimation, then the tactile servo control task performance.

{\color{black}A video of the experiments is included as supplementary material and is publicly available on \href{https://www.youtube.com/watch?v=xVs4hd34ek0}{YouTube}. We have released the data and code for this paper on \href{https://github.com/dexterousrobot}{github.com/dexterousrobot} with a guide and summary on \href{https://lepora.com/pose-and-shear}{lepora.com/pose-and-shear}.}

\section{Background and related work}
\label{sec:background}

\subsection{Tactile pose-and-shear estimation}
\label{sec:tactile_pose_shear_estimation}

Contemporary methods for tactile pose estimation can be broadly categorised according to whether they estimate a \emph{local} contact pose or a \emph{global} object pose. Local contact pose estimation tends to be easier because it can always use tactile information provided in a single contact, although accuracy can be improved by using a sequence of observations or contacts. As such, it has been studied over a longer period of time than the second problem. The global object pose estimation problem is generally harder to solve because it involves fusing information from several different contact interactions, and using this information together with some form of object model to estimate the object pose (\cite{bimbo2015global, suresh2021tactile, villalonga2021tactile, bauza2022tac2pose, kelestemur2022tactile, caddeo2023collision}). As such, work on this type of pose estimation problem is more recent, and this has largely been driven by progress in deep learning models. Comprehensive reviews of pose estimation in the context of robotic tactile perception can be found in \cite{luo2017robotic} and \cite{li2020review}.

In early work on tactile pose estimation, Bicchi et al. proposed a theoretical model for estimating pose-and-shear information, and described a framework for designing tactile sensors that have this capability (\cite{bicchi1993contact}). More specifically, their theoretical model addressed the problem of how to determine the location of a contact, the force at the interface and the moment about the contact normals.

In the context of high-resolution vision-based tactile sensors, Yuan et al. showed that the GelSight sensor can be used to estimate the normal contact pose between the sensor and an object surface, but was limited in its contact angle range due to its rather flat sensor geometry (\cite{yuan2017gelsight}). Similarly, Lepora et al. showed that the TacTip soft biomimetic optical tactile sensor could be used to predict 2D contact poses (\cite{lepora2017exploratory,lepora2019pixels}) and, more recently, 3D contact poses (\cite{lepora2020optimal,lepora2021pose}).

The estimation of post-contact shear is less well-explored than pose estimation. Yuan et al. showed how a GelSight sensor can be used to measure post-contact shear by including printed markers on the sensing surface (\cite{yuan2015measurement}). Cramphorn et al. described a similar approach for the TacTip sensor based on the shear of marker-tipped pins (\cite{cramphorn2018voronoi}). More recent work has considered data-efficient methods of decoupling the confounding post-contact shear from the primary goal of contact pose estimation, either through Principal Component Analysis~(\cite{aquilina2019}) or the latent feature space of a CNN model (\cite{pmlr-v164-gupta22a}). The present study takes a different approach in seeking a predictive model of the components of pose-contact shear that can be used alongside a model of contact pose for tactile servo control.

\subsection{Tactile servoing and object pushing}
\label{sec:tactile_servoing_pushing}

Methods for robotic tactile servoing can be grouped according to whether they control attributes in the \emph{signal space} or \emph{feature space} of tactile sensor signals or features, or attributes in the \emph{task space} associated with the problem under consideration. For vision-based tactile sensors or tactile sensors that produce a taxel image, if control is performed in the sensor feature space it can be referred to as \emph{image-based tactile servoing} (IBTS). Conversely, if control is performed in the task space and the task involves tracking a reference pose with respect to a surface feature it can be referred to as \emph{pose-based tactile servoing} (PBTS) (\cite{lepora2021pose}). In principle, a hybrid approach could also be used, where some aspects of control are performed in the task space and some in the signal or feature space. The tactile servo control methods used in this paper can be viewed as pose-based tactile servoing methods (and more generally as task-space methods) because we combine the contact pose-and-shear motion into a single ``surface contact pose" and use it in a feedback loop to control the robot arm motion. However, because of the importance of shear in the control, we refer to it as \emph{pose and shear-based tactile servoing}.

Historically, Berger and Khoslar first used image-based tactile feedback on the location and orientation of edges together with a feedback controller to track straight and curved edges in 2D (\cite{berger1991using}). Chen et al. used a task-space tactile servoing approach, using an ``inverse tactile model" similar in concept to a pose-based tactile servoing model, to follow straight-line and curved edges in 2D (\cite{chen1995edge}). Zhang and Chen used an image-based tactile servoing approach and introduced the concept of a ``tactile Jacobian" to map image feature errors to task space errors (\cite{zhang2000control}). They used their system to track straight and curved edges in 2D and to follow cylindrical and spherical surfaces in 3D. Sikka et al. drew inspiration from image-based \emph{visual} servoing to develop a tactile analogy using the taxel images produced by a tactile sensor to control the robot arm movement. They applied their tactile servoing system to the task of rolling a cylindrical pin on a planar surface (\cite{sikka2005tactile}).

Later on, Li et al. advanced the tactile servoing approach of Zhang and Chen to demonstrate a wider selection of servoing tasks including 3D object tracking and surface following (\cite{li2013control}). Lepora et al. used a TacTip soft optical tactile sensor with a bio-inspired active touch perception method and a simple proportional controller to demonstrate contour following around several complex 2D edges and ridges (\cite{lepora2017exploratory}), following a related contour-following method with an iCub fingertip~(\cite{martinez2017active,martinez2013active}). Sutanto et al. used learning-from-demonstration to build a tactile servoing dynamics model and used it to demonstrate 3D contact-point tracking (\cite{sutanto2019learning}). Kappassov et al. developed a task-space tactile servoing system, similar to the earlier system developed by Li et al., and used it for 3D edge following and object co-manipulation (\cite{kappassov2020touch}). More recently, Lepora and Lloyd described a pose-based tactile-servoing approach that uses a deep learning model to map from the tactile image space to pose space, firstly in 2D (\cite{lepora2019pixels}) and then in 3D (\cite{lepora2021pose}). They used this approach to demonstrate robotic surface and edge following on complex 2D and 3D objects.

Most current approaches for robotic object pushing also fall into two main categories: \emph{analytical physics-based approaches}, which are used in conventional robot planning and control systems; and \emph{data-driven approaches} for learning forward or inverse models of pusher-object interactions, or for directly learning control policies (e.g., using reinforcement learning). We summarise work on these two approaches in the following paragraphs. A comprehensive survey on robotic object pushing can be found in \cite{stuber2020let}.

In the case of analytical, physics-based object pushing, Mason derived a simple rule known as the \emph{voting theorem} for determining the direction of rotation of a pushed object (\cite{mason1986mechanics}). Goyal et al. introduced the concept of a \emph{limit surface} to describe how the sliding motion of a pushed object depends on its frictional properties (\cite{goyal1989limit}). Lee and Cutkosky derived an ellipsoid approximation to the limit surface, aiming to reduce the computational overhead of applying it in real applications (\cite{lee1991fixture}). Lynch et al. used the ellipsoid approximation to obtain closed-form analytical solutions for sticking and sliding pushing interactions (\cite{lynch1992manipulation}). Howe and Cutkosky explored other, non-ellipsoidal geometric forms of limit surface and provided guidelines for selecting them (\cite{howe1996practical}). Lynch and Mason analysed the mechanics, controllability and planning of object pushing and developed a planner for finding stable pushing paths between obstacles (\cite{lynch1996stable}).

In the case of data-driven approaches, Kopicki et al.used a modular data-driven approach for predicting the motion of pushed objects (\cite{kopicki2011learning}). Bauza et al. developed models that describe how an object moves in response to being pushed in different ways and embedded these models in a model-predictive control (MPC) system (\cite{bauza2018data}). Zhou et al. developed a hybrid analytical/data-driven approach that approximated the limit surface for different objects using a parametrised model (\cite{zhou2018convex}). Other researchers have used deep learning to model the forward or inverse dynamics of pushed object motion (\cite{agrawal2016learning, byravan2017se3, li2018push}), or to learn end-to-end control policies for pushing (\cite{clavera2017policy, dengler2022learning}). In general, analytical approaches are more computationally efficient and transparent in their operation than data-driven approaches, but may not perform well if their underlying assumptions and approximations do not hold in practice ({\cite{yu2016more}}).

While most object pushing methods rely on computer vision systems to track the pose and other state information of the pushed object, a few (including ours) use tactile sensors to perform this function. Lynch et al. were the first to employ tactile sensing to manipulate a rectangular object and circular disk on a moving conveyor belt (\cite{lynch1992manipulation}). Jia and Erdmann used a theoretical analysis to show that the pose and motion of a planar object with known geometry can be determined using only the tactile contact information generated during pushing (\cite{jia1999pose}). More recently, Meier et al. used a tactile-based method for pushing an object using frictional contact with its upper surface (\cite{meier2016distinguishing}).

From a control perspective, the most similar approaches to our method for single-arm robotic pushing are the ones described by Hermans (\cite{hermans2013decoupling}) and Krivic (\cite{krivic2019pushing}). The similarities and differences are described in more detail in \cite{lloyd2021goal}, but the main difference from our method is that they both used computer vision techniques to track the state of the pushed object, rather than tactile sensing and proprioception.

{\color{black}\section{Computational methods}}
\label{sec:methods}

\subsection{Contact pose-and-shear prediction with uncertainty}
\label{sec:nn_pose_shear_estimate}

\subsubsection{Surface contact pose and shear}
\label{sec:surface_contact_poses}

In previous work on tactile pose estimation (\cite{lepora2020optimal, lepora2021pose, lloyd2021goal}), we assumed that a surface contact pose can be represented by a 6-component vector, $\left( 0,0,z,\alpha,\beta,0 \right)$, where the $z$-component denotes the contact depth and the $\left( \alpha,\beta \right)$-components denote the two orientation angles of the sensor with respect to the surface normal. The three remaining components were set to zero because we assumed that all surface contacts are invariant to $\left( x,y \right)$-translation and $\gamma$-rotation parallel to an idealized flat surface. 

In this paper, we instead train a model to estimate all six non-zero components of a surface contact pose, $\left(x,y,z,\alpha,\beta,\gamma\right)$. To do this, we identify the $\left(z,\alpha,\beta\right)$ components with surface contact pose, as previously, and the $\left(x,y,\gamma\right)$ components with post-contact shear. This combination of pose and shear into a single vector arises from the geometry of a flat planar surface, in that the pose components where the surface remains invariant $\left(x,y,\gamma\right)$ happen to be along the primary motions that cause shear.  

Thus, in our new definition of a \emph{surface contact pose and post-contact shear}, we make the following two {\color{black}simplifying modeling assumptions}:
\begin{enumerate}[leftmargin=0.45cm]
	\item All contacted surfaces can be locally approximated as~flat.
	\item All sensor-surface contacts that produce a sensor output can be approximately decomposed into an equivalent normal contact motion followed by a tangential post-contact shear motion.
\end{enumerate}

\begin{figure}
	\centering
	\includegraphics[width=\columnwidth,trim=220 100 220 90,clip]{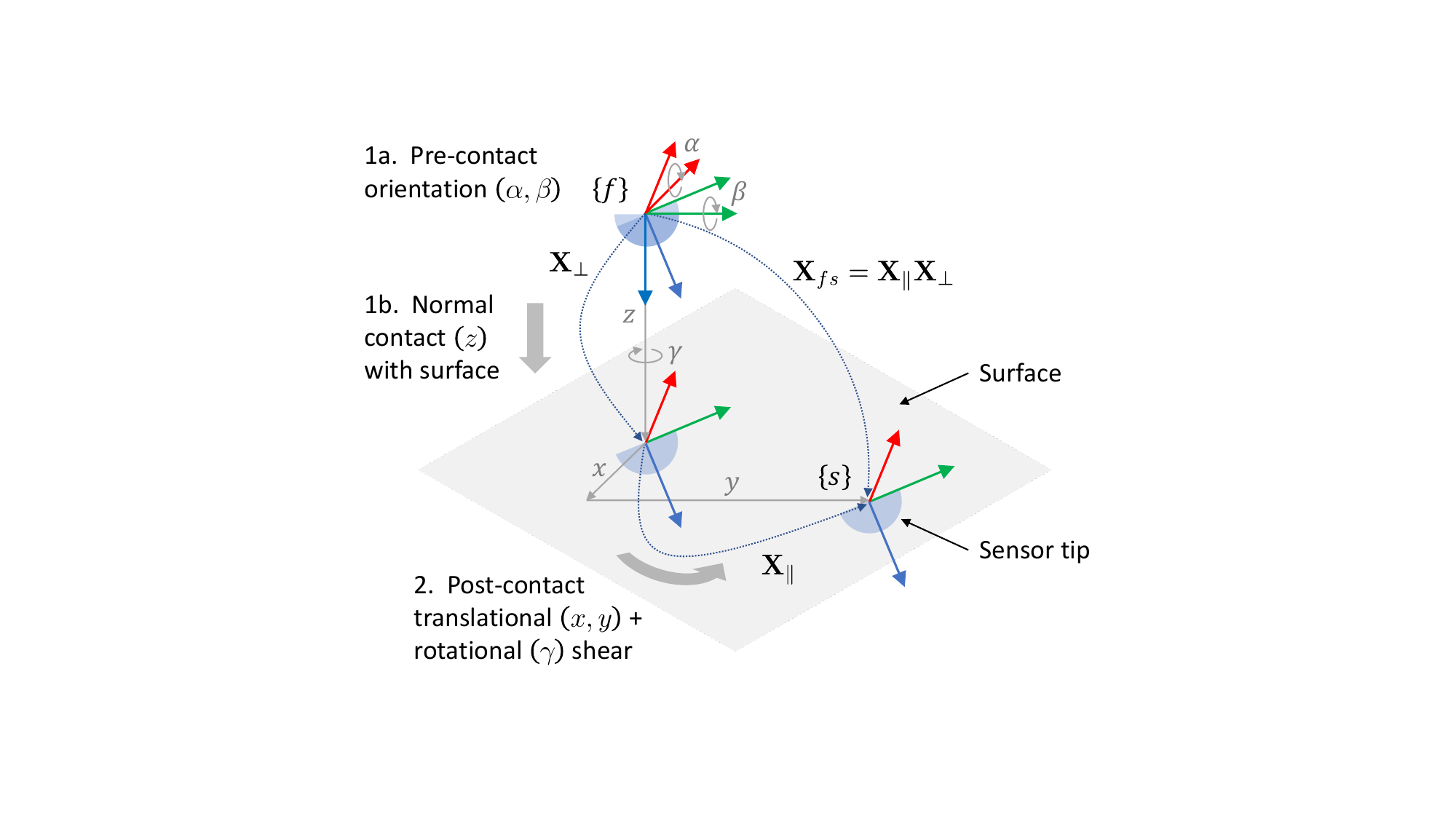}
	\caption{Definition and generation of surface contact poses using a two-step process of normal contact motion followed by translational and rotational shear. Step 1a: prior to normal contact motion, the sensor is rotated by Euler angles $(\alpha,\beta)$ with respect to the surface plane. Step 1b: the sensor is brought into normal contact with the surface through distance $z$. Step 2: the sensor is translated by $(x,y)$ parallel to the surface and rotated through angle $\gamma$ about the normal contact axis.
	\label{fig:surface_contact_pose_def}}
\end{figure}

\noindent {\color{black}In practise, we find that the tactile servo control methods apply equally well to curved objects and in situations when the sensor output depends on the time history of the combined normal and shear motion. Thus, with these two simplifying} assumptions in mind, we now define the surface contact poses we use to train our pose-and-shear estimation models (Figure~\ref{fig:surface_contact_pose_def}) and describe the process we use to sample and generate the data.

We start by attaching a sensor coordinate frame $\{s\}$ to the centre of the hemispherical sensor tip so that the $z$-axis is directed outwards from the tip of the sensor, along its radial axis. We also attach a surface feature frame $\{f\}$ to the surface so that its $z$-axis is normal to and directed inwards towards the surface. The $\{f\}$-frame is also located so that it is aligned with the sensor frame $\{s\}$ when the sensor is in its initial position, just out of normal contact with the surface.

As discussed above, the surface contact motion is assumed equivalent to one that is carried out in two stages: a normal contact motion followed by a post-contact, tangential shear motion. The normal contact motion, represented by an $\{f\}$-frame $SE(3)$ transform $\boldsymbol{\mathrm{X}}_{\perp}$, rotates the sensor by Euler angles $(\alpha,\beta)$ with respect to the surface (assuming an extrinsic-$xyz$ Euler convention) and then brings it into normal contact with the surface through distance $z$. The tangential shear motion is represented by another $\{f\}$-frame $SE(3)$ transform $\boldsymbol{\mathrm{X}}_{\parallel}$, which translates the sensor by a displacement $(x,y)$ parallel to the surface, while simultaneously rotating it about the normal contact axis through an angle $\gamma$. Composing these transformations into a single $SE(3)$ transform: $\boldsymbol{\mathrm{X}}_{fs} = \boldsymbol{\mathrm{X}}_{\parallel} \boldsymbol{\mathrm{X}}_{\perp}$, we take components in the Euler representation $(x,y,z,\alpha,\beta,\gamma)$, to represent the surface contact pose of the sensor in the surface feature frame $\{f\}$. Combining the contact pose and post-contact shear information in this way simplifies the subsequent filtering and control stages because it avoids the need for two separate filters and two pose-based controllers.

\subsubsection{Training Data Collection.}
\label{sec:data_collection}

We collect data for training the pose-and-shear prediction models by using a robot arm to move the sensor into different surface contact poses with a flat surface, then apply a shear motion before recording the tactile sensor image. Each data sample consists of a tactile image together with the corresponding surface pose and shear in extrinsic-$xyz$ Euler format.  

The tactile images associated with surface contact poses and shears $(x,y,z,\alpha,\beta,\gamma)$ are sampled according to a two-step procedure that mirrors the definition of the surface contact pose and shears given in the previous section (Algorithm~\ref{alg:data_collection}). Ranges for data collection were as follows:
\begin{enumerate}[leftmargin=0.5cm]
	\item $x$ and $y$ are sampled so that the translational shear displacements are distributed uniformly over a disk of radius $r_{\mathrm{max}}=5$\,mm centred on the initial point of normal contact with the surface.
	\item $z$ is sampled uniformly in the range $[0.5, 6]$\,mm, {\color{black}chosen to provide sufficient variation but not damage the sensor.}
	\item $\alpha$ and $\beta$ are sampled so that contacts with the sensor are distributed uniformly over a spherical cap of the sensor, which is subtended by angle $\phi_{\mathrm{max}}=25^\circ$ with respect to its central axis.
	\item $\gamma$ is sampled uniformly at random over a $[-5^\circ, 5^\circ]$ range, {\color{black}chosen to provide sufficient variation but limit slippage.}
\end{enumerate}

\noindent The sampling of $x$ and $y$ over a disk, is specified as follows:
\begin{equation}
	\label{eqn:disk_random_sampling}
	\begin{gathered}
		x = r \cos \theta, \ \; y = r \sin \theta, \ \; r = r_{\mathrm{max}}\sqrt{r^{\prime}},\\
		r^{\prime} \sim \mathcal{U}(0, 1), \;\; \theta \sim \mathcal{U}(0, 360^\circ),
	\end{gathered}
\end{equation}
where $\mathcal{U}(a, b)$ denotes a continuous uniform probability density function (PDF) over the interval, $\left[ a, b \right]$. The sampling of $\alpha$ and $\beta$ over a spherical cap is based on the method outlined by \cite{simon2015generating}:
\begin{equation}
	\label{eqn:sphere_random_sampling}
	\begin{gathered}
		\alpha = -\arcsin q, \;\; \beta = -\mathrm{atan2}(p, r),\\
		p = \sin \phi \cos \theta, \;\; q = \sin \phi \sin \theta, \;\; r = \cos \phi,\\
		\phi = \arccos (1 - (1 - \cos \phi_{\mathrm{max}}) \phi^{\prime}),\\
		\phi^{\prime} \sim \mathcal{U}(0, 1), \;\; \theta \sim \mathcal{U}(0, 360^\circ).
	\end{gathered}
\end{equation}

\begin{algorithm}[t]
	\caption{Training data collection}
	\label{alg:data_collection}
	\begin{algorithmic}
	\Require Surface contact poses $\boldsymbol{\mathrm{X}}_{i}$, $i=1 \ldots N$
	\Ensure Sensor tactile images $\boldsymbol{\mathrm{I}}_{i}$ for poses $X_{i}$, $i=1 \ldots N$
	\For{each pose $(x,y,z,\alpha,\beta,\gamma) = \boldsymbol{\mathrm{X}}_i$, }
		\State 1. Move sensor to start point above surface
		\State 2. Rotate sensor to contact angle $(\alpha, \beta)$
		\State 3. \parbox[t]{\dimexpr\linewidth-\algorithmicindent}{%
			Move sensor normal to surface to make contact at depth $z$
		}
		\State 4. \parbox[t]{\dimexpr\linewidth-\algorithmicindent}{%
			Move sensor parallel to surface through translation $(x,y)$ and rotation $\gamma$
		}
		\State 5. Capture sensor tactile image $\boldsymbol{\mathrm{I}}_{i}$
	\EndFor
	\end{algorithmic}
\end{algorithm}

Three distinct data sets were used to develop the pose-and-shear models: a training set of 6000 samples, a validation set of 2000 samples for model selection and hyper-parameter tuning, and a test set of 2000 samples for independently verifying the model performance post training. {\color{black}All data were collected using a 3D-printed flat surface (VeroWhite material, shown later in Figure 6(c)).}

\begin{figure*}[t]
	\centering
	\includegraphics[width=\textwidth,trim=0 30 0 75,clip]{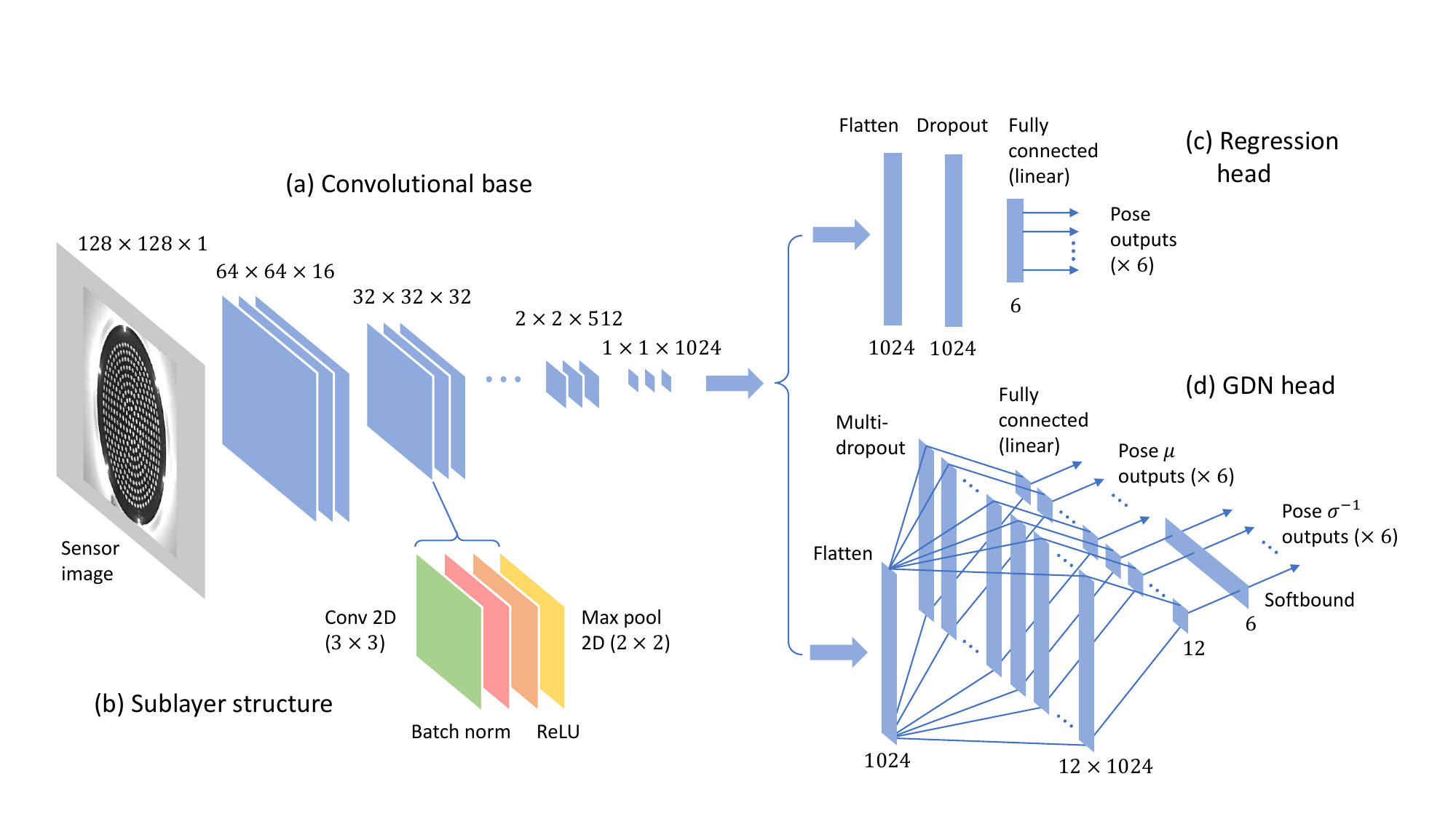}
	\caption{CNN {\color{black}with regression} and GDN architectures used for surface contact pose estimation. (a) Convolutional base for CNN and GDN models. (b) Convolutional block sub-layer structure. (c) CNN multi-output regression head. (d) GDN PDF estimation head.
	\label{fig:cnn_gdn_architecture}}
\end{figure*}

\subsubsection{Pre- and post-processing.}
\label{sec:pre_post_processing}

We used the following steps to collect and pre-process the tactile images of the training, validation and test sets, and to pre-process tactile images after the model is deployed:
\begin{enumerate}[leftmargin=0.5cm]
    \item Collect at $640 \times 480$ pixel resolution and convert to 8-bit grayscale.
	\item Crop to a $430 \times 430$ pixel square enclosing the circular region within which the markers are located.
	\item Apply a $5 \times 5$ median blur to remove sensor noise.
	\item Apply an adaptive threshold to binarize the image.
	\item Resize the images to $128 \times 128$ pixels.
	\item Convert the 8-bit integer pixel intensities to floating point and normalise to lie in the range $[0, 1]$.
\end{enumerate}

We also pre-processed the pose-and-shear labels so that the trained model predictions are in the correct format for the subsequent filtering and controller stages:
\begin{enumerate}[leftmargin=0.5cm]
	\item Convert pose labels from their Euler representations to $4 \times 4$ homogeneous matrices $\boldsymbol{\mathrm{X}}_{fs} \in \mathbb{R}^{4 \times 4}$.
	\item Invert the $4 \times 4$ matrices, so that instead of representing sensor poses in the surface feature frames $\{f\}$ they now represent surface feature poses in the sensor frames $\{s\}$: $\boldsymbol{\mathrm{X}}_{sf}=\boldsymbol{\mathrm{X}}_{fs}^{-1}$.
	\item Convert the inverted matrices to exponential coordinates: $\boldsymbol\xi_{sf}=\ln(\boldsymbol{\mathrm{X}}_{sf})^{\vee} \in \mathbb{R}^{6}$ (see Appendix~\ref{sec:math_prelim}).
\end{enumerate}

\subsubsection{Convolutional neural network for pose-and-shear estimation.}
\label{sec:cnn_model}

Following previous work on tactile pose estimation (\cite{lepora2020optimal, lepora2021pose, lloyd2021goal}), we consider a baseline model using a {\color{black}CNN with a multi-output regression head}. We configured this CNN architecture to be effective for predicting pose and shear, resulting in a sequence of convolutional layer blocks, where each block is composed of a sequence of sub-layers: a $3 \times 3$ 2D convolution; batch normalisation (\cite{ioffe2015batch}); a rectified linear unit (ReLU) activation function; and $2 \times 2$ max pooling. The feature map dimensions are reduced by half at each block as we move forwards through the blocks, due to the max-pooling. Hence, we balance the progressive loss of feature resolution by doubling the number of features in consecutive layer blocks. 

The output of the convolutional base feeds into a fully-connected, multi-output regression head, composed of a flatten layer, dropout layer with dropout probability $p=0.1$ (\cite{srivastava2014dropout}) and a single fully-connected layer with a linear activation function. When a pre-processed sensor image is applied as input to the CNN, it outputs a surface contact pose-and-shear estimate ${\boldsymbol\xi}\in\mathbb{R}^6$. Where required, we convert these estimates ${\boldsymbol\xi}$ from exponential coordinates in the vector space to $4\times 4$ homogeneous matrices using ${\boldsymbol{\mathrm{X}}} = \exp({\boldsymbol\xi}^{\wedge})$ (see Appendix~\ref{sec:math_prelim}).

We train this {\color{black}CNN regression model} by minimising a weighted mean-squared error (MSE) loss function, defined over $N$ training examples and $M=6$ network outputs:
\begin{equation}
	\label{eqn:mse_loss_def}
	\mathrm{MSE}
	\, = \,
	\frac{1}{N} \sum_{i=1}^{N} \sum_{j=1}^{M} \alpha_{j} \left(\xi_{ij}^{\mathrm{label}} - {\xi}_{ij} \right)^{2}.
\end{equation}
Here, $\xi_{ij}=(\xi_{j})_i$ is the ${j}$th pose-and-shear component of the $i$th sample in exponential coordinates, with ${\xi}_{ij}$ the {\color{black}regression} output and ${\xi}_{ij}^{\mathrm{label}}$ its corresponding label. The loss weights $\alpha_{j}$ are hyperparameters that can compensate for different output scales and avoid over-fitting when some outputs have larger errors than others. Through trial and error, we found a good set of weights to be $\boldsymbol\alpha = (1,1,1,100,100,100)$.

We trained {\color{black}these CNN regression} models using the Adam optimizer with a batch size of 16 and a linear rise, polynomial decay (LRPD) learning rate schedule. In our implementation of this schedule, we initialised the learning rate to 10$^{-5}$ and linearly increased it to 10$^{-3}$ over 3 epochs; we then maintained it for a further epoch before decaying it to 10$^{-7}$ over $e_{\mathrm{max}} = 50$ epochs using a $\sqrt{1-{e}/{e_{\mathrm{max}}}}$ polynomial decay weighting factor. We found that a good learning rate schedule can make the training process less sensitive to a particular choice of learning rate and generally improves the performance of the trained model. We used ``early stopping" to terminate the training process when the validation loss reached its minimum value over a ``patience" of 25 epochs.

\subsubsection{Gaussian density network for pose and shear with uncertainty.}
\label{sec:gdn_model}

In this paper, we introduce a modification of the CNN regression head to estimate the parameters of a (Gaussian) pose-and-shear distribution, rather than produce a single-point estimate. This allows us to estimate both the surface contact pose/shear and its associated uncertainty (Figure~\ref{fig:cnn_gdn_architecture}). The motivation for doing this was discussed in our previous work on tactile aliasing (\cite{lloyd-RSS-21}): if we know the uncertainty associated with a pose, this information can be used to reduce the error and uncertainty using other system components such as the Bayesian filter we describe in the next section. {\color{black}In our previous work, we considered a Mixture Density Network (MDN) composed of a mixture of Gaussians. In the present work, we use a single Gaussian to be consistent with assumptions for deriving the update equations for the Bayesian filter in Algorithm~\ref{alg:se3_bayes_filter}.} We refer to this model as a Gaussian Density Network (GDN) because it predicts the parameters of a multivariate Gaussian PDF that captures uncertainty in the pose-and-shear outputs.

Specifically, we use the GDN outputs ${\boldsymbol\mu}_{i}$ and ${\boldsymbol\sigma}_{i}^{-1}$ for the $i$th tactile data sample to estimate the parameters of a multivariate Gaussian PDF over the pose-and-shear components in exponential coordinates $\boldsymbol{\xi_{i}}\in\mathbb{R}^6$: 
\begin{equation}
	\label{eqn:multi_gaussian_pdf}
	p \left( \boldsymbol\xi_{i} \right)
	\, = \,
	\prod_{j=1}^{M} \frac{{\sigma}_{ij}^{-1}}{\sqrt{2\pi}} 
	\exp \left( -\textstyle\frac{1}{2}
	\left({{\sigma}_{ij}^{-1}} \left( \xi_{ij} - {\mu}_{ij} \right) \right)^{2} \right),
\end{equation}
where ${\mu}_{ij} = ({\mu}_{j})_i$ and ${\sigma}_{ij}^{-1} = ({\sigma}_{j}^{-1})_i$. To simplify the model and reduce the amount of training data needed, we have assumed a diagonal covariance matrix, conditioned on the $i$th data sample: $\mathrm{diag}({\boldsymbol\sigma}_{i}^{2})= \mathrm{diag}\left(1/({\sigma}_{i1}^{-1})^2, \cdots, 1/({\sigma}_{iM}^{-1})^2\right)$.

We train the GDN model by minimising a mean negative log likelihood loss function over the label values:
\begin{multline}
    \label{eqn:gdn_mean_nll_loss_def}
	\hspace{-1em}\overline{\mathrm{NLL}}
	\,=\, 
	-\frac{1}{N}\sum_{i=1}^{N} \ln p \left( \boldsymbol\xi_{i}^{\mathrm{label}} \right)\\
	= 
	\frac{1}{N}\sum_{i=1}^{N} \sum_{j=1}^{M} 
	\left[ {\textstyle\frac{1}{2}} \left( {\sigma}_{ij}^{-1} \left( \xi_{ij}^{\mathrm{label}} - {\mu}_{ij} \right) \right)^{2} - \ln {\sigma}_{ij}^{-1} \right] + c,\!\!\!
\end{multline}
where $c=\frac{1}{2}M\ln(2\pi)$ is a constant term that can be ignored. 
Comparing this definition with Equation~\ref{eqn:mse_loss_def}, if ${\sigma}_{ij}^{-1} = \sigma_j^{-1}$ for all $i$, minimizing $\overline{\mathrm{NLL}}$ is equivalent to minimising the weighted MSE. Moreover, the squared inverse standard deviations play the same role as the loss function weights $\boldsymbol\alpha$.

As mentioned above, the GDN model can be viewed as a single-component mixture density network (MDN), which performs a similar function to the GDN but uses a Gaussian mixture model to model the output distribution (\cite{bishop1994mixture, bishop2006pattern}). This is relevant because the difficulties encountered when training MDNs are well-documented and include problems such as training instability and mode collapse (\cite{hjorth1999regularisation, makansi2019overcoming}). To overcome these difficulties, we incorporated several novel extensions to our architecture, described below.

1) First, rather than directly estimating the component means and standard deviations of a multivariate Gaussian pose-and-shear distribution (assuming a diagonal covariance matrix), we instead estimate the means and \emph{inverse} standard deviations. As a result of this, the estimated values appear as products in the mean negative log-likelihood loss function instead of quotients. Otherwise, we found that using a neural network to simultaneously estimate two variables that appear as quotients in a loss function can cause instability or slow progress during training. 

2) We introduce a new \emph{softbound} activation function layer. We use this layer to bound the values of the (inverse) standard deviation within a pre-specified range $[x_{\mathrm{min}}, x_{\mathrm{max}}]$ to prevent it from becoming too large or small:
\begin{multline}
	\label{eqn:softbound_def}
	\mathrm{softbound} \left( x \right)
	\, = \,
	x_{\mathrm{min}}
	+ \mathrm{softplus} \left( x - x_{\mathrm{min}} \right)\\
	- \mathrm{softplus} \left( x - x_{\mathrm{max}} \right),
\end{multline}
where $\mathrm{softplus} \left( x \right)\, = \,\ln \left( 1 + \exp \left( x \right) \right)$. This softbound layer also helps speed-up training and reduce instability:

3) We introduce a novel \emph{multi-dropout} configuration for multi-output neural networks, which allows us to apply distinct dropout probabilities to different outputs. In general, we found that for our pose-and-shear estimation task, dropout is more effective than other forms of regularisation such as L2 regularization, and so we needed a way to vary the amount of dropout across the different outputs. 

Our GDN architecture uses the same convolutional base as the original CNN architecture, but instead feeds its output through a modified GDN head that includes the enhancements discussed above. Inside the GDN head, the output of the convolutional base is flattened and replicated to a set of 12 dropout layers, one for each of the 12 network outputs $(\mu_j)$ and $(\sigma_j^{-1})$. Each of these dropout layers feeds into a distinct single-output, fully-connected output layer with a linear activation function. The 6 outputs for $(\sigma_j^{-1})$ are each passed through a softbound layer that bounds them to the range $\sigma_{j}^{-1} \in$ [10$^{-6}$, 10$^6$]. Since each single-output output layer has its own dedicated dropout layer, we use a higher level of dropout to increase the regularization on the noisier shear-related outputs, and a lower level of dropout on the remaining pose-related outputs. Through manual tuning, we found a good set of dropout probabilities to be $\boldsymbol{p}_{\mu}=(0.7,0.7,0.1,0,0,0.4)$ and 
$\boldsymbol{p}_{\sigma^{-1}}=(0.1,0.1,0,0,0,0.05)$. 

We trained the GDN model the same way as the CNN model, using the Adam optimizer with a batch size of 16 and the same LRPD learning rate schedule. As before, we terminated the training process when the validation loss reached its minimum value over a ``patience" of 25 epochs.

\begin{figure}[t!]
	\centering
	\includegraphics[width=\columnwidth,trim=136 110 120 120,clip]{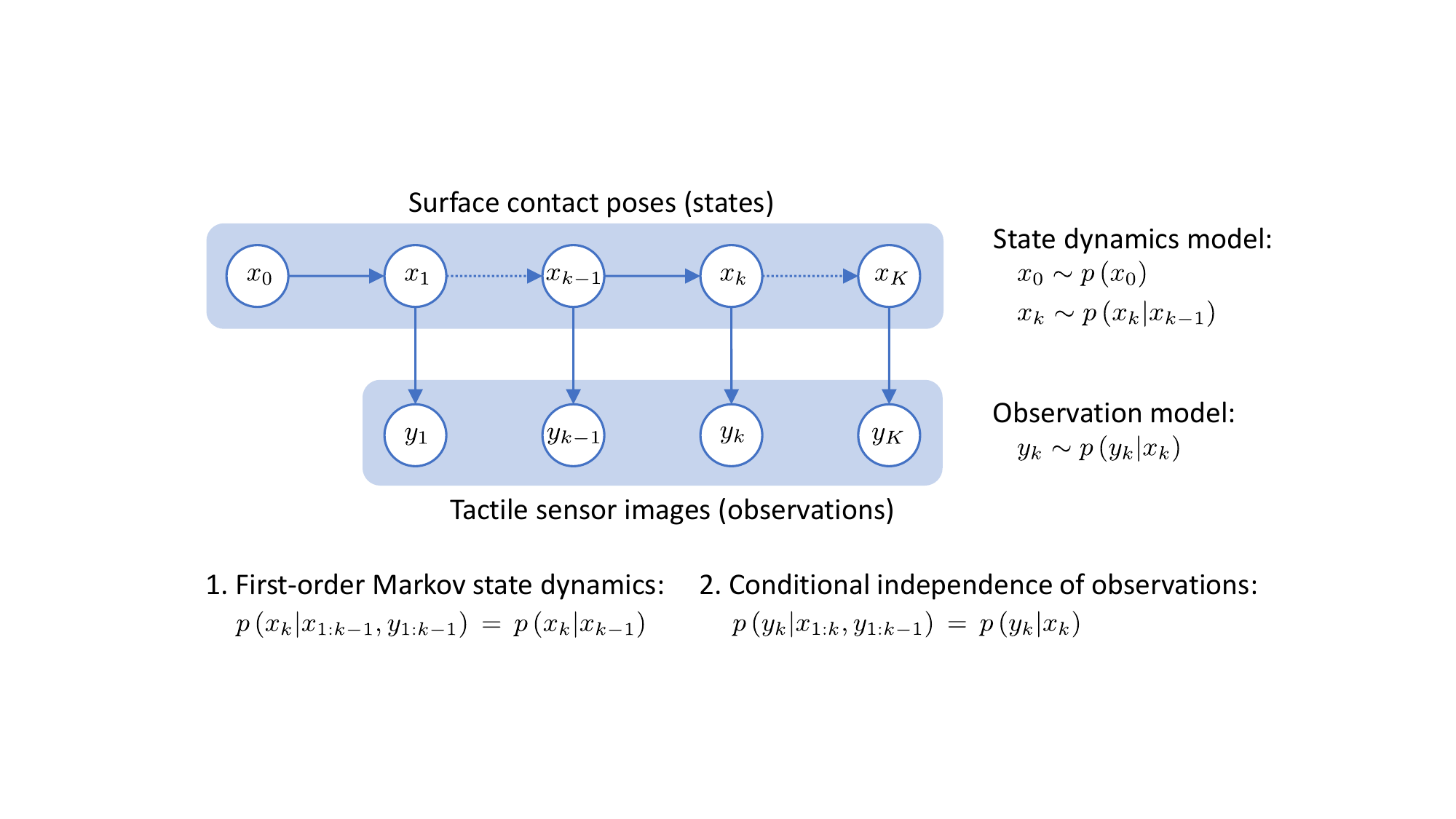}
	\caption{Probabilistic state space model used to describe the relationship between surface contact poses and shears (states) and tactile sensor images (observations) in our Bayesian filter.
	\label{fig:state_space_model}}
\end{figure}

\subsection{Bayesian filtering of pose and shear}
\label{sec:se3_bayes_filter}

\subsubsection{Discriminative Bayesian filtering.}
\label{sec:discriminative_bayes_filter}

We model the sequential pose-and-shear estimation problem using a probabilistic state-space model (Figure~\ref{fig:state_space_model}) that is defined by two interrelated sequences of conditional PDFs. This type of state-space model and inference equations form the basis of many Bayesian filtering algorithms, including the Kalman filter (\cite{kalman1960new, kalman1961new}), extended Kalman filter (EKF) (see \cite{gelb1974applied}), unscented Kalman filter (UKF) (\cite{julier1995new, julier1997new}) and particle filters (see \cite{sarkka2013bayesian}). To simplify the notation in this section, we use lower-case italic letters to represent continuous random variables, regardless of whether they are scalars, vectors or $SE(3)$ elements.

The \emph{state dynamics model} describes how states $x_{k} \sim p(x_{k}|x_{k-1})$ are transformed across steps $k=1,2,\cdots$ beginning from $x_{0} \sim p(x_{0})$. The \emph{observation model} describes how observations $y_{k}\sim p(y_{k}|x_{k})$ are related to the state $x_k$ at time step $k$. As is conventional for this type of model, we assume \emph{first-order Markov state dynamics} and \emph{conditional independence of observations}:
\begin{equation}
	\label{eqn:bayes_filter_assumptions}
	\begin{split}
		p \left( x_{k} | x_{1:k-1}, y_{1:k-1} \right)
		& \, = \,
		p \left( x_{k} | x_{k-1} \right),\\
		p \left( y_{k} | x_{1:k}, y_{1:k-1} \right)
		& \, = \,
		p \left( y_{k} | x_{k} \right).\nonumber
    \end{split}
\end{equation}

The (conditional) PDF over states $x_{k}$ can then be inferred recursively from the following pair of equations:
\begin{align}
	\label{eqn:bayes_filter_predict}	
    p(x_{k}|y_{1:k-1}) &=\int p(x_{k}|x_{k-1}) \, p(x_{k-1}|y_{1:k-1}) \, dx_{k-1},\\
    \label{eqn:bayes_filter_correct}
    p(x_{k}|y_{1:k}) &= \frac{1}{Z_{k}} \, p(y_{k}|x_{k}) \, p(x_{k}|y_{1:k-1}),
\end{align}
where the normalisation coefficient $Z_{k}$ is
\begin{equation}
	\label{eqn:bayes_filter_correct_normaliser}
	Z_{k} \, = \, \int p(y_{k}|x_{k}) \, p(x_{k}|y_{1:k-1}) \, dx_{k}.
\end{equation}
Here, the first relation (Equation~\ref{eqn:bayes_filter_predict}) is known as the \emph{prediction step} or the \emph{Chapman--Kolmogorov} equation, and it computes an interim PDF over states $x_{k}$ at time step $k$ given observations up to time step $k-1$. Since the integral marginalises over the state distribution at the previous time step, it can be viewed as computing the PDF of the \emph{probabilistic transformation} of the previous state by the state dynamics model. Meanwhile, the second relation (Equation~\ref{eqn:bayes_filter_correct}) is known as the \emph{correction step} and uses Bayes' rule to compute the PDF over states at time step $k$ given observations up to time step $k$. This step can be viewed as \emph{probabilistic fusion} of the current observation with the interim state computed in the prediction step. 

\begin{algorithm}[t]
	\caption{$SE(3)$ discriminative Bayesian filter}
	\label{alg:se3_bayes_filter}
	\begin{algorithmic}
	\Require A sequence of surface pose estimates $\big(\bar{\boldsymbol{\mathrm{X}}}_{k}^{{\mathrm{obs}}}, \boldsymbol\Sigma_{k}^{{\mathrm{obs}}}\big)$ and sensor poses $\boldsymbol{\mathrm{X}}_{k}^{\mathrm{sens}}$;
	$k = 0,1,2,\ldots$
	\Ensure A sequence of filtered surface pose estimates $\big(\bar{\boldsymbol{\mathrm{X}}}_{k}^{\mathrm{fil}}, \boldsymbol\Sigma_{k}^{\mathrm{fil}}\big)$; $k = 0,1,2,\ldots$
	\State \textit{Initialisation:} $p(x_{0}^{\mathrm{fil}}) = p(x_{0}^{\mathrm{bel}}) = p(x_{0}^{{\mathrm{obs}}})$
    \State\hspace{\algorithmicindent} $\big(\bar{\boldsymbol{\mathrm{X}}}_{0}^{\mathrm{fil}}\!\!, \boldsymbol\Sigma_{0}^{\mathrm{fil}}\big)=\big(\bar{\boldsymbol{\mathrm{X}}}_{0}^{\mathrm{bel}}\!\!, \boldsymbol\Sigma_{0}^{\mathrm{bel}}\big)=\big(\bar{\boldsymbol{\mathrm{X}}}_{0}^{{\mathrm{obs}}}\!\!, \boldsymbol\Sigma_{0}^{{\mathrm{obs}}}\big)$
	\For{each $\big(\bar{\boldsymbol{\mathrm{X}}}_{k}^{{\mathrm{obs}}}, \boldsymbol\Sigma_{k}^{{\mathrm{obs}}}\big)$}
		\State \textit{Prediction step:}
		\State \parbox[t]{\dimexpr\linewidth-\algorithmicindent}{%
		Compute $p(x_{k}^{\mathrm{bel}})\!=\!\int p(x_{k}^{\mathrm{bel}} | x_{k - 1}^{\mathrm{fil}}) \, p(x_{k - 1}^{\mathrm{fil}}) \, dx_{k - 1}^{\mathrm{fil}}$\\
		using:}
		\State\hspace{\algorithmicindent} $\bar{\boldsymbol{\mathrm{X}}}_{k}^{\mathrm{bel}} = \bar{\boldsymbol{\mathrm{T}}}_{k}\, \bar{\boldsymbol{\mathrm{X}}}_{k-1}^{\mathrm{fil}}$
		\State\hspace{\algorithmicindent} $\boldsymbol\Sigma_{k}^{\mathrm{bel}} = 
        \mathrm{Ad}(\bar{\boldsymbol{\mathrm{T}}}_{k}) \,\boldsymbol\Sigma_{k - 1}^{\mathrm{fil}}\,  
        \mathrm{Ad}(\bar{\boldsymbol{\mathrm{T}}}_{k})^{\top} + \boldsymbol{\Sigma}_{\boldsymbol{\phi}}$
		\State\hspace{\algorithmicindent} and $\bar{\boldsymbol{\mathrm{T}}}_{k} = (\boldsymbol{\mathrm{X}}_{k}^{\mathrm{sens}})^{- 1} \boldsymbol{\mathrm{X}}_{k - 1}^{\mathrm{sens}}$
		\State \textit{Correction step:}
		\State \parbox[t]{\dimexpr\linewidth-\algorithmicindent}{%
	  	Compute $p(x_{k}^{\mathrm{fil}}) = \frac{1}{Z}p\big(x_{k}^{{\mathrm{obs}}}\big) \, p\big(x_{k}^{\mathrm{bel}}\big)$
	  	using the $SE(3)$ data fusion algorithm (Algorithm~\ref{alg:se3_data_fusion}) to find:
		} 
		\State\hspace{\algorithmicindent}\!\!\!$\big(\bar{\boldsymbol{\mathrm{X}}}_{k}^{\mathrm{fil}}\!\!, \boldsymbol\Sigma_{k}^{\mathrm{fil}}\big)$ from $\big(\bar{\boldsymbol{\mathrm{X}}}_{k}^{{\mathrm{obs}}}\!\!,\boldsymbol\Sigma_{k}^{{\mathrm{obs}}}\big)$,$\big(\bar{\boldsymbol{\mathrm{X}}}_{k}^{\mathrm{bel}}\!\!,\boldsymbol\Sigma_{k}^{\mathrm{bel}}\big)$
	\EndFor
	\end{algorithmic}
\end{algorithm}

The observation model can be viewed as a \emph{generative} model because it specifies how to generate observations $y_{k}$ given a state $x_{k}$. However, as pointed out in \cite{burkhart2020discriminative}, we do not always have access to such a model but instead have a \emph{discriminative} model of the form $x_{k} \sim p(x_{k}|y_{k})$. This alternative type of model corresponds to the situation we are dealing with here, where the GDN model estimates a PDF over states (poses and shears) given an observation (tactile image). To use this type of model in the Bayesian filter equations, we must first invert the original observation model using a second application of Bayes' rule and then substitute the result back into the correction step of Equation~\ref{eqn:bayes_filter_predict} to give a modified correction step:
\begin{equation}
	\label{eqn:bayes_filter_prob_fusion}
	p(x_{k}|y_{1:k})
	\, = \,
	\frac{1}{Z_{k}^{\prime}} \,
	\frac{p(x_{k}|y_{k}) \, p(x_{k}|y_{1:k-1})}{p(x_{k})}.
\end{equation}
Here, the $p(y_{k})$ term has been absorbed in the modified normalisation constant $Z_{k}^{\prime}$. If we also assume a constant prior $p(x_{k})$, we can further simplify this modified correction step to a normalised product of PDFs:
\begin{equation}
	\label{eqn:bayes_filter_prob_fusion_approx}
	p(x_{k}|y_{1:k})
	\, = \,
	\frac{1}{Z_{k}^{\prime\prime}} \,
	p(x_{k}|y_{k}) \, p(x_{k}|y_{1:k-1}),
\end{equation}
where the constant prior $p(x_{k})$ has been absorbed in the modified normalisation constant $Z_{k}^{\prime \prime}$.

Then we can reinterpret Equations~\ref{eqn:bayes_filter_predict} and \ref{eqn:bayes_filter_prob_fusion_approx} as a discriminative Bayesian filter that updates a filtered state PDF $p\big(x_{k}^{\mathrm{fil}}\big)$ over steps $k=0,1,2,\cdots$. This filter uses an intermediate computation of the belief PDF $p\big(x_{k}^{\mathrm{bel}}\big)$ in the prediction step, which is fused with the observation PDF $p\big(x_{k}^{{\mathrm{obs}}}\big)$ in the correction step:
\begin{align}    
    \label{eqn:bayes_filter_prediction}
    p\big(x_{k}^{\mathrm{bel}}\big) &\,=\, \int p\big(x_{k}^{\mathrm{bel}} | x_{k - 1}^{\mathrm{fil}}\big) \, p\big(x_{k - 1}^{\mathrm{fil}}\big) \, dx_{k - 1}^{\mathrm{fil}},\\    	
    \label{eqn:bayes_filter_correction}
    p\big(x_k^{\mathrm{fil}}\big) &\,=\, \frac{1}{Z}\,p\big(x_{k}^{{\mathrm{obs}}}\big) \, p\big(x_{k}^{\mathrm{bel}}\big).
\end{align}

{\color{black}For Kalman filters, it is standard to} assume Gaussian PDFs. Then the filter is equivalent to updating the means and covariance matrix for $x_{k}^{\mathrm{fil}}\sim \mathcal{N}\big(\boldsymbol{\mu}_k^{\mathrm{fil}},\boldsymbol{\Sigma}_k^{\mathrm{fil}}\big)$ using an intermediate estimate of the belief PDF $x_{k}^{\mathrm{bel}}\sim \mathcal{N}\big(\boldsymbol{\mu}_k^{\mathrm{bel}},\boldsymbol{\Sigma}_k^{\mathrm{bel}}\big)$ that is then fused with the observation PDF $x_{k}^{{\mathrm{obs}}}\sim \mathcal{N}\big(\boldsymbol{\mu}_k^{{\mathrm{obs}}},\boldsymbol{\Sigma}_k^{{\mathrm{obs}}}\big)$. The state dynamics model $p\big(x_{k}^{\mathrm{bel}} | x_{k-1}^{\mathrm{fil}}\big)$ can be as simple as shifting the filtered mean $\boldsymbol{\mu}_{k-1}^{\mathrm{fil}}$ by an approximate displacement and increasing the covariance matrix $\boldsymbol{\Sigma}_{k-1}^{\mathrm{fil}}$ by a constant amount.

A similar approach has been used to derive some discriminative variations of the Kalman filter, referred to as the Discriminative Kalman Filter (DKF) and robust DKF (\cite{burkhart2020discriminative}). However, in that work the authors modified the inference equations \emph{after} specialising the state-space model to a linear-Gaussian model. {\color{black}We do not follow that approach here because we cannot specialise to standard Gaussian state distributions due to the complexities with $SE(3)$ state variables discussed in the next section.} However, it is nevertheless reassuring to know that if we had assumed a linear-Gaussian model with our more general equations (Equation~\ref{eqn:bayes_filter_predict} together with Equation~\ref{eqn:bayes_filter_prob_fusion} or Equation~\ref{eqn:bayes_filter_prob_fusion_approx}), we would obtain the same filter update equations as other works.

\begin{algorithm}[t]
	\caption{$SE(3)$ data fusion}
	\label{alg:se3_data_fusion}
	\begin{algorithmic}
	\Require $SE(3)$ factor PDF parameters:\  $(\bar{\boldsymbol{\mathrm{X}}}_{1},\boldsymbol{\mathrm{\Sigma}}_{1}),(\bar{\boldsymbol{\mathrm{X}}}_{2},\boldsymbol{\mathrm{\Sigma}}_{2})$
	\Ensure Normalised product PDF parameters:\  $(\bar{\boldsymbol{\mathrm{X}}}_{*},\boldsymbol{\mathrm{\Sigma}}_{*})$
	\State \textit{Initialise trial solution (operating point):}\  $\bar{\boldsymbol{\mathrm{X}}} = \bar{\boldsymbol{\mathrm{X}}}_{1}$
	\While {not converged}
		\State $\boldsymbol\xi_{1} = \ln \left(\bar{\boldsymbol{\mathrm{X}}} \bar{\boldsymbol{\mathrm{X}}}_{1}^{-1} \right)^{\vee}$\!, \ \ $\boldsymbol\xi_{2} = \ln \left(\bar{\boldsymbol{\mathrm{X}}} \bar{\boldsymbol{\mathrm{X}}}_{2}^{-1} \right)^{\vee}$
		\State $\mathcal{J}_{1}^{-1} \equiv \mathcal{J}(\boldsymbol\xi_{1})^{-1}$,\ \ \ $\mathcal{J}_{2}^{-1} \equiv \mathcal{J}(\boldsymbol\xi_{2})^{-1}$ \\
	  \hspace{2em}using $\mathcal{J}(\boldsymbol\xi)^{-1}\!\approx\boldsymbol{\mathrm{1}} - \frac{1}{2}\boldsymbol\xi^{\curlywedge} + \frac{1}{12}(\boldsymbol\xi^{\curlywedge})^{2}$, $\boldsymbol\xi^{\curlywedge}\!=\mathrm{ad}\big( \boldsymbol\xi^{\wedge} \big)$
		\State ${\boldsymbol{\mathrm{\Sigma}}}
			=
			\left(
			\mathcal{J}_{1}^{-\top}
			\boldsymbol{\mathrm{\Sigma}}_{1}^{-1}
			\mathcal{J}_{1}^{-1}
			+
			\mathcal{J}_{2}^{-\top}
			\boldsymbol{\mathrm{\Sigma}}_{2}^{-1}
			\mathcal{J}_{2}^{-1}
			\right)^{-1}$
		\State $\boldsymbol{\mathrm{\mu}}
			=
			-\boldsymbol{\mathrm{\Sigma}}
			\left(
			\mathcal{J}_{1}^{-\top}
			\boldsymbol{\mathrm{\Sigma}}_{1}^{-1} \boldsymbol{\mathrm{\xi}}_{1}
			+
			\mathcal{J}_{2}^{-\top}
			\boldsymbol{\mathrm{\Sigma}}_{2}^{-1} \boldsymbol{\mathrm{\xi}}_{2}
			\right)$
		\State \textit{Update trial solution (operating point):}
		\State $\bar{\boldsymbol{\mathrm{X}}}
		\, \leftarrow \,
		\exp \left( \boldsymbol\mu^{\wedge} \right) \bar{\boldsymbol{\mathrm{X}}}$
	\EndWhile
	\State \textit{Return solution at convergence:} $(\bar{\boldsymbol{\mathrm{X}}}_{*},{\boldsymbol{\mathrm{\Sigma}}}_{*}) = (\bar{\boldsymbol{\mathrm{X}}},{\boldsymbol{\mathrm{\Sigma}}})$
	\end{algorithmic}
\end{algorithm}

\subsubsection{Discriminative Bayesian filtering in SE(3).}
\label{sec:discriminative_bayesian_filtering}

We now describe how the discriminative Bayesian filtering from the previous section is implemented on a sequence of $SE(3)$ pose-and-shear observations with uncertainty. Algorithm~\ref{alg:se3_bayes_filter} iteratively implements the filter by applying the prediction step in Equation~\ref{eqn:bayes_filter_prediction} and correction step in Equation~\ref{eqn:bayes_filter_correction} to the sequence of observation PDFs. As part of this computation, Algorithm~\ref{alg:se3_data_fusion} implements the correction step as this requires combining two $SE(3)$ PDFs, which is complicated by the fusion of two $SE(3)$ PDFs being no longer of the same form, so an approximate method is needed (see Appendix~\ref{sec:se3_probabilistic_fusion}).

We assume a sequence of observations of the surface pose and shear with uncertainty $\big(\bar{\boldsymbol{\mathrm{X}}}_{k}^{{\mathrm{obs}}}, \boldsymbol\Sigma_{k}^{{\mathrm{obs}}}\big)$ estimated by the GDN model from the $k$th tactile sensor image $\boldsymbol{\mathrm{I}}_{k}$, along with a corresponding sequence of sensor poses $\boldsymbol{\mathrm{X}}_{k}^{\mathrm{sens}}$ from the robot arm kinematics. The pose-and-shear (mean) estimates from the GDN model also need converting from the $\boldsymbol\mu_k\in\mathbb{R}^{6}$ exponential coordinate representation of $\mathfrak{se}(3)$ to \mbox{$4 \times 4$} matrices $\bar{\boldsymbol{\mathrm{X}}}_{k}\in SE(3)$ representing the translational and rotational components of the pose and shear (Appendix~\ref{sec:math_prelim}):
\begin{equation}
\label{eq:se3_inverse_transform}
\bar{\boldsymbol{\mathrm{X}}}_{k}^{{\mathrm{obs}}} \!= \exp\big({\boldsymbol\mu}_{k}^{\wedge}\big),\ \ \  
{\boldsymbol{\mathrm{\Sigma}}}_{k}^{{\mathrm{obs}}} \!= \mathcal{J}\big(\boldsymbol\mu_{k}\big) \,\mathrm{diag}({\boldsymbol\sigma}_{k}^{2})\, \mathcal{J}\big(\boldsymbol\mu_{k}\big)^{\top}\!.
\end{equation}
Here, the covariance matrix ${\boldsymbol{\mathrm{\Sigma}}}_{k}=\mathrm{diag}({\boldsymbol\sigma}_{k}^{2})\in\mathbb{R}^{6\times 6}$ is transformed from the GDN output by the Jacobian $\mathcal{J}\big(\boldsymbol\mu\big)$ because our chosen method of representing $SE(3)$ random variables is to use a mean that is left-perturbed by a zero-mean Gaussian random variable. With reference to Appendix~\ref{sec:se3_probabilistic_transform}, we invert the covariance expression in Equation~\ref{eqn:glob_tangent_space_pdf} to find this covariance matrix ${\boldsymbol{\mathrm{\Sigma}}}_k$.

The sequence of sensor poses $\boldsymbol{\mathrm{X}}_{k}^{\mathrm{sens}}$ is used in the state dynamics model to find the deterministic component $\bar{\boldsymbol{\mathrm{T}}}_{k} \in SE(3)$ of the probabilistic transformation from the change in sensor pose between steps $k-1$ and $k$:
\begin{equation}
\bar{\boldsymbol{\mathrm{T}}}_{k} = \big(\boldsymbol{\mathrm{X}}_{k}^{\mathrm{sens}}\big)^{- 1} \boldsymbol{\mathrm{X}}_{k - 1}^{\mathrm{sens}},
\end{equation}
which we assume approximates the change in object pose between those steps. Around this deterministic component, a zero-mean Gaussian noise term $\boldsymbol{\phi}$ (with covariance $\boldsymbol{\Sigma}_{\boldsymbol{\phi}}$) represents the uncertainty in the change in object pose between steps ({\em e.g.} due to motion of the object relative to the sensor). Unless otherwise specified, we use the following state dynamics noise covariance in the Bayesian filter algorithm for the experiments in this paper:
\begin{equation}
	\label{eqn:state_dynamics_noise_cov}
	\boldsymbol\Sigma_{\boldsymbol\phi}
	=
 \sigma_{\boldsymbol{\phi}}^{2}\,\,\boldsymbol{1}_{6\times 6},
\end{equation}
with $\sigma_{\phi}=0.5$, which is equivalent to using a standard deviation ({\em i.e.} uncertainty) of 0.5\,mm/s and 0.5\,deg/s in the three translational and three rotational components of $\boldsymbol{\phi}_{k}$.

The observations and states described above are combined using an $SE(3)$ Bayesian filter (Algorithm~\ref{alg:se3_bayes_filter}) that computes the filtered surface pose and shear with uncertainty. The prediction step implements a state dynamics model that updates $\big(\bar{\boldsymbol{\mathrm{X}}}_{k}^{\mathrm{bel}}, \boldsymbol\Sigma_{k}^{\mathrm{bel}}\big)$ according to the change in sensor pose $\bar{\boldsymbol{\mathrm{T}}}_{k}$, adding a noise term $\boldsymbol\Sigma_{\boldsymbol\phi}$ to the covariance matrix. The correction step updates $\big(\bar{\boldsymbol{\mathrm{X}}}_{k}^{\mathrm{fil}}, \boldsymbol\Sigma_{k}^{\mathrm{fil}}\big)$ by combining the belief output of the prediction step and the observed poses and shears with uncertainty, using an approximate $SE(3)$ data fusion method~(Algorithm~\ref{alg:se3_data_fusion}).

The derivation of the prediction step of the $SE(3)$ Bayesian filter is given in Appendix~\ref{sec:se3_probabilistic_transform} and the derivation of the correction/data fusion step is given in Appendix~\ref{sec:se3_probabilistic_fusion}.

\subsection{Feedforward-feedback control of pose and shear}
\label{sec:se3_feedforward_feedback_control}

\begin{figure*}
	\centering
	\includegraphics[width=\textwidth,trim=0 30 0 0,clip]{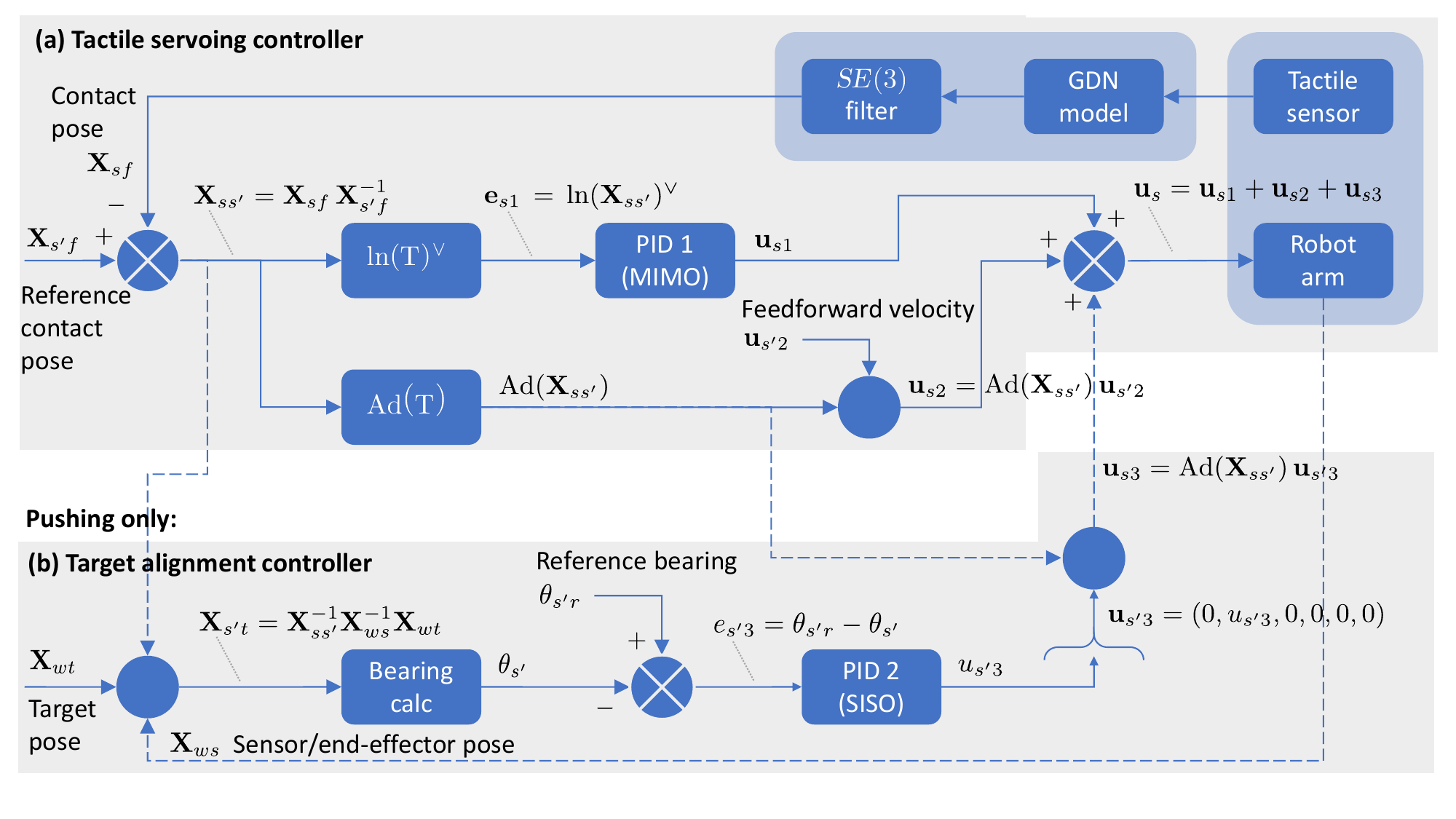}
	\caption{(a) Tactile servoing controller used for all tasks, which is the sole controller for the object tracking and surface following tasks. (b) For the tactile pushing controller the tactile servoing controller is supplemented  with a target alignment controller.
	\label{fig:pushing_controller}}
\end{figure*}

\subsubsection{Feedforward-feedback control in SE(3).}
\label{sec:tangent_space_control}

In our past work, we made extensive use of feedback control systems for pose-based tactile servo control (\cite{lepora2021pose, lloyd2021goal}). For the feedback control, we defined the pose error as the $SE(3)$ transformation that moves the observed sensor pose to a reference pose in the same coordinate frame. We do the same here, with pose error:
\begin{equation}
	\label{eqn:local_coord_frame_error}
	\boldsymbol{\mathrm{E}}_{\boldsymbol{\mathrm{X}}}
    \,=\,
	\boldsymbol{\mathrm{X}}^{-1} \boldsymbol{\mathrm{X}}_{\mathrm{ref}},
\end{equation}
where $\boldsymbol{\mathrm{E}}_{\boldsymbol{\mathrm{X}}}$ and $\boldsymbol{\mathrm{X}}_{\mathrm{ref}}$ are specified in the local frame associated with the pose $\boldsymbol{\mathrm{X}}\in SE(3)$. Then the pose error right-multiplies the observed pose $\boldsymbol{\mathrm{X}}$ to give the reference pose $\boldsymbol{\mathrm{X}}_{\mathrm{ref}}=\boldsymbol{\mathrm{X}}\,\boldsymbol{\mathrm{E}}_{\boldsymbol{\mathrm{X}}}$.


To define the control operations, we project this pose error into the exponential coordinates for the Lie algebra $\mathfrak{se}(3)$, mapped onto the vector tangent space $\mathbb{R}^6$:
\begin{equation}
	\label{eqn:local_tangent_space_error}
	\boldsymbol{\mathrm{e}}_{\boldsymbol{\mathrm{X}}}
	\,=\,
	\ln(\boldsymbol{\mathrm{X}}^{-1} \boldsymbol{\mathrm{X}}_{\mathrm{ref}})^{\vee}.
\end{equation}
Likewise, $\boldsymbol{\mathrm{e}}_{\boldsymbol{\mathrm{X}}}$ is regarded as the pose error in the local tangent space to the $SE(3)$ transformations at pose $\boldsymbol{\mathrm{X}}\in SE(3)$. This error is also the right-perturbation that transforms the observed pose $\boldsymbol{\mathrm{X}}$ to the reference pose $\boldsymbol{\mathrm{X}}_{\mathrm{ref}}=\boldsymbol{\mathrm{X}}\exp(\boldsymbol{\mathrm{e}}_{\boldsymbol{\mathrm{X}}}^{\wedge})$.

This control signal $\boldsymbol{\mathrm{e}}_{\boldsymbol{\mathrm{X}}}\in\mathbb{R}^6$ can be used to directly control the robot for velocity-based control, or treated as a right-perturbation of the current $SE(3)$ pose for position-based control. Since the error is defined in a Euclidean vector space $\mathbb{R}^6$, we can employ all the control frameworks that have been developed for such spaces ({\em e.g.} state feedback). Another advantage of using this representation is that for velocity-based control, it can be convenient to use the control signals generated in these spaces to directly control the robot.


In the case of multi-input multi-output (MIMO) proportional control, we use Equation~\ref{eqn:local_tangent_space_error} to map the observed pose $\boldsymbol{\mathrm{X}}\in SE(3)$ to a 6-component vector $\boldsymbol{\mathrm{e}}_{\boldsymbol{\mathrm{X}}}\in \mathbb{R}^6$, and then compute the control signal using $\boldsymbol{\mathrm{u}}(t) = \,\boldsymbol{\mathrm{K}}_{p} \, \boldsymbol{\mathrm{e}}_{\boldsymbol{\mathrm{X}}}(t)$, where $\boldsymbol{\mathrm{K}}_{p}$ is a $6 \times 6$ diagonal gain matrix that contains the corresponding proportional gain coefficients. For full MIMO proportional-integral-derivative (PID) control, we use:
\begin{equation}
	\label{eqn:ff_fb_pid_control}
	\boldsymbol{\mathrm{u}}(t)
	= \boldsymbol{\mathrm{v}}(t)
	+ \boldsymbol{\mathrm{K}}_{p} \, \boldsymbol{\mathrm{e}}_{\boldsymbol{\mathrm{X}}}(t)
	+ \boldsymbol{\mathrm{K}}_{i} \int_{t'=0}^{t}\!\!\!\!\!\! \boldsymbol{\mathrm{e}}_{\boldsymbol{\mathrm{X}}}(t') \, dt^{\prime}
	+ \boldsymbol{\mathrm{K}}_{d} \, \frac{d \boldsymbol{\mathrm{e}_{\boldsymbol{\mathrm{X}}}}}{dt}(t),
\end{equation}
where $\boldsymbol{\mathrm{K}}_{i}$ and $\boldsymbol{\mathrm{K}}_{d}$ are the $6 \times 6$ diagonal gain matrices associated with the integral and derivative errors at time $t$. For this type of controller, we include a feedforward term $\boldsymbol{\mathrm{v}}(t)$ that can generate a control signal in the absence of any error. This term is useful in surface following tasks, where the tactile sensor on the robot arm should move tangentially to a surface while the sensor remains normal to the surface at a fixed contact depth. Similarly, for object pushing tasks, the tactile sensor on the robot arm should move forwards oriented normal to the contacted surface of the pushed object.

Since our system operates in discrete time, we typically use simple backward-Euler approximations for computing the integral and derivative errors. To reduce noise in the error signal before computing the derivative, we smooth the error using an exponentially-weighted moving average filter with decay coefficient 0.5. We also sometimes clip the integral error between pre-defined limits to mitigate any integral wind-up problems, and clip the output to limit the control signal range. Details of gain coefficients, error or output clipping ranges, feedback reference poses and feedforward velocities (velocity twists) are provided in Appendix~\ref{sec:controller_parameters}.

\subsubsection{Tactile servoing controller.}
\label{sec:servoing_controller}

For object tracking and surface following, we use a tactile servoing controller (Figure~\ref{fig:pushing_controller}, top part only) that performs MIMO feedforward-feedback PID control as described in the previous section (see Equation~\ref{eqn:ff_fb_pid_control}). The goal of this controller is to align the sensor with a reference contact pose in the surface feature frame, while at the same time moving it with a feedforward velocity (set to zero for object tracking) specified in relation to the desired pose. The reference contact pose is usually set so that the sensor is normal to the surface at a fixed contact depth. For surface following, the feedforward velocity is usually set to be tangential to the surface. The overall effect is that the sensor moves smoothly to track or follow a surface while maintaining normal contact at a constant depth. 

In each control cycle, we start by computing the $SE(3)$ error in the sensor coordinate frame using:
\begin{equation}
	\label{eqn:se3_tactile_servo_error}
	\boldsymbol{\mathrm{E}}_{\boldsymbol{\mathrm{X}}}
    \,=\, \boldsymbol{\mathrm{X}}_{ss^{\prime}}
	\, = \,
	\boldsymbol{\mathrm{X}}_{fs}^{-1} \, \boldsymbol{\mathrm{X}}_{s^{\prime}f}^{-1}
     \, = \,
	\boldsymbol{\mathrm{X}}_{sf} \, \boldsymbol{\mathrm{X}}_{s^{\prime}f}^{-1},
\end{equation}
where $\boldsymbol{\mathrm{X}}_{sf}$ is the observed feature pose ({\em i.e.} the surface contact pose) in the current sensor frame that is predicted by the GDN model and subsequently filtered by the Bayesian filter; the other term $\boldsymbol{\mathrm{X}}_{\mathrm{ref}}=\boldsymbol{\mathrm{X}}_{fs^{\prime}}=\boldsymbol{\mathrm{X}}_{s^{\prime}f}^{-1}$ is the target/reference sensor pose in the feature frame. This error is then mapped onto the $\mathbb{R}^6$ representation of $\mathfrak{se}(3)$ using the logarithmic map (Equation~\ref{eqn:local_tangent_space_error}). Then the transformed error is sent to a 6-channel MIMO PID controller with the resulting control signal added to the feedforward velocity. 


For surface following, we set the reference sensor frame so that its $z$-axis is normal to and pointing towards the surface and the feedforward velocity lies in the $xy$-plane of that frame (tangential to the surface). The adjoint representation of the $SE(3)$ error is used to map the feedforward velocity to the observed sensor frame before adding it to the feedback signal, so that $\boldsymbol{\mathrm{v}}$ in Equation~\ref{eqn:ff_fb_pid_control} is ${\rm Ad}(\boldsymbol{\mathrm{X}}_{ss^{\prime}})\boldsymbol{\mathrm{u}}_{s'2}$. Finally, the resulting control signal is used to update the robot end-effector velocity during each control cycle.

\subsubsection{Tactile pushing controller.}
\label{sec:pushing_controller}

For pushing objects across a surface towards a target, we augment the tactile servoing controller with an additional feedback control element that we refer to as the \emph{target alignment controller} (Figure~\ref{fig:pushing_controller}, bottom part). The target alignment controller tries to steer the object towards the target as it is pushed forward, using sideways tangential movements while the sensor remains in frictional contact with the object. In this configuration, the controller feedforward velocity is specified normal to and into the object surface (rather than tangential to the surface as was done for surface following). The combined effect of the tactile servoing and target alignment controllers is to get the sensor to push the object towards the target point while trying to maintain normal contact with the pushed object's surface. Since the tactile serviong controller was discussed in the previous section, we only describe its integration with the target alignment controller here.


The object pushing target is specified as a target pose $\boldsymbol{\mathrm{X}}_{wt}$ in the robot work frame $\{w\}$. The target pose is transformed to the reference sensor frame using the sensor pose $\boldsymbol{\mathrm{X}}_{ws}$ obtained from the robot ({\em i.e.} proprioceptive information), and the sensor error $\boldsymbol{\mathrm{X}}_{ss^{\prime}}$ computed by the tactile servoing controller in Equation~\ref{eqn:se3_tactile_servo_error}:
\begin{equation}
	\label{eqn:transformed_target_pose}
	\boldsymbol{\mathrm{X}}_{s^{\prime}t}
	\,=\,
	\boldsymbol{\mathrm{X}}_{ss^{\prime}}^{-1}\,\, \boldsymbol{\mathrm{X}}_{ws}^{-1} \,\,\boldsymbol{\mathrm{X}}_{wt}.
\end{equation}
The target bearing and distance are computed in the reference sensor frame using:
\begin{equation}
	\label{eqn:target_bearing_distance}
	\begin{gathered}
		\theta_{s^{\prime}} = \mathrm{atan2} \left( y, z \right),\ \ \ 
		r_{s^{\prime}} = \sqrt{ y^{2} + z^{2} },
	\end{gathered}
\end{equation}
where $y$ and $z$ are the target pose translation components extracted from $\boldsymbol{\mathrm{X}}_{s^{\prime}t}\in SE(3)$. The target bearing is subtracted from the reference bearing, which is zero in our case $\theta_{s^{\prime}r} = 0$, to obtain the bearing error in the reference sensor frame. This error is sent to a single-input single-output (SISO) PID controller, which generates a scalar control signal that is used as the tangential $y$-component of the velocity control signal in the reference sensor frame (with other components zero). Since the tactile servoing controller generates a control signal relative to the current sensor pose, the target alignment control signal must be transformed from the reference sensor frame to that of the current sensor frame. We do this using the adjoint representation of the $SE(3)$ error, in the same way that we transform the feedforward velocity signal in the tactile servoing controller. The transformed target alignment control signal is then added to the tactile servoing control signal. Finally, we use the resulting control signal to update the robot end-effector velocity during each control cycle. 


As in our previous work on pushing (\cite{lloyd2021goal}), we zero the output of the target alignment controller when the sensor is less than a pre-defined distance $\rho^{*}$ away from the target (manually tuned to $\rho^{*}=120$\,mm), so as to maintain stability close to the target. After this point, only the tactile servoing controller remains active. The pushing sequence is terminated when the centre of the sensor tip is closer than its radius of 20\,mm from the target. This ensures that the sensor-object contact point is moved close to the target with minimal overshooting.

\begin{figure}[t]
	\centering
	\includegraphics[width=\columnwidth,trim=30 75 150 80,clip]{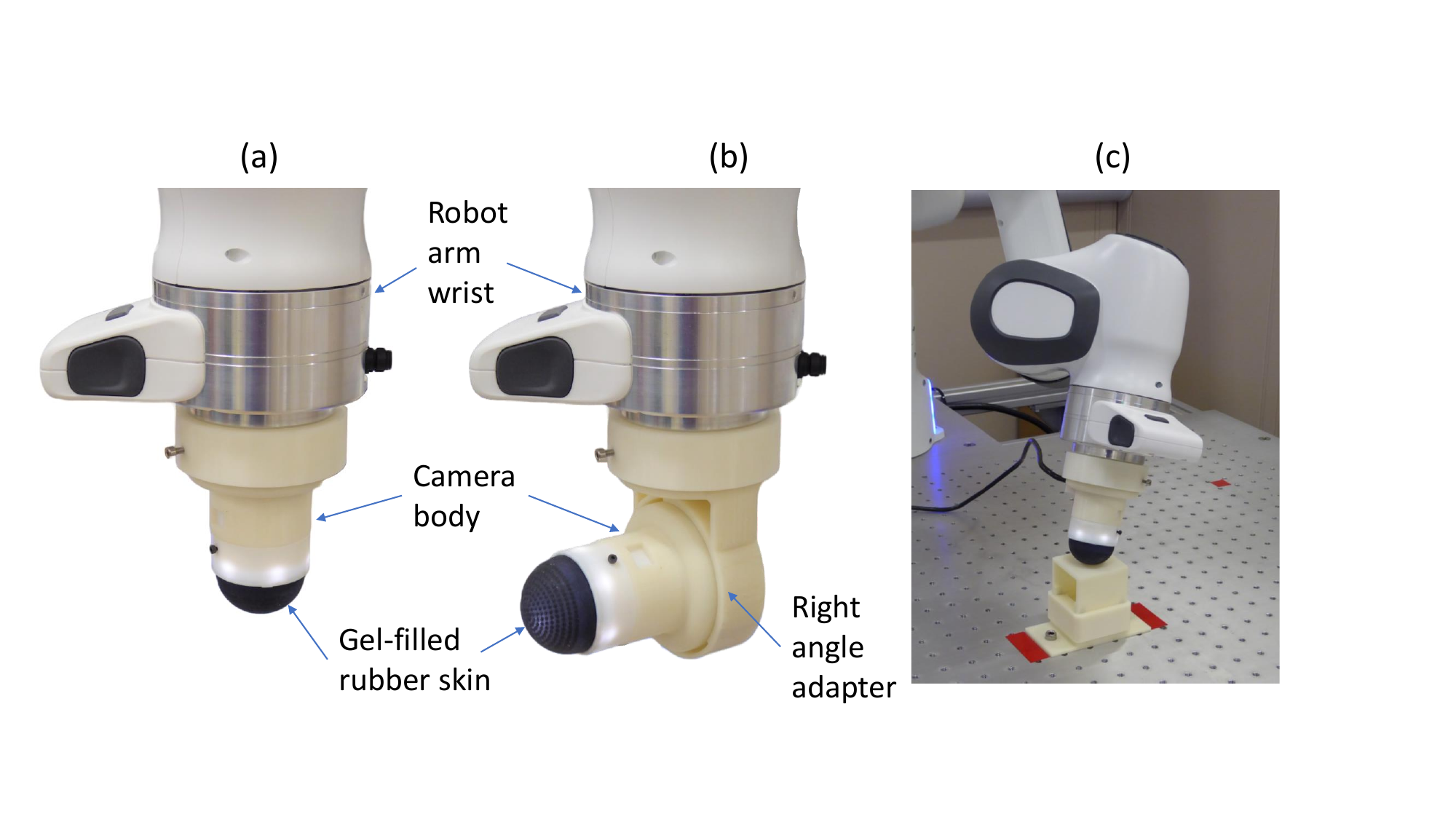}
	\caption{TacTip soft biomimetic optical tactile sensor, with (a)~sensor mount for a robot arm and (b) right-angled mount. (c)~Robot-arm mounted tactile sensor collecting training data.
	\label{fig:tactile_sensor}}
	\centering
	\includegraphics[width=\columnwidth,trim=20 120 20 70,clip]{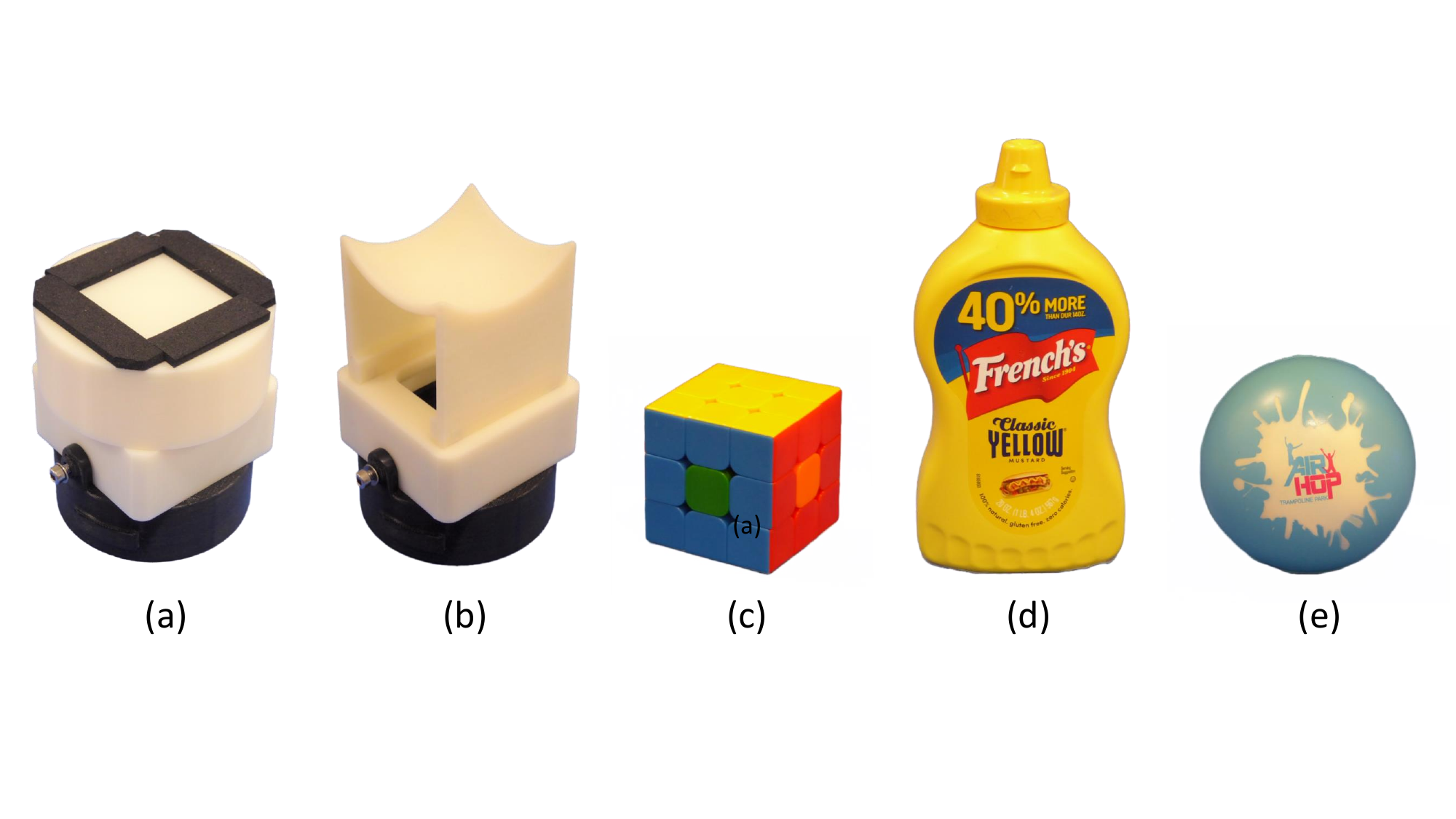}
	\caption{End-effector mounts used for object tracking: (a) Flat mount with non-slip foam pads (used for Rubik's cube and mustard bottle), (b)~Concave curved mount (used for foam ball). Objects: (c) Rubik's cube, (d) Mustard bottle, (e) Soft foam ball. {\color{black}Object (d) is from the YCB Object set~(\cite{calli2015benchmarking})}.
	\label{fig:end_effector_adaptor}}
\end{figure}

\subsubsection{Single-arm and dual-arm control configurations.}
\label{sec:single_dual_arm_controllers}

The tactile servoing and object pushing controllers described above can either be used in isolation to control a single robot arm for object tracking, surface following or single-arm pushing tasks, or they can be used in combination to control multiple robot arms. In the dual-arm pushing task, one arm is controlled by an active/leader pushing controller, while the second arm is controlled using a passive/follower object tracking controller. This dual-arm configuration allows the active pushing arm to control the movement of the object towards the target while the second passive arm helps to stabilise the object to prevent it from toppling.

Another way of viewing the operation of these multi-arm configurations is that each robot arm is attempting to follow a control signal via the feedforward path, while simultaneously trying to satisfy the constraints imposed by the reference contact pose specified in the feedback path. In this scenario, the feedforward control signals can either be generated separately for each arm in a decentralised approach or they can be generated in a centralised, more coordinated manner. The ``leader-follower" configuration we use in our dual-arm pushing task is an example of the decentralised approach.

\section{{\color{black}Tactile robot experimental platform}}
\label{sec:platform}

\begin{figure}[t]
	\centering
	\includegraphics[width=\columnwidth,trim=110 210 220 185,clip]{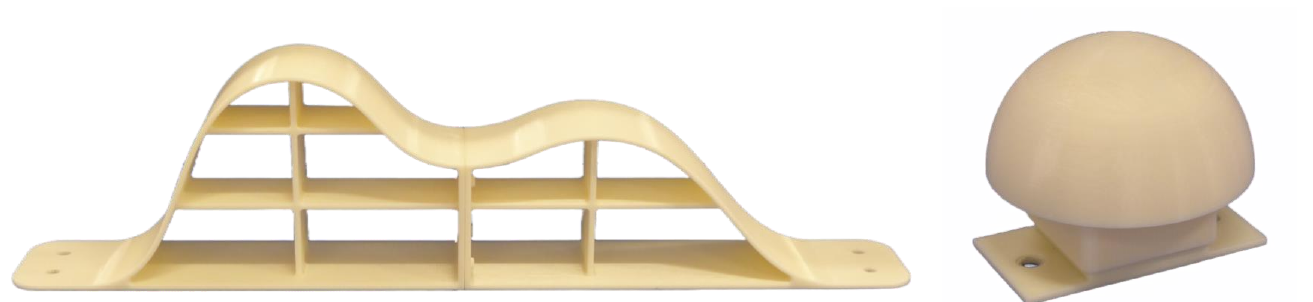}
	\caption{3D-printed objects used in surface following experiments: (left)~curved ramp, (right) hemispherical dome.
	\label{fig:surface_follow_objects}}
	\centering
	\includegraphics[width=\columnwidth,trim=170 160 170 170,clip]{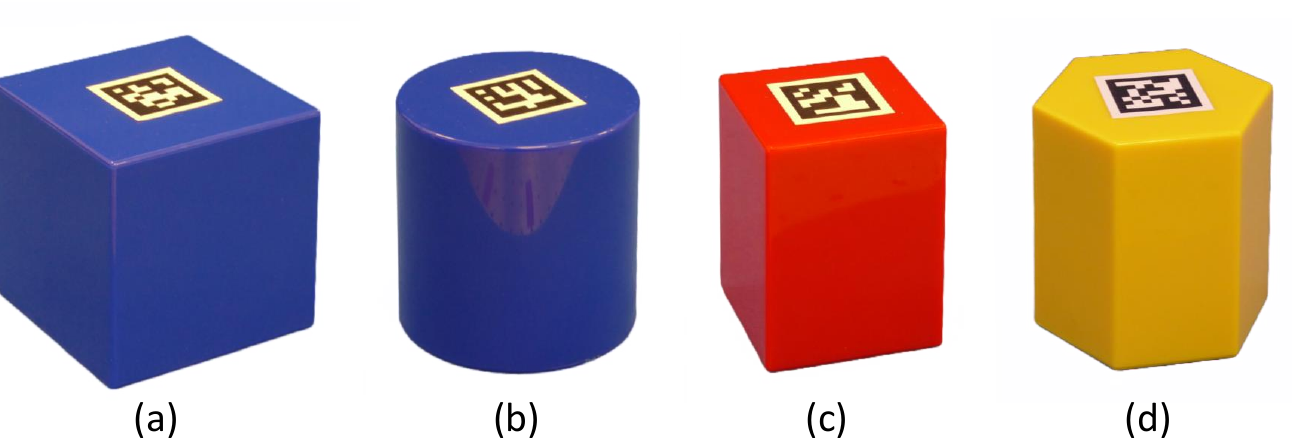}
	\caption{Regular geometric objects used in single-arm pushing experiments: (a) Large blue square prism (479\,g), (b)~Blue circular prism (363\,g), (c) Small red square prism (264\,g), (d) Yellow hexagonal prism (310\,g).
	\label{fig:single_arm_pushing_objects}}
	\centering
	\includegraphics[width=\columnwidth,trim=200 180 150 100,clip]{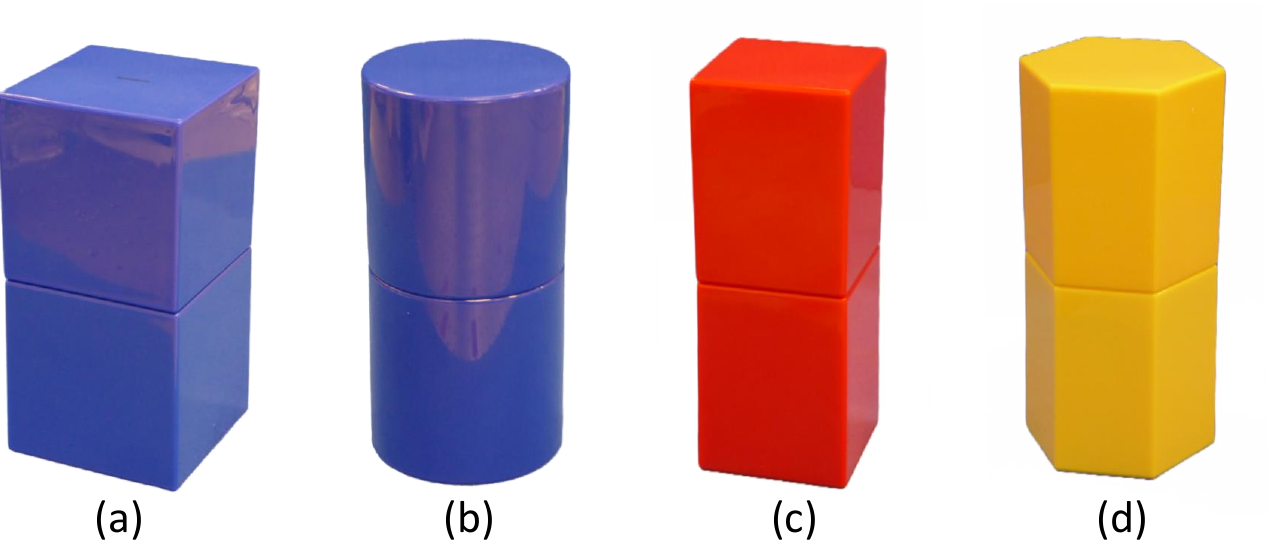}
	\includegraphics[width=\columnwidth,trim=160 125 130 125,clip]{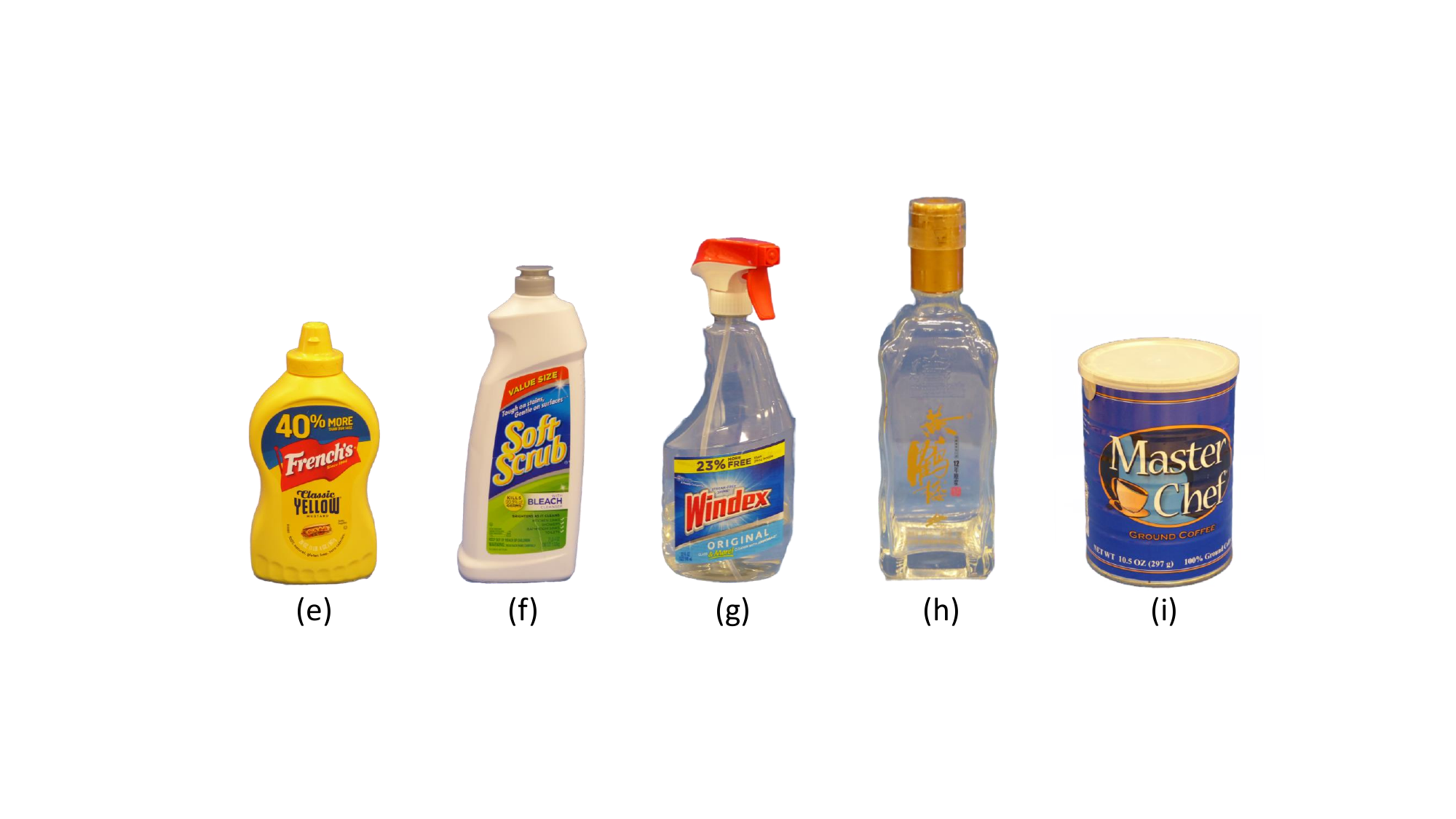}
	\caption{Tall (double height) geometric objects used in dual-arm object pushing experiments: (a) Large blue square prism (566\,g), (b) Blue circular prism (436\,g), (c) Small red square prism (324\,g), (d) Yellow hexagonal prism (373\,g). Tall everyday objects used in dual-arm object pushing experiments: (e) Mustard bottle (237\,g), (f) Cream cleaner bottle (161\,g), (g)~Windex window cleaner spray bottle (339\,g), (h) Glass bottle (641\,g), (i) Large coffee tin (214\,g). {\color{black}Objects (e),(f),(g) and (i) are from the YCB Object set~(\cite{calli2015benchmarking})}.
	\label{fig:tall_geometric_pushing_objects}}
\end{figure}

\subsection{Dual-arm robot platform}
\label{sec:dual_arm_robot_platform}

For our experiments and demonstrations, we use a dual robot arm system with two Franka Emika Panda, 7 degree-of-freedom (DoF) robot arms. The robot arms are mounted on custom aluminium trolleys with base plates, which are bolted together so that the arms are separated by 1.0\,m at their bases and can be used individually or together for collaborative tasks (Figure~\ref{fig:system_architecture}(a)-(d)). Depending on the task, the robots can either be fitted with a TacTip tactile sensor (Figure~\ref{fig:tactile_sensor}) or a stimulus adaptor as an end-effector (Figures~\ref{fig:end_effector_adaptor}(a)-(b)). The tactile sensor can be mounted in a standard downwards-pointing configuration or at a right angle using an adaptor mount (Figure~\ref{fig:tactile_sensor}(b)).

\subsection{Tactile sensor}
\label{sec:tactile_sensor}

The TacTip soft biomimetic optical tactile sensor (Figure~\ref{fig:tactile_sensor}) has been used in a wide variety of robotic touch applications and integrated into many robot hands (for reviews, see \cite{ward2018tactip, lepora2021soft}). The 3D-printed sensor tip consists of a black, gel-filled, rubber-like skin with an internal array of pins capped with white markers, which are imaged with a standard USB camera and LED lighting contained within the sensor body. The TacTip is considered biomimetic because these pins mimic the epidermal papillae structure in human skin on the boundary of the epidermal (outer) and dermal (inner) skin layers~(\cite{chorley2009development}), {\color{black}as verified in a comparison to real sensory neuronal data on matched stimuli~(\cite{pestell2022artificial}). Practically, the use of marker tips on pins means the sensor is highly sensitive to both normal contact and shear, because the pins act as levers that amplify small contacts into larger patterns of shear.

The TacTip is well-suited for investigating tactile control because its 3D-printed outer surface (Agilus 30, Stratesys) is fairly robust to abrasion and tears, while also being inexpensive and easy to replace. The soft inner gel (Techsil, Shore A hardness 15) gives a conformability similar to the soft parts of the human hand, making the sensor responsive and forgiving of errors in physical contact. Many variations of the TacTip have been created, from fingertip-sized sensors for anthropomorphic robot hands~(\cite{ford2023tactile}) to the DigiTac version of the low-cost DIGIT~(\cite{lepora2022digitac}).}

{\color{black}In this work, we use one or two TacTip sensors with 40\,mm diameter hemispherical tips containing} 331 marker-tipped pins arranged in a circular array. As in other work using the TacTip with deep learning, we use the raw sensor image with minimal pre-processing as input to a neural network model.

\begin{figure*}[t]
	\centering
	\includegraphics[width=\textwidth,trim=20 140 20 100,clip]{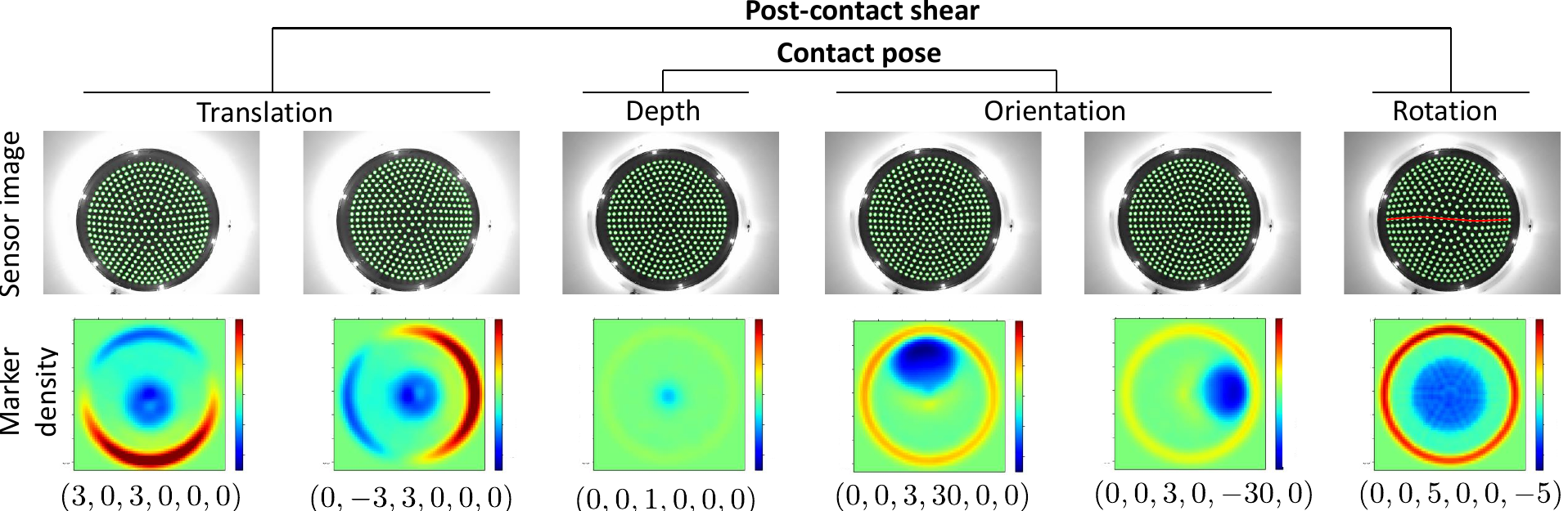}
	\caption{Visualization of tactile images as corresponding changes in marker density with respect to an undeformed tactile image across the relative surface contact poses and shears annotated below the marker density images. 
	\label{fig:marker_density_images}}
\end{figure*}

\subsection{Test objects}
\label{sec:test_objects}

Various test objects and mounts (Figures~\ref{fig:tactile_sensor}-\ref{fig:tall_geometric_pushing_objects}) were used in the experiments reported in the results sections of this paper. For training/validation/testing tactile data collection, the tactile sensor was mounted vertically on the end effector of the arm and brought into contact with a flat 3D-printed surface mounted to the base plate (Figure~\ref{fig:tactile_sensor}(c)). For the object-tracking experiments, end-effector mounts were used to attach flat or concave curved objects to the end of the leader arm (Figures~\ref{fig:end_effector_adaptor}(a)-(b)), using a similar flat surface to the one used to collect training data and a set of everyday objects (Rubik's cube, mustard bottle and soft foam ball) held against an adaptor by the tactile sensor mounted on the second robot arm (Figure~\ref{fig:end_effector_adaptor}(c)-(e)). For the surface following experiments, we used a 3D-printed curved ramp and hemisphere (Figure~\ref{fig:surface_follow_objects}) attached to mounts bolted onto the base plate. For the single-arm pushing experiments, we used four distinct plastic regular geometric objects (Figure~\ref{fig:single_arm_pushing_objects}). For the dual-arm pushing experiments, we used double-height (stacked) versions of the four geometric objects and five everyday objects (mustard bottle, cream cleaner bottle, window cleaner spray bottle, glass bottle and large coffee tin) as tall objects that are challenging because they usually topple when pushed~(Figure~\ref{fig:tall_geometric_pushing_objects}).

\subsection{Software infrastructure}
\label{sec:software}

We control the robot arms using a layered software API built on top of the \emph{libfranka} C++ library (version 0.8.0) of the Franka Control Interface (\cite{franka2015fci}). This library provides several software methods that allow users to specify callback functions within a 1\,kHz real-time control loop. On top of this, we have a developed an in-house library called \emph{pyfranka} that provides smooth trajectory generation for position- and velocity-based control, so that velocity, acceleration and jerk constraints are not violated; it also handles any background threads needed for velocity-based control and provides a python wrapper via \emph{pybind11}. The pyfranka library sits underneath the \emph{Common Robot Interface} (CRI) python library that has been used in most recent work on tactile sensing with the TacTip. Since the only critical functionality needed for our experiments and demonstrations is the ability to perform Cartesian position/velocity control and query the state of the robot, it should be possible to replace the Franka robot arms and API with any 6-DOF or 7-DOF robot arms that support this functionality.

The OpenCV library (version 4.5.2) is used to capture and process images from the tactile sensor, and TensorFlow (version 2.4) with the included Keras API to develop our neural network models for those tactile images. We also use the \emph{transforms3d} python library and the software provided with the book {\em Modern Robotics} (\cite{lynch2017modern}) to manipulate 3D poses, transforms and velocity twists.

We run all of the software components in a Pyro5 distributed object environment on an Ubuntu 18.04 desktop PC. The Pyro5 environment allows us to run several communicating python processes in parallel to ensure real-time performance. Using this approach, we were able to run the low-level 1\,kHz control loops, image capture, neural network inference (but not training) and high-level control loops for both robot arms and tactile sensors on a single PC.

\section{Experimental results}
\label{sec:experiments_demos}

\subsection{Pose-and-shear information in tactile images}
\label{sec:nn_pose_shear_information}

In the first experiment, we examine images from the tactile sensor used here (the TacTip) to check that the six considered components of contact pose and shear are represented in the tactile data. To do this, we visualize the marker densities of tactile images using a kernel density model (see \cite{silverman2018density}), with Gaussian kernels located at marker centroids and a constant kernel width (15 pixels) equal to the mean distance between adjacent markers.

From these visualizations, we were confident that the sensor images contained enough information to produce these estimates (Figure~\ref{fig:marker_density_images}). The size of the low-density blue region in the centre of the image depends on the contact depth, while its location in the image depends on the sensor orientation. Changes in marker density around the periphery of the sensor depend on the post-contact translational shear, and subtler changes within the contact region depend on the post-contact rotational shear. These are a type of feature that CNNs can easily replicate if required by applying a sequence of convolution and down-sampling operations. 

\subsection{Neural network-based pose-and-shear estimation}
\label{sec:nn_pose_shear_estimate_exp}

 \begin{table}[t]
	\small\sf\centering
	\caption{Overall MSE / mean NLL loss and pose component MAEs for 10 CNN {\color{black}regression and 10 GDN} models (mean values $\pm$ standard deviation across 10 models). The lowest mean MAE values for each component are highlighted in bold.
	\label{tab:nn_pose_shear_estimate_results}}
	\begin{tabular}{ccc}
	\toprule
	Metric & CNN {\color{black}regression} & GDN\\
	\midrule
	MSE/NLL loss & 1.43 $\pm$ 0.02 & -7.20 $\pm$ 0.25\\
	\midrule
	MAE $v_{x}$ (mm/s) & 0.448 $\pm$ 0.005 & \textbf{0.426} $\pm$ 0.005\\	
	MAE $v_{y}$ (mm/s) & 0.451 $\pm$ 0.004 & \textbf{0.422} $\pm$ 0.005\\	
	MAE $v_{z}$ (mm/s) & 0.164 $\pm$ 0.002 & \textbf{0.123} $\pm$ 0.002\\	
	MAE $\omega_{x}$ (deg/s) & 0.88 $\pm$ 0.03 & \textbf{0.45} $\pm$ 0.01\\	
	MAE $\omega_{y}$ (deg/s) & 1.04 $\pm$ 0.02 & \textbf{0.64} $\pm$ 0.01\\	
	MAE $\omega_{z}$ (deg/s) & 1.44 $\pm$ 0.02 & \textbf{1.16} $\pm$ 0.02\\
	\bottomrule
	\end{tabular}
\end{table}

In this experiment, we compare the performance of our GDN pose-and-shear estimation model against a baseline CNN model {\color{black}with regression head}. To ensure a fair comparison, both models were developed using the same three data sets (6000 training set samples, 2000 validation set samples and 2000 test set samples), which were collected as described in Section~\ref{sec:data_collection} and pre-processed as described in Section~\ref{sec:pre_post_processing}. We trained the CNN {\color{black}regression} model as described in Section~\ref{sec:cnn_model}, and the GDN models as described in Section~\ref{sec:gdn_model}. For statistical robustness, we trained 10 models of each from different random weight initializations and then computed the mean and standard deviation loss (MSE loss for the CNN {\color{black}regression} model and mean NLL for the GDN model) and component Mean Absolute Errors (MAEs) for all models on the test data set (Table~\ref{tab:nn_pose_shear_estimate_results}).

The results show that the GDN model produces lower component MAEs than the CNN {\color{black}regression} model when evaluated on the test data set. An explanation for this is that the mean NLL loss function used to train the GDN model directly has a variable, estimated uncertainty for each pose component, which in effect increases the error weighting on more confident estimates and decreases the error weighting on less confident ones (see Section~\ref{sec:gdn_model}). This contrasts with the MSE loss function used to train the CNN {\color{black}regression} model, which implicitly assumes a constant, pre-specified uncertainty for each pose component and hence is unable to incorporate variations in the estimated uncertainty.

\begin{figure*}
	\centering
	\includegraphics[width=\textwidth,trim=5 100 5 115,clip]{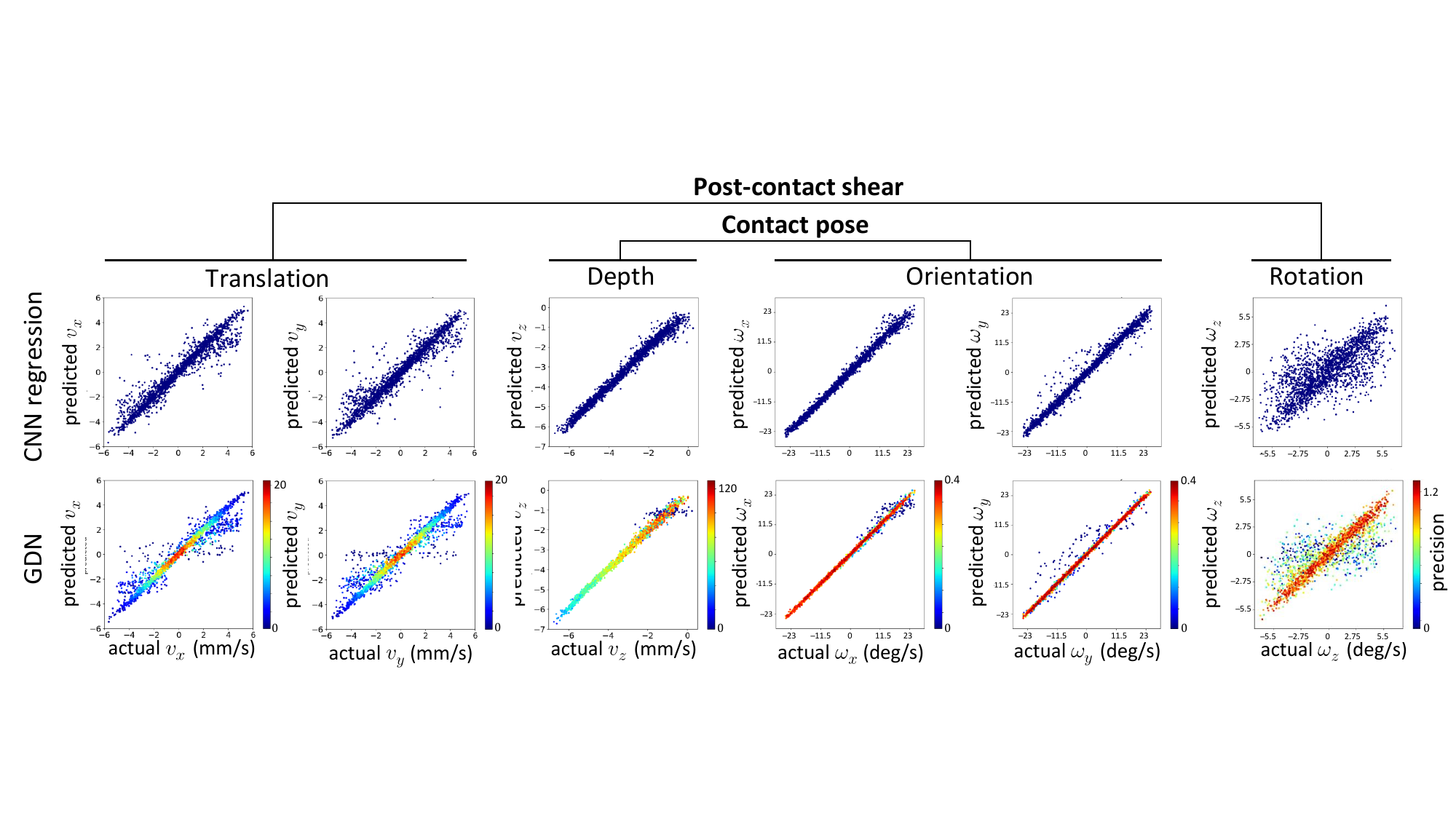}
	\caption{Distribution of test errors for best-performing (lowest loss) CNN {\color{black}regression} and GDN models. Predicted pose-and-shear values are plotted against their actual values. GDN estimates are coloured by their predicted precision (reciprocal variance).
	\label{fig:nn_pose_shear_estimate_performance}}
	\centering
	\includegraphics[width=\textwidth,trim=5 260 5 275,clip]{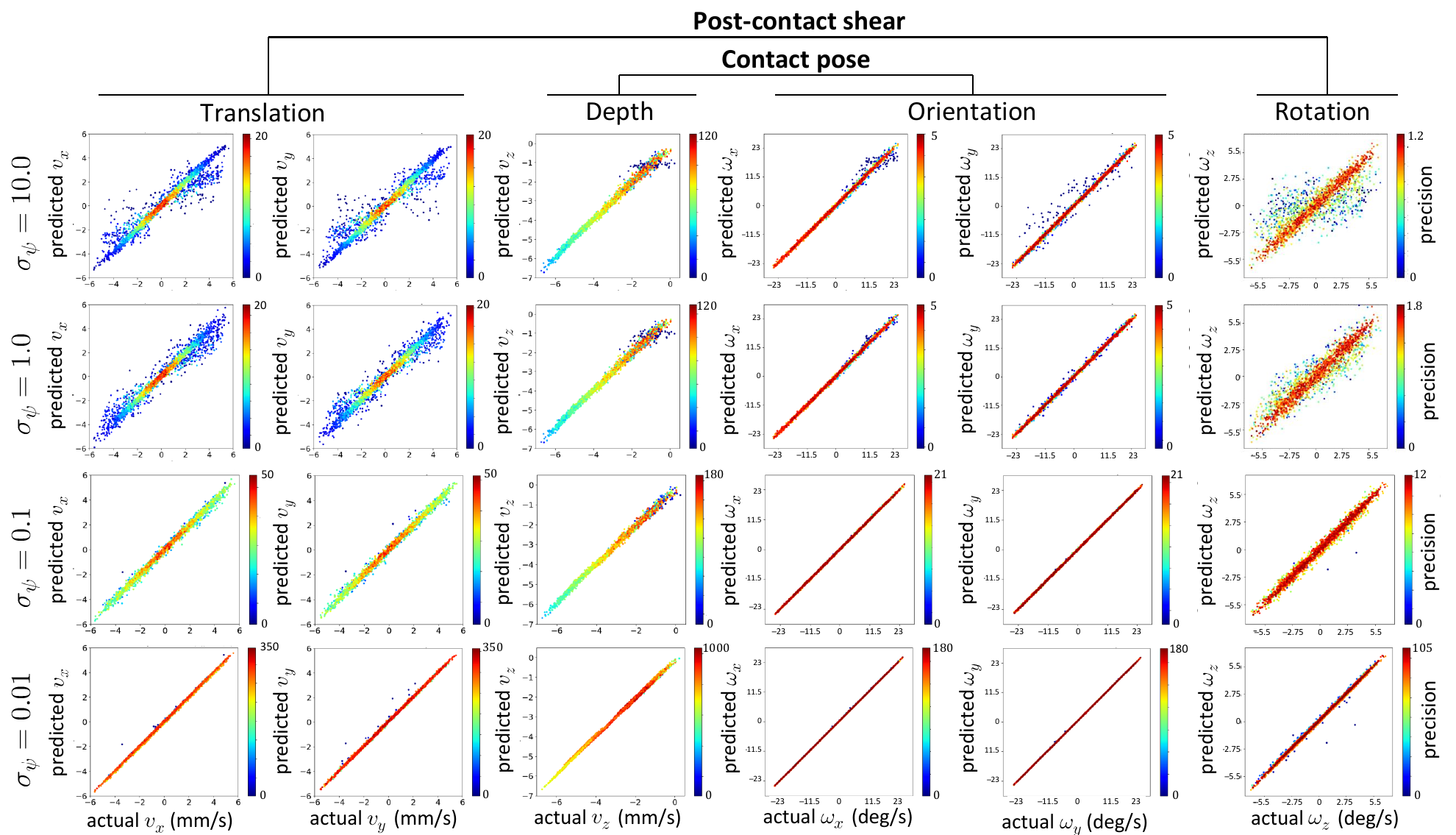}
	\caption{Distribution of test errors for best-performing (lowest loss) GDN model followed by Bayesian filter with different state dynamics noise levels $\sigma_{\psi}$. Pose-and-shear values (predicted vs actual) are coloured by their predicted precision.
	\label{fig:bayes_filter_performance}}
\end{figure*}

We visualise the distribution of test set errors by plotting the estimated pose components against the ground truth pose components for the best-performing model of each type (Figure~\ref{fig:nn_pose_shear_estimate_performance}). For the GDN model, we also colour each point according to the precision (inverse variance) estimated by the model for the corresponding pose component. Points coloured in red denote high precision (low uncertainty) estimates and points coloured in blue denote low precision (high uncertainty) estimates. 

\begin{table*}[t!]
	\small\sf\centering
	\caption{Pose component MAEs for 10 GDN models followed by Bayesian filter with different state dynamics noise levels, $\sigma_{\psi}$ (mean values $\pm$ standard deviation across 10 models). The lowest mean MAE values are highlighted in bold.
	\label{tab:bayes_filter_results}}
	\begin{tabular}{cccccc}
	\toprule
	Metric & $\sigma_{\psi} = \infty$ & $\sigma_{\psi} = 10.0$ & $\sigma_{\psi} = 1.0$ & $\sigma_{\psi} = 0.1$ & $\sigma_{\psi} = 0.01$\\
	\midrule
	MAE $v_{x}$ (mm/s) & 0.426 $\pm$ 0.005 & 0.422 $\pm$ 0.006 & 0.360 $\pm$ 0.007 & 0.160 $\pm$ 0.004 & \textbf{0.062} $\pm$ 0.004\\	
	MAE $v_{y}$ (mm/s) & 0.422 $\pm$ 0.005 & 0.421 $\pm$ 0.006 & 0.352 $\pm$ 0.006 & 0.161 $\pm$ 0.005 & \textbf{0.065} $\pm$ 0.005\\	
	MAE $v_{z}$ (mm/s) & 0.123 $\pm$ 0.002 & 0.123 $\pm$ 0.002 & 0.121 $\pm$ 0.002 & 0.098 $\pm$ 0.001 & \textbf{0.069} $\pm$ 0.003\\	
	MAE $\omega_{x}$ (deg/s) & 0.50 $\pm$ 0.01 & 0.49 $\pm$ 0.01 & 0.40 $\pm$ 0.01 & 0.18 $\pm$ 0.01 & \textbf{0.08} $\pm$ 0.01\\	
	MAE $\omega_{y}$ (deg/s) & 0.64 $\pm$ 0.01 & 0. 62 $\pm$ 0.01 & 0.42 $\pm$ 0.02 & 0.18 $\pm$ 0.01 & \textbf{0.07} $\pm$ 0.01\\
    MAE $\omega_{z}$ (deg/s) & 1.16 $\pm$ 0.01 & 1.15 $\pm$ 0.02 & 0.82 $\pm$ 0.02 & 0.30 $\pm$ 0.02 & \textbf{0.11} $\pm$ 0.02\\
 \bottomrule
	\end{tabular}
\end{table*}

With reference to these plots, we make the following observations. Firstly, while the errors are significant across all pose components estimated by both models, they are larger for the shear-related components than the normal contact ones. This could be due to aliasing effects, which are more prevalent during shear motion than normal contact motion; for example, at small contact depths, the tactile sensor is prone to slip under translational shear, which would lead to a similar tactile image for a range of shear values (see~\cite{lloyd-RSS-21} for an explanation of the effects of tactile aliasing). Secondly, the GDN estimates appear more accurate than the CNN {\color{black}regression} estimates, in that the distribution of predicted values is more concentrated around the true values for the GDN model than the CNN {\color{black}regression} model, which is consistent with the statistical results presented in Table~\ref{tab:nn_pose_shear_estimate_results}. Finally, the precision (uncertainty) values estimated by the GDN model appear to correlate with the errors, in the sense that the red points tend to lie closer to the imaginary ground-truth line than the blue points. In the following section, we consider the impact that our $SE(3)$ Bayesian filter has on reducing these estimation errors and the associated uncertainty.

\subsection{Error and uncertainty reduction using an SE(3) Bayesian filter}
\label{sec:se3_bayes_filter_exp}

To evaluate the effect of the $SE(3)$ discriminative Bayesian filter on the pose-and-shear predictions and uncertainty values produced by the GDN model, we treated the test data set as a \emph{sequence} of sensor images with corresponding pose-and-shear estimates on consecutive random contacts. 

Since the pose changes between consecutive sensor contacts can be computed from the test data labels, we can compute the state dynamics transformation $\boldsymbol{\mathrm{T}}_k$ whose mean is computed in Algorithm~\ref{alg:se3_bayes_filter} for time step $k$:
\begin{equation}
	\label{eqn:se3_filter_exp_state_trans_model}
	\boldsymbol{\mathrm{T}}_{k}
	\, = \,
	\exp \left( \boldsymbol\psi_{k}^{\wedge} \right)
	\boldsymbol{\mathrm{X}}_{k} \boldsymbol{\mathrm{X}}_{k-1}^{-1}.
\end{equation}
Here, $\boldsymbol{\mathrm{X}}_{k}$ and $\boldsymbol{\mathrm{X}}_{k-1}$ are the contact poses ({\em i.e.} the test data labels) at time steps $k$ and $k-1$, and $\boldsymbol{\psi}_{k} \sim \mathcal{N} \left(\boldsymbol{\mathrm{0}}, \boldsymbol\Sigma_{\boldsymbol\psi} \right)$ is a Gaussian noise perturbation applied at time step $k$, which represents simulated noise in the state dynamics model. We specify the noise perturbation covariance as:
\begin{equation}
	\label{eqn:se3_filter_exp_state_trans_cov}
	\boldsymbol\Sigma_{\boldsymbol\psi}
	=
	\sigma_{\psi}^{2}\,\,\boldsymbol{1}_{6\times 6}.
\end{equation}

The discriminative Bayesian filter (Algorithm~\ref{alg:se3_bayes_filter}) was then applied to the GDN pose estimates generated in response to the sequence of test inputs, using Equation~\ref{eqn:se3_filter_exp_state_trans_model} to compute the $SE(3)$ transformation in the state dynamics model at each time step. We set the corresponding perturbation noise covariance $\boldsymbol{\Sigma}_{\boldsymbol{\phi}}$ in Algorithm~\ref{alg:se3_bayes_filter} equal to the perturbation noise covariance defined in Equation~\ref{eqn:se3_filter_exp_state_trans_cov}: $\boldsymbol{\Sigma}_{\boldsymbol{\phi}} = \boldsymbol\Sigma_{\boldsymbol\psi}$.

As in the single-prediction results, we improved statistical robustness by applying the Bayesian filter to each of the 10 GDN models we trained from different random weight initializations and evaluated the mean and standard deviation component MAEs for all models on the test data set. We repeated the experiment for four different noise levels $\sigma_{\psi}$, which are specified on a logarithmic scale between minimum and maximum values $\sigma_{\psi}=0.01$ and $\sigma_{\psi}=10.0$. 

The statistical results presented in Table~\ref{tab:bayes_filter_results} show that the filtered estimates become more accurate as the noise levels in the real state dynamics and the state dynamics model are reduced. As the noise levels increase, the accuracy reduces to the single time-step prediction results for the GDN models in Table~\ref{tab:nn_pose_shear_estimate_results}, which we consider as the $\sigma_{\psi}=\infty$ case and have included for comparison in Table~\ref{tab:bayes_filter_results}.

To show explicitly how the GDN model depends on the state dynamics noise, we visualise the distribution of test sequence errors by plotting the filtered pose-and-shear predictions against the actual components for the best-performing GDN model at the different noise levels (Figure~\ref{fig:bayes_filter_performance}). With reference to these plots, we make the following observations. Firstly, the accuracy of the filtered estimates increases as the state dynamics noise level $\sigma_{\psi}$ is decreased (the magnitude of errors about the imaginary ground-truth line decreases). Secondly, the filtered uncertainty estimates get smaller as the state dynamics noise level $\sigma_{\psi}$ is decreased (the proportion of points coloured red and blue increases and decreases respectively). Both of these observations are a consequence of the state dynamics model becoming more accurate as the noise levels are reduced. This allows more effective combination of consecutive pose estimates, which increases their accuracy and reduces the associated uncertainty.

{\color{black}In the above analysis, the state dynamics noise level $\sigma_\psi$ is known, so the noise covariance in the Bayesian filter update can be set to that value, $\sigma_\phi=\sigma_\psi$. In the following experiments, the noise covariance in the state dynamics model is set to a constant $\sigma_\phi=0.5$ (see text following Equation~\ref{eqn:state_dynamics_noise_cov}), which allows for precise control while being able to react quickly to changes in the environment.}   

\subsection{Task 1: Object pose tracking}
\label{sec:obj_pose_tracking_exp}

In this experiment, we show how our tactile robotic system can be configured to track the pose of a moving object. We demonstrate this capability using two robot arms: the first arm (the \emph{leader robot}) moves an object around in 3D space, while a second arm fitted with a tactile sensor (the \emph{follower robot}) tracks the motion of the object using the tactile servoing controller described in Section~\ref{sec:servoing_controller}.

There are two parts to this experiment. In the first part, we show that the follower arm can track changes to individual pose components of a moving object. More specifically, we track translational motion along the $x$, $y$ and $z$ axes of the robot work frame, and $\alpha$, $\beta$ and $\gamma$ rotational motion around these axes. In the second part of the experiment, we show that the follower arm can track simultaneous changes to all pose components while the leader arm moves the object in a complex periodic motion. Another key difference between these experiments is that in the first part we only track a flat surface attached to the end of the leader arm, whereas in the second part we also track several everyday objects that are held in position against the leader arm by the follower arm (Figure~\ref{fig:end_effector_adaptor}). Hence, the second part of the experiment also demonstrates a form of dual-arm object manipulation.

For both parts of this experiment, we used the controller parameters listed in Table~\ref{tab:object_tracking_control_params} (Appendix~\ref{sec:controller_parameters}) in the tactile servoing controller (Figure~\ref{fig:pushing_controller}(a), top controller only). The feedback reference pose specifies that the tactile sensor should be orientated normal to the contacted surface at a contact depth of 6\,mm. Since the feedforward velocity is not required for object tracking tasks, it is set to zero.

\subsubsection{Tracking changes to single pose components.}
\label{sec:object_pose_tracking_single_exp}

\begin{figure*}
	\centering
	\includegraphics[width=\textwidth,trim=0 270 0 200,clip]{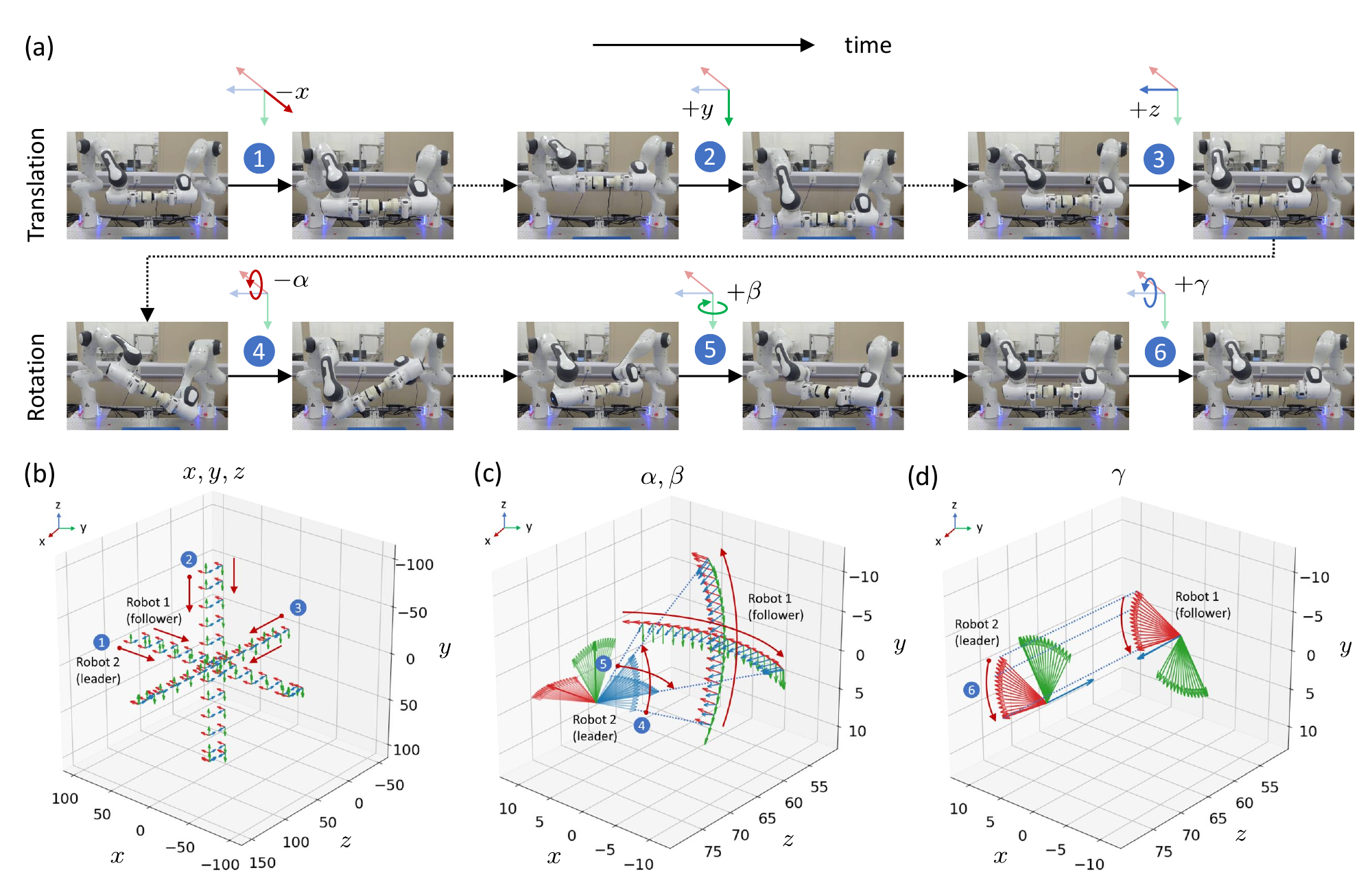}
	\caption{Using the follower arm to track changes to individual components of the leader arm pose. (a) Tracking sequence: translation along $-x \rightarrow y \rightarrow z$ axes (1-3), followed by $-\alpha \rightarrow \beta \rightarrow \gamma$ rotation around these axes (4-6). (b) Leader and follower arm pose trajectory as leader arm translates along $-x$, $y$ and $z$ axes. (c) Leader and follower arm pose trajectory as leader arm rotates about $x$ and $y$ axes ($\alpha$ and $\beta$). (d) Leader and follower arm pose trajectory as leader arm rotates about $z$ axis ($\gamma$). 
	\label{fig:object_tracking_single}}
	\centering
	\includegraphics[width=\textwidth,trim=0 270 0 180,clip]{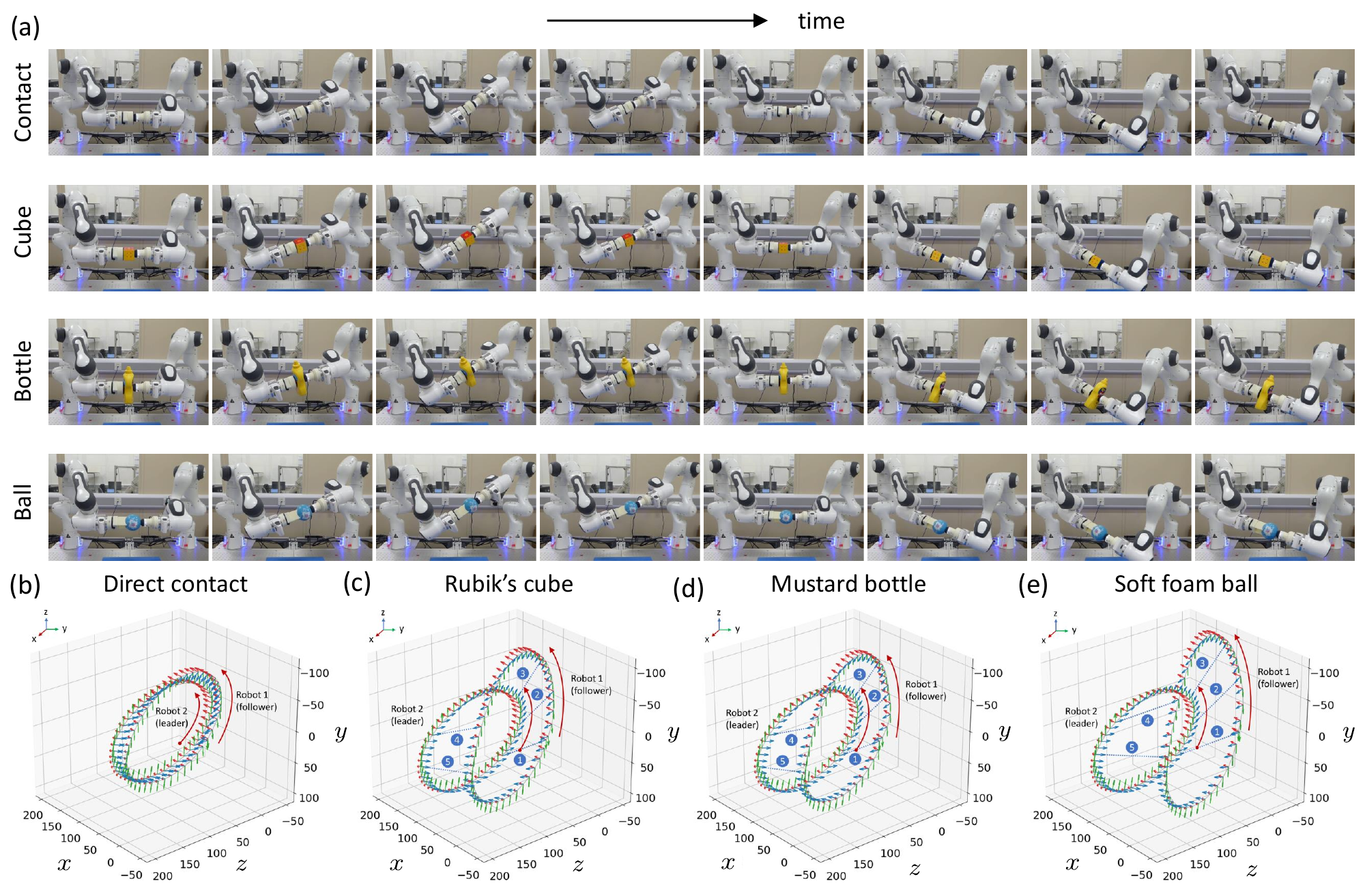}
	\caption{Using the follower arm to track simultaneous changes to all components of the leader arm pose. (a) Tracking sequences for direct (flat) surface contact, Rubik's cube, mustard bottle and soft foam ball. Leader and follower robot pose trajectory for a single period of the leader arm trajectory for, (b) direct surface contact, (c) Rubik's cube, (d) mustard bottle, and (e) soft foam ball. In (c)-(e), the numbered points indicate corresponding poses of the two robot arms at several points in the pose trajectory .
	\label{fig:object_tracking_multi}}
\end{figure*}

In the first part of the experiment, we initially positioned the follower arm tactile sensor in direct contact with the leader arm flat surface at a contact depth of approximately 6\,mm and so that its central axis was normal to the flat surface. We then used the leader robot to move the flat surface through a sequence of 200\,mm translations along the $-x$, $y$ and $z$ axes (of the robot work frame), and then through 60 degree $\alpha$, $\beta$ and $\gamma$ rotations about these axes (Figure~\ref{fig:object_tracking_single}(a)). During the tracking sequence, we recorded the end-effector poses and corresponding time stamps for both robot arms at the start of each control cycle. This allowed us to (approximately) match up the corresponding poses for the two arms and plot them in 3D for different points in the trajectory after the experiment had finished (Figure~\ref{fig:object_tracking_single}(b)-(d)).

For plots that relate to translational pose components (Figure~\ref{fig:object_tracking_single}(b))), we removed variation in the rotational components from the response of the follower arm. Similarly, for plots that relate to rotational pose components (Figures~\ref{fig:object_tracking_single}(c)-(d)), we removed any variation in the translational components from the response of the follower arm. This allows us to focus on individual pose components when evaluating how the follower arm responds to changes in leader arm pose. If we did not do this, but instead plotted the raw unaltered poses, it would make it extremely difficult to compare individual pose components of the leader and follower arms at any point in time, particularly for the rotational components ($\alpha$, $\beta$ and $\gamma$). That said, in the second part of the experiment below where we vary all components of the pose together, we will be able to plot the raw unaltered poses for both arms to see the correspondence.

The pose trajectory plots (Figures~\ref{fig:object_tracking_single}(b)-(d)) show that the tactile sensor on the follower arm tracks changes to individual pose components of the flat surface on the end effector of the leader arm. The coordinates are for the tool centre point of each robot, which on the leader arm is in the centre of the flat surface and on follower arm is in the centre of the sensor tip. For translations along $-x$, $y$ and $z$, the follower pose is displaced by $z\approx15$\,mm from the coordinate frame of the leader end effector, where $z$ points towards the follower end effector and the follower coordinate frame of the sensor is reversed compared to that of the leader (Figure~\ref{fig:object_tracking_single}(b)). This displacement is consistent with the 6\,mm contact depth and sensor tip radius of 40\,mm. Likewise for rotations $\alpha$ and $\beta$ around the $x$ and $y$ axes, the follower coordinate frame of the sensor tracks along circular arcs of radius $\sim$15\,mm around a point centred on the leader, with the leader coordinate frame reversed and pivoting around that same point (Figure~\ref{fig:object_tracking_single}(c)). For rotations $\gamma$ around the $z$-axis, the follower again tracks the leader with coordinate frame displaced by $\sim$15\,mm and reversed in $z$ (Figure~\ref{fig:object_tracking_single}(d)). 

\subsubsection{Tracking simultaneous changes to all pose components.}
\label{sec:object_pose_tracking_multi_exp}

In the second part of the experiment, we moved the leader robot arm in a more complex velocity trajectory $\boldsymbol{\mathrm{v}}(t)$ where all of the pose components were varied at the same time using the periodic function:
\begin{equation}
	{\mathrm{v}_j}(t) = \frac{2 \pi {\mathrm{b}_j}}{T}\, \cos \left(\frac{2 \pi t}{T} + \phi_j \right), 
\end{equation}
where we set an amplitude $\boldsymbol{\mathrm{b}}=\left[ 75, 75, 75, 25, 25, 25 \right]^{\top}$, phase $\boldsymbol\phi=\left[ \frac{\pi}{2}, 0, 0, 0, 0, 0 \right]^{\top}$ and period $T=30$\,sec. The translational and rotational components of the amplitude $\boldsymbol{\mathrm{b}}$ have units of mm and degrees respectively. The velocity trajectory was tracked over three full periods ({\em i.e.} 90\,sec).

In addition to tracking a flat surface attached to the leader arm, as in the first part of the experiment above, in this second part we also tracked several everyday objects (Rubik's cube, mustard bottle and soft foam ball; Figure~\ref{fig:end_effector_adaptor}) held between the two arms as they followed the leader arm trajectory. Again, we recorded the end-effector poses and corresponding time stamps for both robot arms at the start of each control cycle to match up the corresponding poses from both robots and plot them after the experiment had finished.

The time-lapse photos and pose trajectory plots (Figure~\ref{fig:object_tracking_multi}) show that the follower arm tracks simultaneous changes to all components of the leader arm pose as the leader arm follows a complex periodic trajectory. Moreover, it can also hold an object against the leader arm while it is following its trajectory, thereby implementing a form of 3D object manipulation guided by the leader arm. 

The leader pose trajectory is the same for all objects and forms {\color{black}an approx.} 150\,mm diameter circle with the $z$-axis of the coordinate frame (blue axis) rotating from orthogonal to tangential to the plane containing the circle at the bottom and top antipodal points (Figure~\ref{fig:object_tracking_multi})(b)-(d)). This $z$-axis points towards the position of the follower arm, which has its coordinate frame reversed compared to the leader arm. For direct contact, the two arm end-effector positions are {\color{black}about} 15\,mm apart along the $z$-axis, resulting in a slightly larger, tilted pose-trajectory for the follower compared to that of the leader. As the object size increases, the pose-trajectory of the follower becomes larger relative to the leader but is still separated along the same orientation of the $z$-axis, increasing from direct contact ($\sim15$\,mm) to the Rubik's cube and mustard bottle (both $\sim 60$\,mm) to the ball ($\sim 80$\,mm).

\subsection{Task 2: Surface following}
\label{sec:surface_follow_exp}

In this experiment, we show how our tactile robotic system can be configured for surface following tasks for two scenarios: traversing a straight line projection on the surface of a curved ramp, and traversing a sequence of eight straight line projections outwards from the centre of a hemispherical dome at 45 degree intervals. The surfaces used in these two scenarios are shown in Figure~\ref{fig:surface_follow_objects}.

For both parts of this experiment, we use the tactile servoing controller described in Section~\ref{sec:servoing_controller} with the controller parameters listed in Table~\ref{tab:surface_follow_control_params} of Appendix~\ref{sec:controller_parameters}. The feedback reference pose specifies that the sensor should be orientated normal to the contacted surface at a contact depth of 3\,mm. Since, the feedforward velocity depends on the particular surface following task being performed, it is specified in each of the following subsections that describe each task.

\subsubsection{Surface following on a curved ramp.}
\label{sec:surface_follow_ramp}

For this first surface following task, we initially positioned the robot arm so that the tactile sensor made contact with the highest part of the curved ramp at a contact depth of 3\,mm with the $y$-axis of the sensor aligned with the $y$-axis of the robot work frame pointing along the length of the ramp. We set the feedforward velocity to 10\,mm/s, with $\boldsymbol{\mathrm{u}}_{s^{\prime}2}=(0,10,0,0,0,0)$ in the tactile servoing controller (Figure~\ref{fig:pushing_controller}(a), top controller only). During surface following, we recorded the end-effector poses and associated time stamps during each control cycle. 

\begin{figure*}
	\centering
	\includegraphics[width=\textwidth,trim=25 421 20 223,clip]{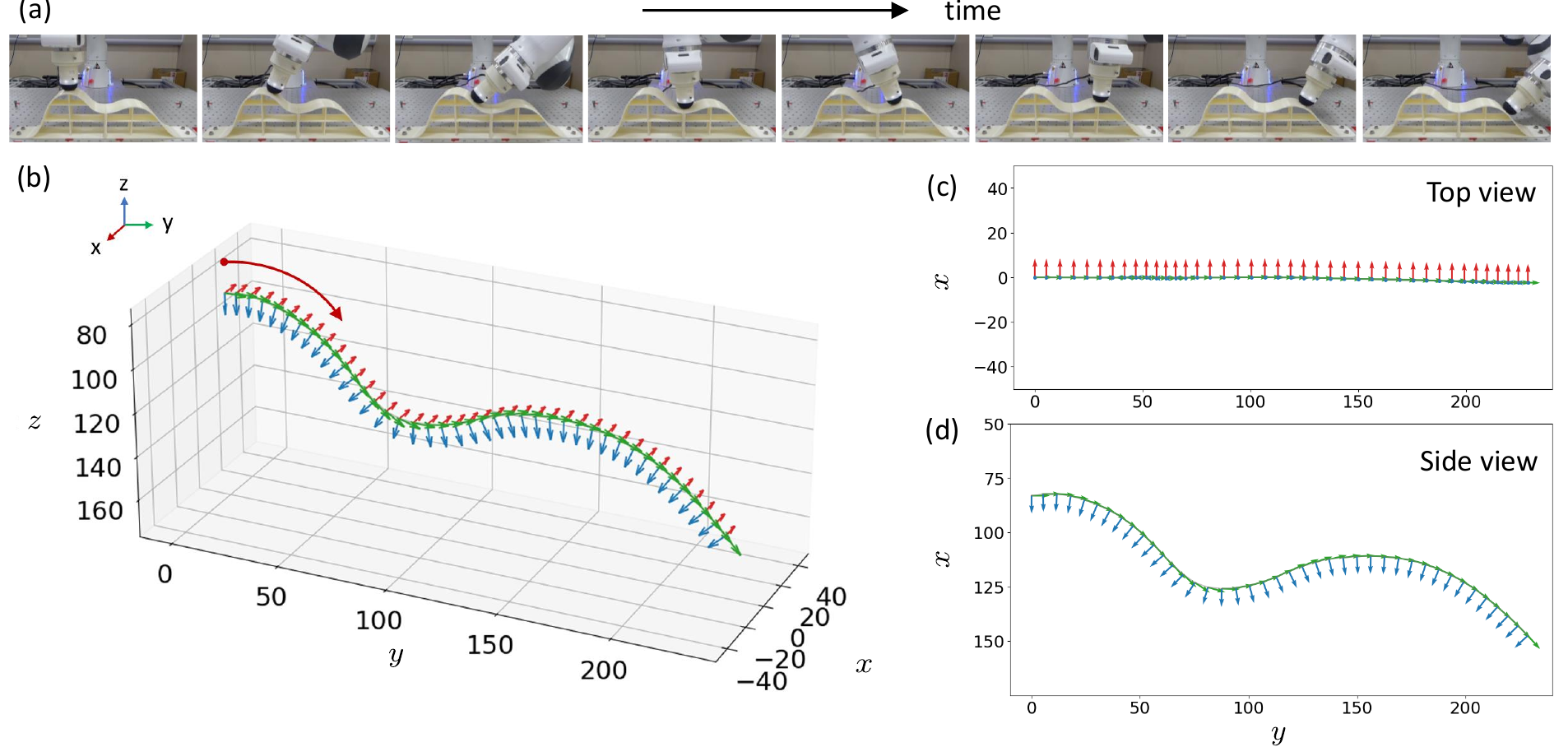}
	\caption{Tactile servoing to follow a curved ramp surface. (a) Time-lapse photos. (b)-(d) Robot arm end-effector pose trajectory. {\color{black}The red/green/blue arrows in this and the other figures correspond to about one every second, although precise timings vary due to factors such as individual controller details and the real-time processing requirements.}
	\label{fig:surface_follow_ramp}}
	\centering
 \vspace{1em}
	\includegraphics[width=\textwidth,trim=75 345 25 220,clip]{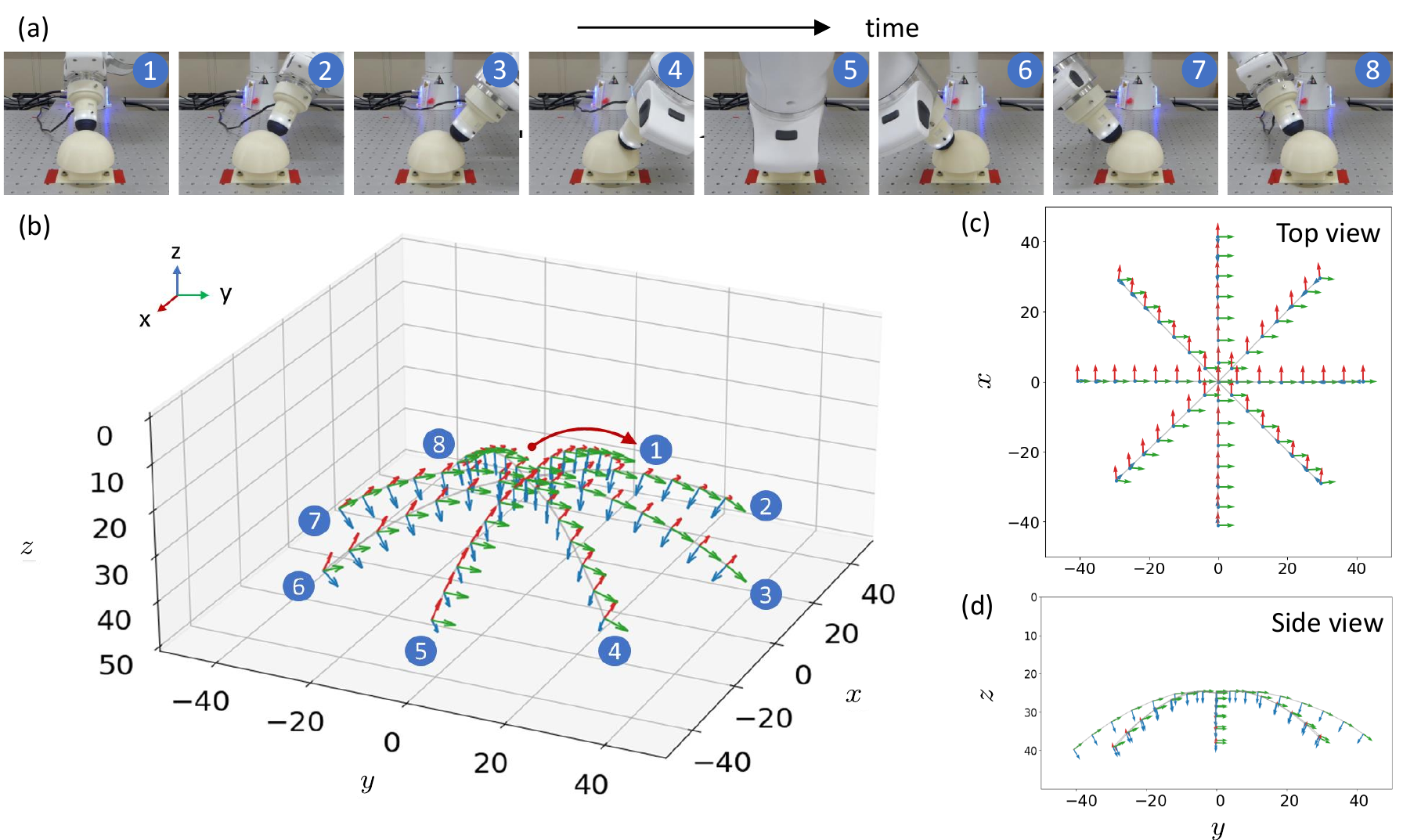}
	\caption{Tactile servoing to follow radial paths from the centre of the surface of a hemispherical dome. (a) Time-lapse photos. (b)-(d)~Robot arm end-effector pose trajectories. In (b), the numbered points correspond to different radial paths over the surface.
	\label{fig:surface_follow_hemisphere}}
\end{figure*}

For this surface-following task, the time-lapse photos and pose trajectory plots show that the robot arm successfully follows this type of gently curving surface while the sensor remains in contact with it and orientated normal to the surface (Figure~\ref{fig:surface_follow_ramp}). The pose-trajectory is almost straight along the $x$-axis of the work frame, but has a small drift, presumably because of a slight surface tilt in the $x$ direction. 

\subsubsection{Surface following on a hemispherical dome.}
\label{sec:surface_follow_hemisphere}

For this second surface-following task, we initially positioned the robot arm so that the tactile sensor made contact with the centre of the dome at a contact depth of 3\,mm, with the sensor $y$-axis aligned with the $y$-axis of the robot work frame. 

When following the $i$th radial path from the centre of the hemisphere, at angle $\theta_{i} = 0, 45^\circ, 90^\circ, \ldots, 315^\circ$, we set the feedforward velocity $\boldsymbol{\mathrm{u}}_{s^{\prime}2}=\left( 10 \cos \theta_{i}, 10 \sin \theta_{i}, 0, 0, 0, 0\right)$, over 8 radial paths ($1\leq i\leq 8$). Specifying the reference pose and feedforward velocity in this way causes the sensor to move at 10\,mm/s tangentially to the surface in direction $\theta_{i}$ while remaining at a contact depth of 3 mm. During the surface following sequence, we recorded the end-effector poses and corresponding time stamps as the sensor moved along each of the radial paths.

For this surface following task, the time-lapse photos and pose trajectory plots show that the robot arm end effector successfully follows this curved surface while the sensor remains in contact and orientated normal to the surface (Figure~\ref{fig:surface_follow_hemisphere}).

\subsection{Task 3: Single-arm object pushing}
\label{sec:single_arm_object_pushing_exp}

In our first object pushing experiment, we demonstrate how our tactile robotic system can be used for single-arm object pushing tasks, similar to those demonstrated in earlier work (\cite{lloyd2021goal}). A major improvement on that earlier work is that the present system can push an object in a smooth continuous manner rather than the previous discrete point motion, because we now use velocity control rather than the position control used previously. We also now show that our new system can push objects over surfaces with different frictional properties, considering both medium-density fibreboard (MDF) and a soft foam surface.

For the single-arm pushing configuration, we mounted the tactile sensor as an end effector of the robot arm using a right-angle adapter (Figure~\ref{fig:system_architecture}(c)) so that it can be moved parallel to the surface during the pushing sequence without the arm getting caught on the surface. At the start of each trial, we positioned the tactile sensor end-effector 45\,mm above the surface with central axis parallel to the $y$-axis of the robot work frame at position $(y,z)=(-250,100)$\,mm in the $yz$-plane parallel to the surface. We then placed the object centrally in front of the tactile sensor so that the contacted surface of the object was about normal to the sensor axis.

For this experiment, we pushed several regular geometric objects (Figure~\ref{fig:single_arm_pushing_objects}) across the MDF and foam surface. These objects were also used in previous work (\cite{lloyd2021goal}), except we do not use the triangular prism because it cannot be used for dual-arm object pushing below with both arms contacting a flat surface.

For each trial of the experiment, the robot arm was used to push the object towards the target at position $(y,z)=(0,375)$\,mm while remaining in contact with the object. The location of the target relative to the object's starting pose means that the robot has to push the object around a bend to reach the target. To control the robot arm, we used the pushing controller described in Section~\ref{sec:pushing_controller}, with the parameters listed in Table~\ref{tab:pushing_control_params} of Appendix~\ref{sec:controller_parameters}.

During each trial, we recorded the end-effector poses and corresponding time stamps over each control cycle. To improve statistical robustness, we repeated the trial five times for each object and computed the mean $\pm$ standard deviation final target error across all five trials (Table~\ref{tab:single_arm_pushing_results}). As in \cite{lloyd2021goal}, we define the final target error as the perpendicular distance from the target to the sensor-object contact normal on completion of the push sequence. This provides a measure of how close the pusher is able to approach the target with the object.

\begin{figure*}
	\centering
	\includegraphics[width=\textwidth,trim=50 360 45 150,clip]{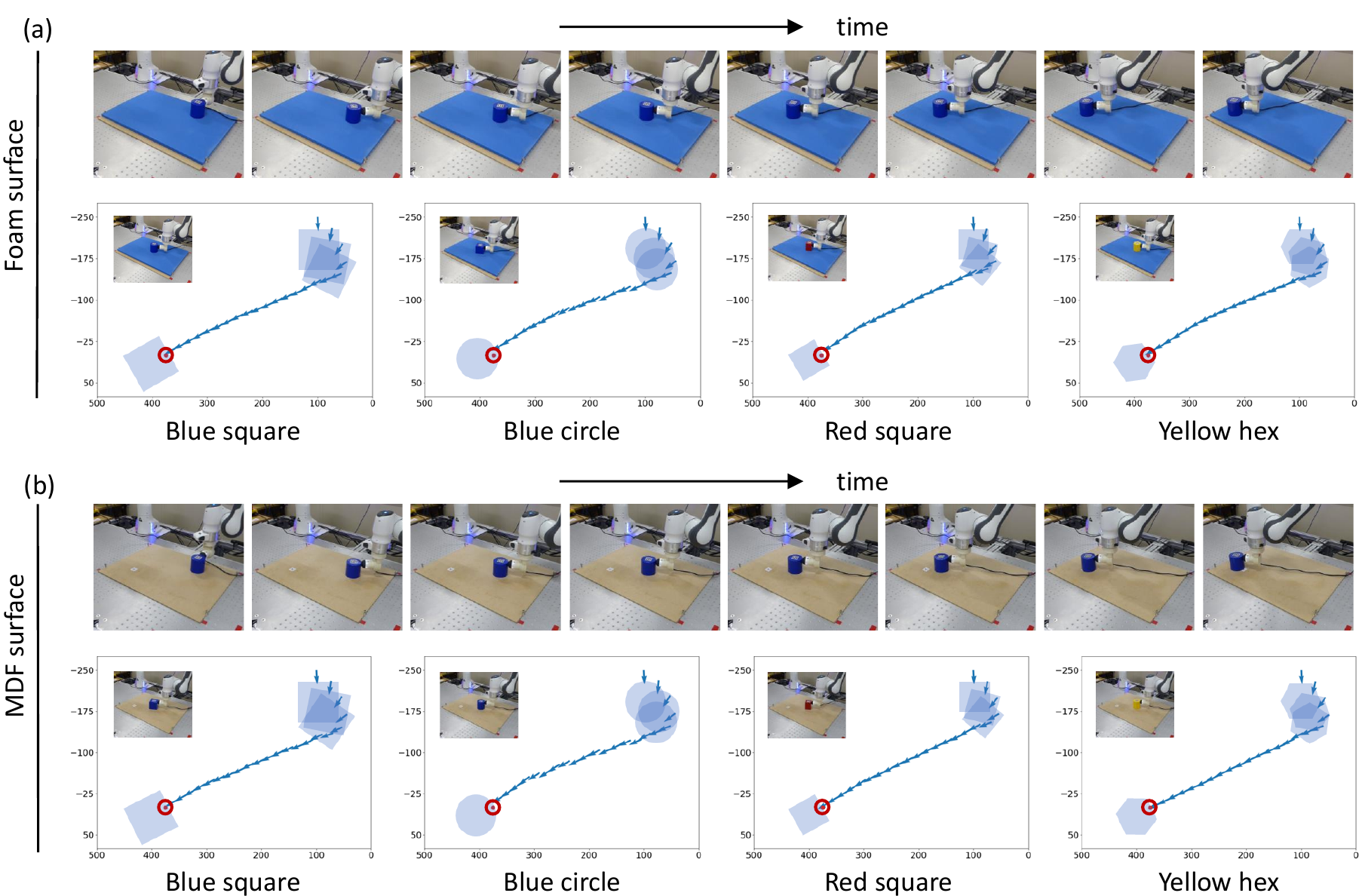}
	\caption{Using a single robot arm to push regular geometric objects across: (a) a foam surface (time lapse photos show sequence for blue square prism), and (b) an MDF surface (time lapse photos show sequence for blue circular prism). In the 2D pose plots, the target is identified by a small red circle and dot.
	\label{fig:single_arm_pushing}}
\end{figure*}

The push sequences are visualized by plotting the end-effector poses in 2D overlaid with approximate poses of the pushed objects at the start and finish points of the trajectory (Figure~\ref{fig:single_arm_pushing}). Our tactile robotic system can push all these regular geometric objects over foam and MDF surfaces to the target, approaching within 10\,mm for the blue circular prism and within 5\,mm for the other objects (Table~\ref{tab:single_arm_pushing_results}).

\begin{table}[t]
	\small\sf\centering
	\caption{Single-arm pushing final target error (mean $\pm$ standard deviation perpendicular distance from target to sensor-object contact normal on completion of push sequence). All statistics are computed over 5 independent trials.
	\label{tab:single_arm_pushing_results}}
	\begin{tabular}{ccc}
	\toprule
	Object & Foam surface & MDF surface\\
	\midrule
	Blue square & 0.50 $\pm$ 0.56 mm & 4.45 $\pm$ 0.78 mm\\	
	Blue circle & 6.93 $\pm$ 0.06 mm & 9.45 $\pm$ 0.13 mm\\	
	Red square & 1.88 $\pm$ 0.71 mm & 4.38 $\pm$ 0.77 mm\\	
	Yellow hexagon & 0.33 $\pm$ 1.32 mm & 2.87 $\pm$ 0.68 mm\\
	\bottomrule
	\end{tabular}
\end{table}

\subsection{Task 4: Dual-arm object pushing}
\label{sec:dual_arm_object_pushing_exp}

In the second pushing experiment, we use a follower robot arm to constrain and stabilise objects as they are pushed across a flat surface by the leader arm. In many ways, this configuration is similar to that used in the object tracking experiment (Section~\ref{sec:object_pose_tracking_multi_exp}), where a leader robot arm moved an object in a complex trajectory while a follower arm tracked its motion and held the object with the first arm.

The experiment is split into two parts. In the first part, we use two robot arms to push the objects used in the previous single-arm experiment across foam and MDF surfaces. In the second part of the experiment, we replace the original set of geometric objects with a set of taller, double-height versions together with several taller everyday objects (e.g. bottles and containers). These taller objects cannot be pushed by a single robot arm without toppling over, so the second stabilising follower arm is essential for the task. 

For this dual-arm configuration, we mounted tactile sensors on both robot arms using right-angle adapters (see Figure~\ref{fig:system_architecture}(d)). At the start of each trial, the leader arm and object were positioned as they were positioned at the start of each single-arm pushing trial. Then we positioned the follower arm so that its tactile sensor was approximately opposite the leader arm tactile sensor and normal to the opposite contacted surface. During each trial, we used the leader robot arm to push the object towards the same target as before, at position $(y,z)=(0,375)$\,mm while both end-effectors remained in contact with the object.

To control the leader robot arm, we used the same pushing controller and parameters as for the single-arm configuration. To control the stabilising follower arm, we used the tactile servoing controller described in Section~\ref{sec:servoing_controller} with the parameters listed in Table~\ref{tab:stabiliser_control_params} of Appendix~\ref{sec:controller_parameters}.

During each trial for both parts of the experiment, we recorded the end-effector poses and corresponding time stamps at each control cycle to match up the corresponding poses at different trajectory points for plotting.

\subsubsection{Pushing regular geometric objects.}
\label{sec:dual_arm_short_object_pushing_exp}

\begin{table}[t]
	\small\sf\centering
	\caption{Dual-arm pushing target error for short geometric objects (mean $\pm$ standard deviation) of the distance from target to sensor-object contact normal on completion of push sequence. All statistics are over 5 independent trials.
	\label{tab:dual_arm_pushing_results}}
	\begin{tabular}{ccc}
	\toprule
	Object & Foam surface & MDF surface\\
	\midrule
	Blue square & 4.43 $\pm$ 0.25 mm & 5.24 $\pm$ 0.24 mm\\	
	Blue circle & 4.46 $\pm$ 3.35 mm & 3.58 $\pm$ 2.04 mm\\	
	Red square & 4.97 $\pm$ 0.18 mm & 4.28 $\pm$ 0.15 mm\\	
	Yellow hexagon & 4.43 $\pm$ 0.31 mm & 4.44 $\pm$ 0.62 mm\\
	\bottomrule
	\end{tabular}
 \vspace{2em}
	\small\sf\centering
	\caption{Dual-arm pushing target error for tall objects (mean $\pm$ standard deviation perpendicular distance from target to sensor-object contact normal on completion of push sequence). All statistics are over 5 independent trials.
	\label{tab:dual_arm_pushing_tall_results}}
	\begin{tabular}{cc}
	\toprule
	Object & MDF surface\\
	\midrule
	Tall blue square & 3.62 $\pm$ 0.31 mm\\	
	Tall blue circle & 7.11 $\pm$ 2.18 mm\\	
	Tall red square & 4.94 $\pm$ 0.19 mm\\	
	Tall yellow hexagon & 4.73 $\pm$ 0.38 mm\\
	\midrule
	Mustard bottle & 6.24 $\pm$ 2.33 mm\\	
	Cleaner bottle & 7.02 $\pm$ 0.95 mm\\	
	Windex spray & 4.59 $\pm$ 1.36 mm\\	
	Glass bottle & 6.42 $\pm$ 0.39 mm\\
	Coffee tin & 4.81 $\pm$ 2.09 mm\\
	\bottomrule
	\end{tabular}
\end{table}

\begin{figure*}
	\centering
	\includegraphics[width=\textwidth,trim=50 355 45 150,clip]{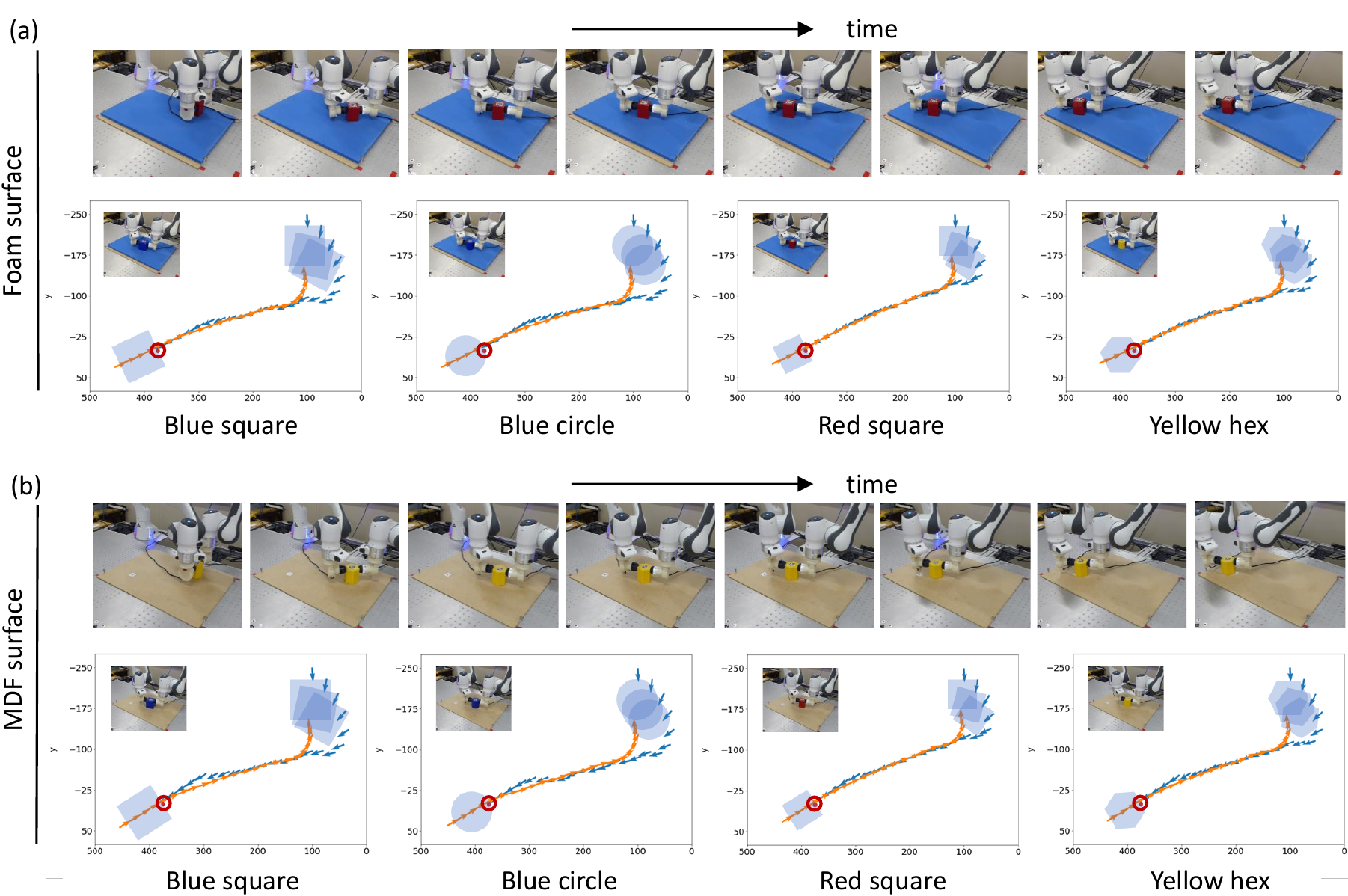}
	\caption{Using a leader and follower robot arm to push regular geometric objects across: (a) a foam surface (time-lapse photos show sequence for red square prism), and (b) an MDF surface (time-lapse photos show sequence for yellow hexagonal prism). In the 2D pose plots, the target is identified by a small red circle and dot.
	\label{fig:dual_arm_pushing}}
	\centering
	\includegraphics[width=\textwidth,trim=50 490 40 60,clip]{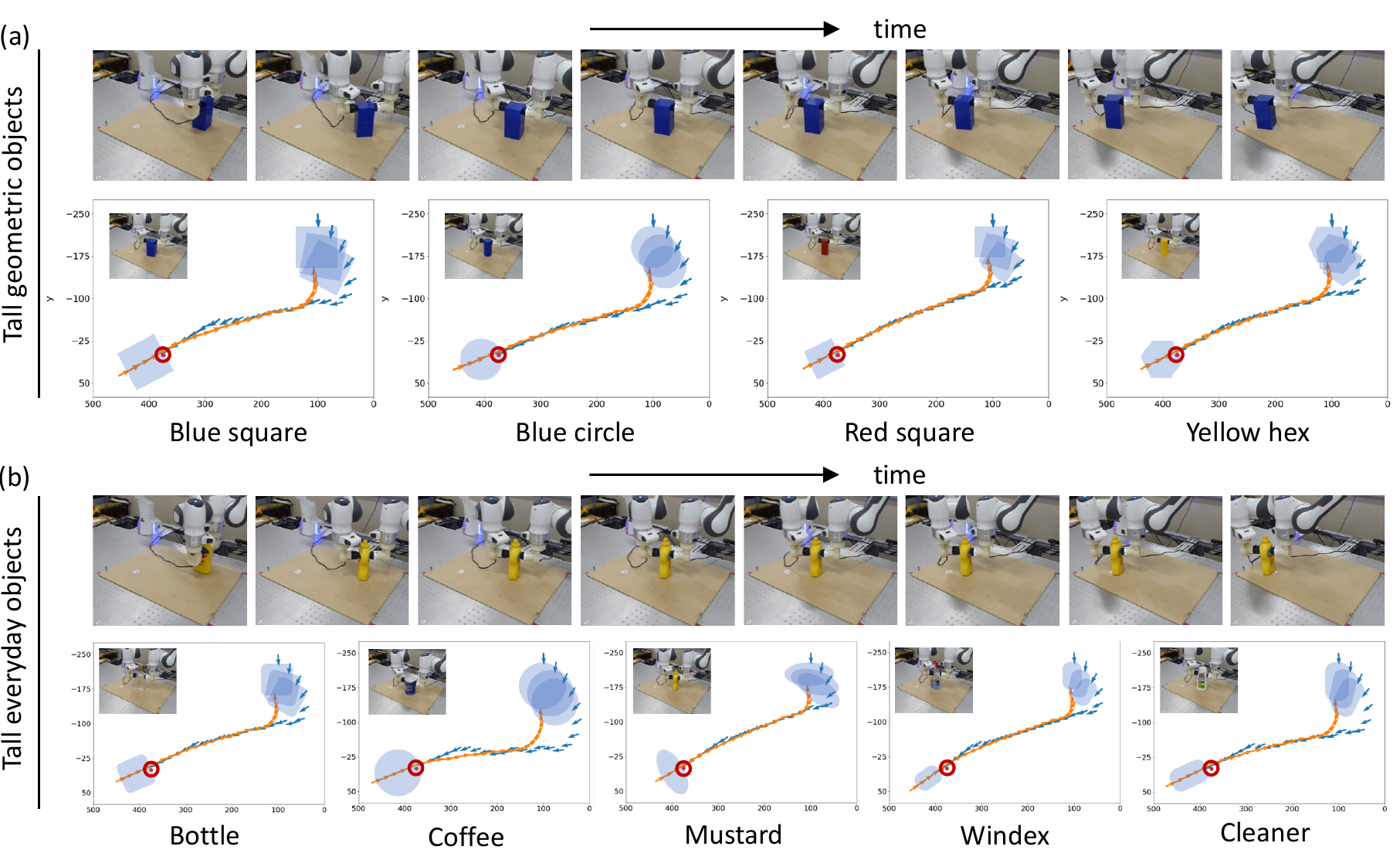}
	\caption{Using a leader and follower robot arm to push tall objects across an MDF surface: (a) tall geometric objects (time-lapse photos show sequence for tall blue square prism), and (b) tall everyday objects (time-lapse photos show sequence for mustard bottle). In the 2D pose plots, the target is identified by a small red circle and dot.
	\label{fig:dual_arm_tall_object_pushing}}
\end{figure*}

In the first part of the dual-arm experiment, we used two robot arms to push the same geometric objects we pushed in the single-arm experiment (Figure~\ref{fig:dual_arm_pushing}). We repeated the experiment five times for each object and then computed the mean $\pm$ standard deviation target error across all five trials (Table~\ref{tab:dual_arm_pushing_results}).

We visualised examples of the push sequences by plotting the end-effector poses of both robot arms in 2D and overlaid the approximate poses of the pushed objects at the start and finish points of the trajectory (Figure~\ref{fig:dual_arm_pushing}).

The results in Table~\ref{tab:dual_arm_pushing_results} and Figure~\ref{fig:dual_arm_pushing} show that our dual-arm system can push the regular geometric objects over foam and MDF surfaces, approaching the target to within less than 5 mm for all objects. In contrast to the results for the single-arm configuration, the accuracy achieved for the blue circular prism did not appear much worse than for the other objects. In fact, for the MDF surface, the accuracy obtained for the circular prism was slightly better than the other objects.

\subsubsection{Pushing tall objects that are prone to toppling.}
\label{sec:dual_arm_tall_object_pushing}

In the second part of the dual-arm experiment, we used the two robot arms to push a set of taller (double-height) geometric objects and tall everyday objects (Figure~\ref{fig:tall_geometric_pushing_objects}) across a surface.

For this part of the experiment, we found that we needed to modify the (feedback) reference contact pose used in the pushing controller of the leader robot to $(0.5,0,0,0,0,0)$ and the reference contact pose used in the servoing controller of the follower robot to $(-0.5,0,3,0,0,0)$. These modified poses only differ from their defaults (Tables~\ref{tab:pushing_control_params} and \ref{tab:stabiliser_control_params}) by 0.5\,mm in the first components. The effect is to apply a slight downward force on the pushed side of the object and slight upward force on the stabilised side. This helps prevent these taller objects from catching their leading edges on the surface as they are being pushed. Even so, we were not able to push these taller objects across the foam surface without their leading edges catching, and so we could only perform this part of the experiment on the harder MDF surface.


Once again, we visualised examples of the push sequences by plotting the end-effector poses of both robot arms in 2D and overlaid the approximate poses of the pushed objects at the start and finish points of the trajectory (Figure~\ref{fig:dual_arm_tall_object_pushing}).

The results in Table~\ref{tab:dual_arm_pushing_tall_results} and Figure~\ref{fig:dual_arm_tall_object_pushing} show that our dual-arm system can push the taller geometric and everyday objects over the MDF surface, approaching the target to within less than 7.5 mm for all objects.

\section{Discussion and limitations}
\label{sec:discussion}

In this paper, we proposed and evaluated a tactile robotic system that uses contact pose and post-contact shear estimation to facilitate object tracking, surface following, and single- and dual-arm object pushing using various configurations of velocity-based control. Our tactile robotic system has two key aspects that enable its generality and ease of control: (a) it estimates both contact pose and post-contact shear, and (b) it enables smooth continuous control of the robot arm by using these estimates to control velocity directly. These aspects enable the robot arm to track objects in six degrees of freedom; this control is either the primary goal, such as in object tracking, or as a secondary constraint on a primary goal, such as when pushing an object along a trajectory while maintaining a contact pose. To achieve these goals, we employed $SE(3)$ Lie group theory to leverage techniques from probability and control theory that were developed originally for Euclidean vector spaces. This novel perspective for our perception and control methodology provides a robust theoretical foundation underlying the experimental demonstrations and results presented here.

\subsection*{Contact pose and post-contact shear estimation}
\label{sec:discussion_contact_pose_shear}

A key simplifying assumption was to merge the contact pose and the post-contact shear into a unified surface contact pose-and-shear vector $(x,y,z,\alpha,\beta,\gamma)$. This was facilitated by the contact pose for a flat planar surface only having meaningful components $(z,\alpha,\beta)$ in contact depth and angle; the other components translate and rotate parallel to the plane and can be ignored in pose-based tactile servo control~(\cite{lepora2021pose,lloyd2021goal}). Here we use these other components $(x,y,\gamma)$ to represent post-contact shear, which in combination with the contact pose is represented by a single $SE(3)$ transformation (Section~\ref{sec:surface_contact_poses}).

This combined surface contact pose-and-shear vector can be estimated using multi-output regression CNNs directly from tactile sensor images, using methods similar to previous work on tactile pose estimation of surfaces and edges~(\cite{lepora2020optimal}). However, for the shear components, the estimates are inaccurate with a larger error even after hyperparameter tuning, particularly for the rotational $\gamma$ component (Figure~\ref{fig:nn_pose_shear_estimate_performance}). A pose-and-shear estimate with this level of error is not suitable for smooth and accurate robot control.


Consequently, we developed a Gaussian density network (GDN) model that combines the CNN base (feature encoding) architecture with output layers that predict both an estimate of the mean and its uncertainty for each pose-and-shear component (Section~\ref{sec:gdn_model}). These estimates are slightly more accurate with the GDN model than with the regression CNN model (Table~\ref{tab:nn_pose_shear_estimate_results}), but more importantly the predicted uncertainties become lower as the predicted means fall closer to the ground truths (Figure~\ref{fig:nn_pose_shear_estimate_performance}). Hence, the GDN model predictions of the means and uncertainties are suitable for Bayesian filtering to reduce error and uncertainty in a sequence of pose-and-shear estimates. 

Therefore, we proposed a novel $SE(3)$ discriminative Bayesian filter to decrease the error and uncertainty of the GDN pose-and-shear estimates. These filtered estimates can be highly accurate for all pose-and-shear components (Table~\ref{tab:bayes_filter_results} and Figure~\ref{fig:bayes_filter_performance}), with this accuracy depending upon the assumed noise in the state dynamics model used with the filter and how well this matches changes in the pose and shear across time steps.  

Both the GDN model and the $SE(3)$ Bayesian filter were technically challenging to implement. The GDN model can be viewed as single-component mixture density network (\cite{bishop1994mixture, bishop2006pattern}), which suffer from problems such as training instability and mode collapse (\cite{hjorth1999regularisation, makansi2019overcoming}). To overcome these difficulties, we incorporated several novel extensions to the model architecture including a {\em softbound} activation function, a new {\em multi-dropout} regularization of the outputs and a loss function that depends on the {\em inverse} standard deviation. Furthermore, there is no simple closed-form method for an $SE(3)$ Bayesian filter because the standard assumption that the normalized product of two Gaussian distributions is Gaussian does not hold on non-Euclidean manifolds such as $SE(3)$. Instead, we used an iterative approximation to the prediction and correction steps of the filter that we re-derived using techniques in Lie group theory (Appendices~\ref{sec:math_prelim}-\ref{sec:se3_probabilistic_fusion}). 

The most obvious limitation of our pose-and-shear estimation methods is the inaccurate CNN regression and GDN model performance on the shear components (Figure~\ref{fig:nn_pose_shear_estimate_performance}). We believe that much of this estimation error is due to tactile aliasing (\cite{lloyd-RSS-21}), whereby similar tactile images in the training set become associated with very different shear labels. Specifically, when the sensor is sheared sufficiently after contacting a surface, it can slip across the surface to result in similar tactile images for a range of post-contact shear labels. If we could prevent this slippage, the complexity of the system might be reduced as the GDN and Bayesian filter could become redundant. However, to do this, the training data would need restricting to samples where slip does not occur, which may be difficult to arrange in practice and could overly restrict the model's applicability, {\em e.g.} just to large contact depths. That said, the TacTip is known to be effective at detecting slip~(\cite{james2020slip,james2018slip}), which potentially could be used to minimize slip in the data collection or provide a label of slip occurrence. Our expectation is that the single-step errors may be reduced but Bayesian filtering will still be needed to reduce the error for accurate control.

Note that including further variation in the trajectories during training could also improve the system performance by giving better model generalization during the task. At present, our training data collection (Algorithm~\ref{alg:data_collection}) first moves the sensor normal to the surface to make contact then parallel with the surface to produce shear. Introducing motions that are more like those during the task could give better model predictions, such as trajectories that vary in shape and have both normal and parallel components of motion while in contact. However, the view taken in this paper is that the issue of aliasing from slippage is of greater initial concern and thus the primary issue to focus upon. 

Another limitation is that in this study we concentrated on estimating contact poses and shears with flat or gently curving surfaces. Clearly, there is a much wider range of surface features that could be tracked, such as following around edges (\cite{lepora2019pixels,lepora2021pose}). In principle, one can train pose estimation models on other object features, for example to predict a 5-component contact pose with a straight edge. However, it would then not be possible to combine a contact pose with a 3-component post-contact shear to form a single 6D pose. Even so, we could still use a GDN model to predict both the contact pose and the post-contact shear motion simultaneously. In this scenario, we would need to combine the outputs of two feedforward-feedback pose controllers ({\em e.g.} one for the contact pose and one for post-contact shear), and there would be subtleties to be addressed in how this would be best realised to transform appropriately under $SE(3)$. 

\subsection*{Experimental servo control task performance}
\label{sec:discussion_servo_manipulation_tasks}

In developing our new tactile robotic system, a primary objective was to achieve smooth and continuous motion of the robot arm using velocity control driven by tactile pose-and-shear estimation. We accomplished this aim by updating the velocity of the end effector during each control cycle, instead of updating its pose based on tactile pose estimation as considered previously (\cite{lepora2021pose,lloyd2021goal}). The underlying tactile servoing uses an MIMO PID feedback controller on the $SE(3)$ error between the estimated and reference pose and shear~(Figure~\ref{fig:pushing_controller}, top) with an additional feedforward velocity supplied for some tasks. For object manipulation, this tactile servo controller is augmented with a {\em target alignment} SISO PID feedback controller that uses the end effector pose relative to the goal to steer the object (Figure~\ref{fig:pushing_controller}, bottom). Specifically, we control the continuous tangential motion, which improves on our previous method of repeatedly breaking contact and discretely pushing the object (\cite{lloyd2021goal}). 

The pose and shear-based tactile servo controller was applied successfully to four distinct tasks: (1)~{\em object pose tracking}, where a leader robot arm moves an object in 3D space while a follower robot arm uses a tactile sensor to track and hold the object; (2) {\em surface following}, where a robot arm uses a tactile sensor to move smoothly over curved surfaces; (3)~{\em single-arm object pushing}, where a robot arm uses a tactile sensor to smoothly push an object to a goal location; and (4) {\em dual-arm object pushing}, where a leader robot arm uses a tactile sensor to push an object while a follower robot arm uses a tactile sensor to track and hold the object. 

Three of these four tasks were made possible by the tactile robotic system's ability to estimate post-contact shear motion in addition to the contact pose. For object pose tracking, the shear motion is essential to track tangential and rotational motion with respect to the contacted surface while maintaining contact, as is visible in the experiments (Figures~\ref{fig:object_tracking_single},\ref{fig:object_tracking_multi}). For single-arm object pushing, controlling shear is essential both to maintain contact with the object and to steer the object via a tangential motion (Figure~\ref{fig:single_arm_pushing}). Likewise, for dual-arm object pushing, estimating post-contact shear is essential for the follower arm to remain in contact (Figure~\ref{fig:dual_arm_pushing}) and hold the tall objects to prevent them being toppled (Figure~\ref{fig:dual_arm_tall_object_pushing}). 

For the other task of surface following, we did not need to control shear to complete the task, so the corresponding gains were set to zero; nevertheless, the pose-and-shear components are mixed in the Bayesian filter over $SE(3)$ so the estimated shear was still used implicitly. In principle, shear control could be used to limit the sliding motion velocity, for example to move slowly when the tactile sensor is pressed strongly into a high friction surface to avoid damage to the sensor or surface.  

One limitation of the tactile robotic system is that we could only successfully push tall objects with two arms on the smooth (MDF) surface, as they kept catching on the foam surface. This is partly due to the nature of the task, as humans can struggle with this too, before they adopt a strategy of partially lifting the object. In principle, the tactile controllers could do this too, but this would be a new task of guiding a lifted object, which is beyond the scope of this investigation.  

Another limitation of the present system is the absence of a planning component, which hinders its ability to anticipate and prevent undesired situations, such as collisions between robot arms, or motion trajectories that approach or reach joint limits or singularities. Implementing this planning capability would also be beneficial in scenarios where the system is unable to achieve a global task objective by following a local control objective, such as non-holonomic object pushing and manipulation tasks where an object must be rotated to a target orientation while also being moved to a target position.

For future developments of this type of tactile robotic system, we believe there are many more manipulation tasks that can be achieved by enabling both robot arms to operate in an active configuration, where they are both functioning as leaders and followers to some extent. This would broaden their capacity to collaborate, particularly on tasks relating to more complex types of tactile-enabled object manipulation. Such tasks could span from guided manipulation of an object to a goal pose or insertion/assembly of one object into/onto another, to more general tasks involving multiple tactile sensors to enable fully-dexterous manipulation with pairs of tactile grippers or multi-fingered tactile robot hands. 

\begin{acks}

This work was supported by an research leadership award from the Leverhulme Trust: “A Biomimetic Forebrain for Robot Touch” (RL-2016-39). We also thank Andy Stinchcombe and
members of the Dexterous Robotics group at Bristol Robotics Laboratory.

\end{acks}

\bibliographystyle{SageH}
\bibliography{references.bib}

\begin{appendices}

\section{Notation and mathematical preliminaries}
\label{sec:math_prelim}

In this appendix, we define our notation and give some basic properties of matrix Lie groups and algebras, focussing on the Special Euclidean group $SE(3)$ of rotations and translations in 3D. We also describe how we represent probability distributions in $SE(3)$. A more comprehensive introduction to Lie groups applied to robotics can be found in \cite{barfoot2017state, sola2018micro}.

A \emph{Lie group} is a group that is also a smooth, differentiable manifold. Hence, the group composition and inversion operations are smooth, differentiable operations. A \emph{matrix Lie group} $G$ is a smooth manifold in the set of $\mathbb{R}^{n \times n}$ matrices that is closed under composition and where the composition and inversion operations are matrix multiplication and inversion, respectively. The group identity is the $n \times n$ identity matrix $\boldsymbol{\mathrm{1}}_{n\times n}$. In this paper, we focus on the Special Euclidean Group of rotations and translations in 3D, $SE(3)$, because it can be used to represent poses and shears, pose-and-shear transformations or changes of coordinate frame in 3D. The elements of $SE(3)$ can be represented as
\begin{equation}
	\label{eqn:se3_matrix_def}
	SE(3)
	\, = \,
	\{ \boldsymbol{\mathrm{X}}
	=
	\begin{bmatrix}
		\boldsymbol{\mathrm{C}} & \boldsymbol{\mathrm{r}}\\
		\boldsymbol{\mathrm{0}}^{\top} & 1
	\end{bmatrix}
	: \, \boldsymbol{\mathrm{C}} \in SO(3)
	, \, \boldsymbol{\mathrm{r}} \in \mathbb{R}^{3} \},
\end{equation}
where $\boldsymbol{\mathrm{C}}$ is a $3 \times 3$ orthonormal matrix ($\boldsymbol{\mathrm{C}}^\top\boldsymbol{\mathrm{C}}=\boldsymbol{\mathrm{C}}\boldsymbol{\mathrm{C}}^\top=\boldsymbol{\mathrm{1}}_{3\times 3}$) giving the rotational component of the transformation, and the column vector $\boldsymbol{\mathrm{r}}$ gives the translational component.


Because Lie groups are manifolds, they have tangent spaces, which at the origin is called the \emph{Lie algebra}, representing directions of motion in the group. For $SE(3)$, the Lie algebra $\mathfrak{se}(3)$ is isomorphic to a 6-dimensional vector space  $\mathbb{R}^{6}$, representing the three translational and three rotational degrees of freedom. Elements $\boldsymbol\Xi$ of the Lie algebra map onto corresponding elements $\boldsymbol{\mathrm{X}}$ of the Lie group via the \emph{exponential map}, $\exp(\boldsymbol\Xi)$, where: 
\begin{equation}
\boldsymbol{\mathrm{X}} \,=\, \exp(\boldsymbol\Xi),\hspace{2em}
\boldsymbol\Xi \, = \, \ln(\boldsymbol{\mathrm{X}}).
\end{equation}
This exponential function is defined by an infinite series analogous to the corresponding scalar exponential series, where the successive powers of $\boldsymbol\Xi$ are found recursively by matrix multiplication. For the inverse map, elements of the matrix group are mapped into the Lie algebra using the \emph{logarithmic map}, $\ln(\boldsymbol{\mathrm{X}})$, defined by an infinite series analogous to the corresponding scalar logarithmic series. 

We use the notation $\left[ \cdot \right]^{\wedge}$ to represent mapping an element $\boldsymbol\xi$ of the Euclidean vector space onto its corresponding element $\boldsymbol\Xi$ of the Lie algebra. For $\mathfrak{se}(3)$, this operation is defined as:
\begin{equation}
	\label{eqn:se3_hat_operator}
	\boldsymbol\Xi
	\, = \,
	\boldsymbol\xi^{\wedge}
	\, = \,
	\begin{bmatrix}
		\boldsymbol\rho\\
		\boldsymbol\phi
	\end{bmatrix} ^ {\wedge}
	\, = \,
	\begin{bmatrix}
		\boldsymbol\phi^{\wedge} & \boldsymbol\rho\\
		\boldsymbol{\mathrm{0}}^{\top} & 0
	\end{bmatrix},
\end{equation}
where $\boldsymbol\rho, \boldsymbol\phi \in \mathbb{R}^{3}$ represents the three translational and three rotational components of $\boldsymbol\xi$, with $\boldsymbol\phi^{\wedge}$ the skew-symmetric matrix representation of $\boldsymbol\phi$:
\begin{equation}
	\label{eqn:so3_hat_operator}
	\boldsymbol\phi^{\wedge}
	\, = \,
	\begin{bmatrix}
		\phi_{1}\\
		\phi_{2}\\
		\phi_{3}
	\end{bmatrix}^{\wedge}
	\, = \,
	\begin{bmatrix}
		0 & -\phi_{3} & \phi_{2}\\
		\phi_{3} & 0 & -\phi_{1}\\
		-\phi_{2} & \phi_{1} & 0
	\end{bmatrix}.
\end{equation}
In this paper, we follow the convention adopted in \cite{barfoot2017state} and \cite{murray2017mathematical} and use the first three components of $\boldsymbol\xi$ to represent $\boldsymbol\rho$ and the last three components to represent $\boldsymbol\phi$. This differs from the convention used in \cite{lynch2017modern}, which reverses this order. 

Note that we use the notation $\left[ \cdot \right]^{\vee}$ to represent the inverse mapping: $\boldsymbol\xi \, = \, \boldsymbol\Xi^{\vee}$. In robot kinematics, $\boldsymbol\xi$ is often referred to as a \emph{velocity twist}.

Elements of the Lie group $\boldsymbol{\mathrm{X}} \in G$ act upon elements of the Lie algebra $\boldsymbol\Xi\in\mathfrak{g}$ using the \emph{adjoint representation}, $\mathrm{Ad}_{\boldsymbol{\mathrm{X}}}\!\left( \boldsymbol{\mathrm{X}} \right)\boldsymbol\Xi=\boldsymbol{\mathrm{X}} \boldsymbol\Xi \boldsymbol{\mathrm{X}}^{-1}$, which for $SE(3)$ is represented by:
\begin{equation}
	\label{eqn:se3_adjoint}
	\begin{split}
		\mathrm{Ad}\!\left( \boldsymbol{\mathrm{X}} \right)
		& \, = \,
		\mathrm{Ad}\! 
		\left(
		\begin{bmatrix}
			\boldsymbol{\mathrm{C}} & \boldsymbol{\mathrm{r}}\\
			\boldsymbol{\mathrm{0}}^{\top} & 1
		\end{bmatrix}
		\right)
        \, = \,
		\begin{bmatrix}
			\boldsymbol{\mathrm{C}} & \boldsymbol{\mathrm{r}}^{\wedge} \boldsymbol{\mathrm{C}}\\
			\boldsymbol{\mathrm{0}}_{3 \times 3} & \boldsymbol{\mathrm{C}}
		\end{bmatrix}.
	\end{split}
\end{equation}
acting on the $\mathbb{R}^{6}$ vector space representation $\boldsymbol\xi=\left[\boldsymbol\rho, \boldsymbol\phi\right]^\top$ of $\mathfrak{se}(3)$.
Then the adjoint representations of the group composition and inverse operations are $\mathrm{Ad}\!\left( \boldsymbol{\mathrm{X}}_{1} \boldsymbol{\mathrm{X}}_{2} \right) = \mathrm{Ad}\!\left( \boldsymbol{\mathrm{X}}_{1} \right) \mathrm{Ad}\!\left( \boldsymbol{\mathrm{X}}_{2} \right)$ and $\mathrm{Ad}\!\left( \boldsymbol{\mathrm{X}} \right)^{-1} = \mathrm{Ad}\!\left( \boldsymbol{\mathrm{X^{-1}}} \right)$. Likewise, the corresponding adjoint representation of elements in the Lie algebra $\boldsymbol\Xi\in\mathfrak{g}$ is denoted as $\mathrm{ad}\!\left( \boldsymbol\Xi \right)$ and for $\mathfrak{se}(3)$ is:
\begin{equation}
	\label{eqn:se3_algebra_adjoint}
		\mathrm{ad}\!\left( \boldsymbol\Xi \right)
          =
  	\boldsymbol\xi^{\curlywedge}
		 = 
		\mathrm{ad}\!\left( \boldsymbol\xi^{\wedge} \right)
		 = 
		\mathrm{ad}\!\left(
		\begin{bmatrix}
			\boldsymbol\rho\\
			\boldsymbol\phi
		\end{bmatrix} ^ {\wedge}
		\right)
         = 
		\begin{bmatrix}
			\boldsymbol\phi^{\wedge} & \boldsymbol\rho^{\wedge} \\
			\boldsymbol{\mathrm{0}} & \boldsymbol\phi^{\wedge}
		\end{bmatrix},
\end{equation}
which acts on the $\mathbb{R}^{6}$ vector space representation $\boldsymbol\xi=\left[\boldsymbol\rho, \boldsymbol\phi\right]^\top$ of $\mathfrak{se}(3)$ with components $\rho_k$ and $\phi_k$.

The product of two exponentials in $\mathfrak{se}(3)$ can be computed using the following approximation, based on the \emph{Baker-Campbell-Hausdorff} (BCH) formula (\cite{barfoot2017state}):
\begin{multline}
	\label{eqn:bch_approx}
	\ln\!\big(\!\exp(\boldsymbol\Xi_{1}) \, \exp(\boldsymbol\Xi_{2})\big)^{\vee}
	\, = \,
	\ln\!\big(\!\exp(\boldsymbol\xi^{\wedge}_{1}) \, \exp(\boldsymbol\xi^{\wedge}_{2})\big)^{\vee}\\
	\approx
	\begin{cases}
		\mathcal{J}(\boldsymbol\xi_{2})^{-1}\boldsymbol\xi_{1} + \boldsymbol\xi_{2} & \text{if $\boldsymbol\xi_{1}$ small}\\
		\boldsymbol\xi_{1} + \mathcal{J}(-\boldsymbol\xi_{1})^{-1}\boldsymbol\xi_{2} & \text{if $\boldsymbol\xi_{2}$ small}\\
	\end{cases} 
\end{multline}
Here, $\mathcal{J} \in \mathbb{R}^{6 \times 6}$ is the \emph{left Jacobian} of $SE(3)$, which can be written as the following series expansion:
\begin{equation}
	\label{eqn:left_jacobian_def}
	\mathcal{J} \left( \boldsymbol\xi \right)
	 = 
	\sum_{n=0}^{\infty}
	\frac{\left( \boldsymbol\xi^{\curlywedge} \right)^{n}} {\left( n+1 \right)!},\ \ 
	\boldsymbol\xi^{\curlywedge}
	 = 
	\mathrm{ad} \left( \boldsymbol\xi^{\wedge} \right)
	 = 
	\begin{bmatrix}
		\boldsymbol\phi^{\wedge} & \boldsymbol\rho^{\wedge} \\
		\boldsymbol{\mathrm{0}} & \boldsymbol\phi^{\wedge}
	\end{bmatrix}.
\end{equation}
Similarly, the inverse (left) Jacobian can be written as another series expansion:
\begin{equation}
	\label{eqn:inv_jacobian_def}
	\mathcal{J} \left( \boldsymbol\xi \right)^{-1}
	\, = \,
	\sum_{n=0}^{\infty}
	\frac {B_{n}} {n!}
	\left( \boldsymbol\xi^{\curlywedge} \right)^{n},
\end{equation}
where $B_{n}$ are the \emph{Bernoulli numbers}, $B_{0}=1, \, B_{1}=-\frac{1}{2}, \, B_{2}=\frac{1}{6}, \, B_{3}=0, \, B_{4}=-\frac{1}{30}, \, \cdots$. In this paper, if we need to calculate the Jacobian or its inverse, we typically truncate the corresponding series after second-order terms because we have found that this is sufficiently accurate for our purposes (this is also consistent with the findings of \cite{barfoot2014associating}). Note that it is also possible to define a \emph{right Jacobian} of $SE(3)$ but we will not need it here and so will just refer to $\mathcal{J}$ as the \emph{Jacobian}. 

\section{SE(3) random variables and probabilistic transformations}
\label{sec:se3_probabilistic_transform}

It is well known that the PDF of a linearly-transformed Gaussian random variable with added Gaussian noise is itself Gaussian, and its parameters can be computed analytically without having to explicitly evaluate a marginalisation integral like the one in Equation~\ref{eqn:bayes_filter_predict}. So, if we were using a linear-Gaussian state dynamics model with our Bayesian filter, we could efficiently compute the prediction step in Equation~\ref{eqn:bayes_filter_predict} in closed form. In this section, we will derive an analogous simplification for $SE(3)$ random variables.

Let us consider a random variable $\boldsymbol{\mathrm{X}}\in SE(3)$. Following \cite{barfoot2014associating, barfoot2017state, bourmaud2016intrinsic}, we define a PDF:
\begin{equation}
	\label{eqn:se3_random_variable_2}
	\boldsymbol{\mathrm{X}} = 
	\exp({\boldsymbol\epsilon}^{\wedge}) \bar{\boldsymbol{\mathrm{X}}};\hspace{1em} 
    \boldsymbol{\epsilon} \sim \mathcal{N}(\boldsymbol{\mathrm{0}},\boldsymbol{\Sigma}),\ \  
    \bar{\boldsymbol{\mathrm{X}}} \in SE(3),
\end{equation}
with deterministic mean $\bar{\boldsymbol{\mathrm{X}}}\in SE(3)$ left-multiplied by~a small zero-mean Gaussian perturbation $\boldsymbol{\epsilon}\in\mathbb{R}^6$ with covariance matrix $\boldsymbol{\Sigma}$. The Gaussian perturbation $\boldsymbol{\epsilon}$ induces a~non-Gaussian PDF over $\boldsymbol{\mathrm{X}}\!\in\!SE(3)$ (\cite{barfoot2014associating})
\begin{equation}
	\label{eqn:se3_pdf}
	p \left( \boldsymbol{\mathrm{X}} \right)
	\, = \,
	\beta \left( \boldsymbol\epsilon \right)
	\exp \left( -\textstyle\frac{1}{2} \boldsymbol\epsilon^{\top}
	\boldsymbol{\mathrm{\Sigma}}^{-1} \boldsymbol\epsilon \right),
\end{equation}
where $ \boldsymbol\epsilon = \ln \big(\boldsymbol{\mathrm{X}} \bar{\boldsymbol{\mathrm{X}}}^{-1} \big)^{\vee}$ and $\beta \left( \boldsymbol\epsilon \right) = {\eta}/{\left| \mathrm{det}( \mathcal{J}(\boldsymbol\epsilon) ) \right|}$. The non-constant normalisation factor $\beta \left( \boldsymbol\epsilon \right)$ originates from the relationship between infinitesimal volume elements in $\mathfrak{se}(3)$ and $SE(3)$ (\cite{barfoot2017state}): $d \boldsymbol{\mathrm{X}}\, = \,
\left| \mathrm{det}( \mathcal{J}(\boldsymbol\epsilon) ) \right| \, d \boldsymbol\epsilon$, which is the reason for the non-Gaussianity.

Sometimes, it is necessary to map $\boldsymbol{\mathrm{X}}$ to a random variable $\boldsymbol\xi\in\mathbb{R}^6$ in the vector space representation of the Lie algebra, and find its PDF. To do this, we first map $\boldsymbol{\mathrm{X}}$ to the Lie algebra using the logarithmic map and then approximate it using the BCH formula (Equation ~\ref{eqn:bch_approx}):
\begin{equation}
	\label{eqn:glob_tangent_space_approx}
	\boldsymbol\xi
	\, = \,
	\ln(\boldsymbol{\mathrm{X}})^{\vee}
	\, \approx \,
	\boldsymbol\mu + \mathcal{J} \left( \boldsymbol\mu \right)^{-1} \boldsymbol\epsilon,
\end{equation}
where $\boldsymbol\mu = \ln(\bar{\boldsymbol{\mathrm{X}}})^{\vee}\in\mathbb{R}^6$. Then, noting that the second term on the right hand side is just a linear transform of the Gaussian random variable $\boldsymbol\epsilon$, we find that the PDF of $\boldsymbol\xi$ is: 
\begin{equation}
	\label{eqn:glob_tangent_space_pdf}
	\boldsymbol\xi \,\sim\, \mathcal{N}(\boldsymbol\mu, \, \mathcal{J}^{-1}(\boldsymbol\mu) \, \boldsymbol{\Sigma} \, \mathcal{J}^{-\top}(\boldsymbol\mu)),
\end{equation}
which is also approximately Gaussian. In the methodology (Section~\ref{sec:discriminative_bayesian_filtering}), we use this expression to map the sensor observations from a distribution defined by $\boldsymbol\mu\in\mathbb{R}^6$ with covariance matrix $\boldsymbol{\Sigma}$ to a distribution defined by $\bar{\boldsymbol{\mathrm{X}}}\in SE(3)$ with perturbation covariance matrix $\mathcal{J}(\boldsymbol\mu)\,\boldsymbol{\Sigma}\,\mathcal{J}^{\top}(\boldsymbol\mu)$. 

Another useful operation is to compound an $SE(3)$ random variable $\boldsymbol{\mathrm{X}}$ with distribution parameters $\bar{\boldsymbol{\mathrm{X}}}$ and $\boldsymbol{\Sigma}$ (Equation~\ref{eqn:se3_random_variable_2}) with a second $SE(3)$ variable sampled from another PDF:
\begin{equation}
	\boldsymbol{\mathrm{T}} = 
	\exp({\boldsymbol\phi}^{\wedge}) \bar{\boldsymbol{\mathrm{T}}};\hspace{1em} 
    \boldsymbol{\phi} \sim \mathcal{N}(\boldsymbol{\mathrm{0}},\boldsymbol{\Sigma}_{\boldsymbol{\phi}}),\ \  
    \bar{\boldsymbol{\mathrm{T}}} \in SE(3),
\end{equation}
where $\bar{\boldsymbol{\mathrm{T}}}$, the deterministic $SE(3)$ mean, is close to $\mathbf{1}$, and $\boldsymbol\phi$ is a second small Gaussian perturbation with covariance matrix $\boldsymbol\Sigma_{\boldsymbol\phi}$.
For $\boldsymbol{\mathrm{X}}^{\prime}=\boldsymbol{\mathrm{T}}\boldsymbol{\mathrm{X}}$, the transformed PDF is to first order in the covariance matrix (see~\cite{barfoot2014associating} for a higher order treatment): 
\begin{equation}
	\label{eqn:se3_rv_probabilistic_transform}
    \boldsymbol{\mathrm{X}}^{\prime}
    =
    \exp \left( {\boldsymbol\epsilon}^{\prime \wedge} \right) \bar{\boldsymbol{\mathrm{X}}}^{\prime};\hspace{1em} 
    \boldsymbol{\epsilon^{\prime}} \sim \mathcal{N}(\boldsymbol{\mathrm{0}},\boldsymbol{\Sigma}^{\prime}),\ \  
    \bar{\boldsymbol{\mathrm{X}}}^{\prime} \in SE(3),
\end{equation}
with corresponding $SE(3)$ mean and covariance matrix:
\begin{equation}
	\label{eqn:se3_rv_probabilistic_transform_2}
    \bar{\boldsymbol{\mathrm{X}}}^{\prime}
    =
    \bar{\boldsymbol{\mathrm{T}}} \bar{\boldsymbol{\mathrm{X}}},\ \ \ \ 
    \boldsymbol\Sigma^{\prime}
     =
    \mathrm{Ad}(\bar{\boldsymbol{\mathrm{T}}}) \, \boldsymbol{\Sigma} \, \mathrm{Ad}(\bar{\boldsymbol{\mathrm{T}}})^{\top} + \boldsymbol{\Sigma}_{\boldsymbol{\phi}}.
\end{equation}
In the methodology (Section~\ref{sec:discriminative_bayesian_filtering}), we use this expression to give the prediction step in the $SE(3)$ discriminative Bayesian filter (Algorithm~\ref{alg:se3_bayes_filter}).

Equation~\ref{eqn:se3_rv_probabilistic_transform_2} is derived by composing Equation~\ref{eqn:se3_random_variable_2} on the left with a deterministic $SE(3)$ transform $\bar{\boldsymbol{\mathrm{T}}}$ to get another random variable of the same form (\cite{barfoot2014associating, barfoot2017state}):
\begin{equation}
	\bar{\boldsymbol{\mathrm{T}}} \boldsymbol{\mathrm{X}}
	\, = \,
	\bar{\boldsymbol{\mathrm{T}}}
	\exp \left( \boldsymbol\epsilon^{\wedge} \right) \bar{\boldsymbol{\mathrm{X}}}
	\, = \,
	\exp \left( \left( \mathrm{Ad}(\bar{\boldsymbol{\mathrm{T}}}) \boldsymbol\epsilon \right)^{\wedge} \right) \bar{\boldsymbol{\mathrm{T}}} \bar{\boldsymbol{\mathrm{X}}}.\nonumber
\end{equation}
The probabilistic transformation to the new random variable $\boldsymbol{\mathrm{X}}^{\prime}$ is completed by adding some Gaussian noise $\boldsymbol{\phi} \sim \mathcal{N}(\boldsymbol{\mathrm{0}}, \, \boldsymbol{\Sigma}_{\boldsymbol{\phi}})$ to the transformed perturbation $\mathrm{Ad}(\bar{\boldsymbol{\mathrm{T}}}) \boldsymbol\epsilon$:
\begin{equation}
	\boldsymbol{\mathrm{X}}^{\prime}
	\, = \,
	\exp \left( \left( \mathrm{Ad}(\bar{\boldsymbol{\mathrm{T}}}) \boldsymbol\epsilon + \boldsymbol\phi \right)^{\wedge} \right) \bar{\boldsymbol{\mathrm{T}}} \bar{\boldsymbol{\mathrm{X}}}.\nonumber
\end{equation}
Therefore, the transformed random variable can be written:
\begin{equation}
        \boldsymbol{\mathrm{X}}^{\prime}
    	=
    	\exp \left( {\boldsymbol\epsilon}^{\prime \wedge} \right) \bar{\boldsymbol{\mathrm{X}}}^{\prime},\ \ \ 
		\bar{\boldsymbol{\mathrm{X}}}^{\prime}
		=
		\bar{\boldsymbol{\mathrm{T}}} \bar{\boldsymbol{\mathrm{X}}},\ \ \ 
		\boldsymbol\epsilon^{\prime}
		 =
		\mathrm{Ad}(\bar{\boldsymbol{\mathrm{T}}}) \, \boldsymbol\epsilon + \boldsymbol\phi.\nonumber
\end{equation}
Since $\boldsymbol{\epsilon}^{\prime} =  \mathrm{Ad}(\bar{\boldsymbol{\mathrm{T}}}) \,\boldsymbol\epsilon + \boldsymbol\phi$
represents a linear transformation of a Gaussian random variable $\boldsymbol\epsilon$ with added Gaussian noise $\boldsymbol\phi$, and both are small, it also has a Gaussian distribution:
\begin{equation}
\label{eqn:se3_prob_transform_perturbation_pdf}
	{\boldsymbol{\epsilon}}^{\prime} \,\sim\, \mathcal{N}(\boldsymbol{\mathrm{0}}, \, \mathrm{Ad}(\bar{\boldsymbol{\mathrm{T}}}) \, \boldsymbol{\Sigma} \, \mathrm{Ad}(\bar{\boldsymbol{\mathrm{T}}})^{\top} + \boldsymbol{\Sigma}_{\boldsymbol{\phi}}),
\end{equation}
which completes the derivation of Equation~\ref{eqn:se3_rv_probabilistic_transform_2}.

\section{SE(3) probabilistic data fusion}
\label{sec:se3_probabilistic_fusion}

In the correction step of our discriminative Bayesian filter, we need to fuse two PDFs by computing their normalised product (Equation~\ref{eqn:bayes_filter_prob_fusion_approx}). For the multivariate Gaussian case, we can do this by computing the normalised product of two Gaussian PDFs, which is also a Gaussian PDF. However, in $SE(3)$, the PDFs are defined over a curved manifold instead of a Euclidean vector space, which necessitates an iterative approach to solve this problem (\cite{barfoot2014associating, bourmaud2016intrinsic, smith2003computing}).

Suppose that our $SE(3)$ state PDFs have the form described in Appendix~\ref{sec:se3_probabilistic_transform}:
\begin{equation}
	\label{eqn:se3_pdf_2}
	p \left( \boldsymbol{\mathrm{X}} \right)
	\, = \,
	\beta \left( \boldsymbol\epsilon \right)
	\exp \left( -\textstyle\frac{1}{2}
	\boldsymbol\epsilon^{\top}
	\boldsymbol{\mathrm{\Sigma}}^{-1}
	\boldsymbol\epsilon \right),\nonumber
\end{equation}
where $\boldsymbol\epsilon = \ln \big(\boldsymbol{\mathrm{X}} \bar{\boldsymbol{\mathrm{X}}}^{-1} \big)^{\!\vee}$ and $\beta \left( \boldsymbol\epsilon \right) = \eta / \left| \mathrm{det}( \mathcal{J}(\boldsymbol\epsilon) ) \right|$. This PDF can then be approximated by (\cite{bourmaud2016intrinsic}):
\begin{equation}
	\label{eqn:se3_pdf_approx}
	p \left( \boldsymbol{\mathrm{X}} \right)
	\approx
	\eta \exp \left( -\textstyle\frac{1}{2}
	\ln \left(\boldsymbol{\mathrm{X}} \bar{\boldsymbol{\mathrm{X}}}^{-1} \right)^{\vee\top}
	\boldsymbol{\mathrm{\Sigma}}^{-1}
	\ln \left(\boldsymbol{\mathrm{X}} \bar{\boldsymbol{\mathrm{X}}}^{-1} \right)^{\vee} \right).
\end{equation}
Here, we replace the variable normalisation factor $\beta \left( \boldsymbol\epsilon \right)$ with the constant normalisation factor $\eta$, because from Equation~\ref{eqn:left_jacobian_def}, $\mathcal{J}(\boldsymbol\epsilon)
 = \boldsymbol{\mathrm{1}} + \frac{1}{2}\epsilon^{\curlywedge} + \dots
\approx \boldsymbol{\mathrm{1}}$ if the perturbation covariance matrix is sufficiently small (i.e., the maximum eigenvalue is sufficiently small).

This \emph{concentrated Gaussian on a Lie group} assumption is made explicitly in \cite{bourmaud2016intrinsic} and is implicit in the data fusion algorithm described in \cite{barfoot2014associating} and \cite{barfoot2017state}. In the latter case, the assumption is implied by the authors' choice of Mahalanobis cost function that they minimise to solve the fusion problem. The reason this assumption is required for probabilistic fusion in $SE(3)$ is that a normalised product of Lie group PDFs is not, in general, equal to another Lie group PDF of the same form (in contrast to the normalised product of Gaussian PDFs in a Euclidean vector space). In fact, the normalised product of two Gaussian Lie group PDFs can only be \emph{approximated} by another PDF of that form.

Using the concentrated Gaussian on a Lie group assumption, the normalised product of two $SE(3)$ PDFs is approximately:
\begin{multline}
\label{eqn:se3_normalised_prod}
	p_{*}\!\left( \boldsymbol{\mathrm{X}} \right)
	\, = \,
	\alpha \, p_{1}\!\left( \boldsymbol{\mathrm{X}} \right) \, p_{2}\!\left( \boldsymbol{\mathrm{X}} \right)\\
	\, \approx \,
	\alpha \eta_{1} \eta_{2} \,
	\exp 
	\left(
	-\textstyle\frac{1}{2}
	\ln \left(\boldsymbol{\mathrm{X}} \bar{\boldsymbol{\mathrm{X}}}_{1}^{-1} \right)^{\vee\top} \!
	\boldsymbol{\mathrm{\Sigma}}_{1}^{-1}
	\ln \left(\boldsymbol{\mathrm{X}} \bar{\boldsymbol{\mathrm{X}}}_{1}^{-1} \right)^{\vee}
	\right)\\
	\times
	\exp 
	\left(
	-\textstyle\frac{1}{2}
	\ln \left(\boldsymbol{\mathrm{X}} \bar{\boldsymbol{\mathrm{X}}}_{2}^{-1} \right)^{\vee\top} \!
	\boldsymbol{\mathrm{\Sigma}}_{2}^{-1}
	\ln \left(\boldsymbol{\mathrm{X}} \bar{\boldsymbol{\mathrm{X}}}_{2}^{-1} \right)^{\vee}
	\right),
\end{multline}
where $\alpha$ is a normalisation factor that ensures the product PDF integrates to $1$ over its support. It can then be shown that this product of $SE(3)$ PDFs can be written as
\begin{multline}
	\label{eqn:se3_normalised_prod_approx}
	p_{*}\! \left( \boldsymbol\epsilon \right)
	\,\approx\,
	\alpha \eta_{1} \eta_{2}\,
	\exp
	\left(
	-\textstyle\frac{1}{2}
	\left(
	\boldsymbol\epsilon - \boldsymbol\mu_{1}^{\prime} \right)^{\top}
	{\boldsymbol{\mathrm{\Sigma}}}_{1}^{\prime -1}
	\left( \boldsymbol\epsilon - \boldsymbol\mu_{1}^{\prime} \right)
	\right)\\
	\times
	\exp
	\left(
	-\textstyle\frac{1}{2}
	\left(
	\boldsymbol\epsilon - \boldsymbol\mu_{2}^{\prime} \right)^{\top}
	{\boldsymbol{\mathrm{\Sigma}}}_{2}^{\prime -1}
	\left( \boldsymbol\epsilon - \boldsymbol\mu_{2}^{\prime} \right)
	\right),
\end{multline}
where the two means and convariance matrices are
\begin{align}
\label{eqn:se3_norm_prod_approx_factor_params_0}
\boldsymbol\mu_{1}^{\prime} = -\mathcal{J}\left( \boldsymbol\xi_{1} \right) \boldsymbol\xi_{1},\ \ \ \ 
\boldsymbol{\mathrm{\Sigma}}_{1}^{\prime} = \mathcal{J} \left( \boldsymbol\xi_{1} \right) \boldsymbol{\mathrm{\Sigma}}_{1}\mathcal{J} \left( \boldsymbol\xi_{1} \right)^{\top},\\
\label{eqn:se3_norm_prod_approx_factor_params}
\boldsymbol\mu_{2}^{\prime} = -\mathcal{J}\left( \boldsymbol\xi_{2} \right) \boldsymbol\xi_{2},\ \ \ \ 
\boldsymbol{\mathrm{\Sigma}}_{2}^{\prime} = \mathcal{J} \left( \boldsymbol\xi_{2} \right) \boldsymbol{\mathrm{\Sigma}}_{2}\mathcal{J} \left( \boldsymbol\xi_{2} \right)^{\top}.
\end{align}
where $\boldsymbol\xi_{1} = \ln \big(\bar{\boldsymbol{\mathrm{X}}} \bar{\boldsymbol{\mathrm{X}}}_{1}^{-1} \big)^{\!\vee}$, $\boldsymbol\xi_{2} = \ln \big(\bar{\boldsymbol{\mathrm{X}}} \bar{\boldsymbol{\mathrm{X}}}_{2}^{-1} \big)^{\!\vee}$. 

To show this, we consider a change of (random) variable using
$\boldsymbol{\mathrm{X}}\, = \,\exp(\boldsymbol\epsilon^{\wedge}) \, \bar{\boldsymbol{\mathrm{X}}}$ and
$\boldsymbol\epsilon\, = \,\ln \big(\boldsymbol{\mathrm{X}} \bar{\boldsymbol{\mathrm{X}}}^{-1} \big)^{\vee}$, so that:
\begin{align}
    \boldsymbol{\mathrm{X}} \bar{\boldsymbol{\mathrm{X}}}_{1}^{-1}
	=
	\exp(\boldsymbol\epsilon^{\wedge}) \, \bar{\boldsymbol{\mathrm{X}}} \bar{\boldsymbol{\mathrm{X}}}_{1}^{-1}
	=
	\exp(\boldsymbol\epsilon^{\wedge}) \, \exp(\boldsymbol\xi_{1}^{\wedge}),\nonumber\\
    \boldsymbol{\mathrm{X}} \bar{\boldsymbol{\mathrm{X}}}_{2}^{-1}
	=
	\exp(\boldsymbol\epsilon^{\wedge}) \, \bar{\boldsymbol{\mathrm{X}}} \bar{\boldsymbol{\mathrm{X}}}_{2}^{-1}
	=
	\exp(\boldsymbol\epsilon^{\wedge}) \, \exp(\boldsymbol\xi_{2}^{\wedge}).\nonumber
\end{align}
Since $\boldsymbol\epsilon$ is small because of the concentrated Gaussian on a Lie group assumption, we can use the BCH approximation (Equation~\ref{eqn:bch_approx}) to write:
\begin{equation}
	\begin{gathered}
	\label{eqn:se3_norm_prod_factor_error_approx}
	\ln \big( \boldsymbol{\mathrm{X}} \bar{\boldsymbol{\mathrm{X}}}^{-1} \big)^{\vee}
	\approx
	\mathcal{J} \left( \boldsymbol\xi\right)^{-1} \boldsymbol\epsilon + \boldsymbol\xi
	=
	\mathcal{J} \left( \boldsymbol\xi \right)^{-1} \left( \boldsymbol\epsilon + \mathcal{J} \left( \boldsymbol\xi \right) \boldsymbol\xi \right).\nonumber
    \end{gathered}
\end{equation}
Substituting this expression in Equation~\ref{eqn:se3_normalised_prod} and rearranging terms, we get Equations~\ref{eqn:se3_normalised_prod_approx}-\ref{eqn:se3_norm_prod_approx_factor_params}.

Equation~\ref{eqn:se3_normalised_prod_approx} is recognizable as a normalized product of Gaussian PDFs, which is itself equal to a Gaussian PDF (\cite{bromiley2003products, petersen2008matrix}):
\begin{equation}
\label{eqn:se3_normalised_prod_approx_3}
	p_{*}\! \left( \boldsymbol\epsilon \right)
	\,\approx\,
	\eta_{*}\,
	\exp
	\left(
	-\textstyle\frac{1}{2}
	\left(
	\boldsymbol\epsilon - \boldsymbol\mu_{*}^{\prime} \right)^{\top}
	{\boldsymbol{\mathrm{\Sigma}}}_{*}^{\prime -1}
	\left( \boldsymbol\epsilon - \boldsymbol\mu_{*}^{\prime} \right)
	\right),
\end{equation}
with normalization factor $\eta_{*} = \alpha \eta_{1} \eta_{2} \gamma$, and the mean and covariance matrices are given by:
\begin{equation}
\label{eqn:se3_normalised_prod_params}
	\begin{gathered}
		{\boldsymbol{\mathrm{\mu}}}_{*}^{\prime}
		=
		{\boldsymbol{\mathrm{\Sigma}}}_{*}^{\prime}
		\left(
		{\boldsymbol{\mathrm{\Sigma}}}_{1}^{\prime -1} {\boldsymbol{\mathrm{\mu}}}_{1}^{\prime}
		+
		{\boldsymbol{\mathrm{\Sigma}}}_{2}^{\prime -1} {\boldsymbol{\mathrm{\mu}}}_{2}^{\prime}
		\right),\ 
  	{\boldsymbol{\mathrm{\Sigma}}}_{*}^{\prime}
		=
		\left(
		{\boldsymbol{\mathrm{\Sigma}}}_{1}^{\prime -1}
		+
		{\boldsymbol{\mathrm{\Sigma}}}_{2}^{\prime -1}
		\right)^{-1}\!\!.
	\end{gathered}
\end{equation}
This scaled product of Gaussian PDFs represents a \emph{non-zero-mean} Gaussian perturbation around a mean $\bar{\boldsymbol{\mathrm{X}}}_* \in SE(3)$. 


Finally, we obtain an approximation for the normalised product PDF as a \emph{zero-mean} Gaussian perturbation:
\begin{equation}
	\label{eqn:se3_normalised_prod_rev}
	p_{*} \left( \boldsymbol{\mathrm{X}} \right)
	\approx
	\eta_{*}
	\exp 
	\left(
	-\textstyle\frac{1}{2}
	\ln \big(\boldsymbol{\mathrm{X}} \bar{\boldsymbol{\mathrm{X}}}_{*}^{-1} \big)^{\vee\top}
	\boldsymbol{\mathrm{\Sigma}}_{*}^{-1}
	\ln \big(\boldsymbol{\mathrm{X}} \bar{\boldsymbol{\mathrm{X}}}_{*}^{-1} \big)^{\vee}
	\right)
\end{equation}
where $\boldsymbol\xi_{*} = 
	\ln \big(\bar{\boldsymbol{\mathrm{X}}} \bar{\boldsymbol{\mathrm{X}}}_{*}^{-1} \big)^{\vee}$ defines:
\begin{equation}
	\label{eqn:se3_norm_prod_rev_params}
	\begin{gathered}
		\bar{\boldsymbol{\mathrm{X}}}_{*}
		=
		\exp \left( \left( -\boldsymbol\xi_{*} \right)^{\wedge} \right) \bar{\boldsymbol{\mathrm{X}}},\ \ \ \ 
        \boldsymbol{\mathrm{\Sigma}}_{*}
		=
		\mathcal{J} \left( \boldsymbol\xi_{*} \right)^{-1}
		\boldsymbol{\mathrm{\Sigma}}_{*}^{\prime}
		\mathcal{J} \left( \boldsymbol\xi_{*} \right)^{-\top}\!\!.
	\end{gathered}
\end{equation}
Under the concentrated Gaussian on a Lie group assumption, the infinitesimal volume elements are related by \mbox{$d\boldsymbol{\mathrm{X}} \approx d \boldsymbol\epsilon$}, and so by an analogous BCH approximation to above:
\begin{equation}
	\label{eqn:se3_norm_prod_factor_error_approx_rev}
    \ln \big( \boldsymbol{\mathrm{X}} \bar{\boldsymbol{\mathrm{X}}}_{*}^{-1} \big)^{\!\vee}
	\approx
	\mathcal{J}\!\left( \boldsymbol\xi_{*} \right)^{\!-1}\!\boldsymbol\epsilon + \boldsymbol\xi_{*}\\
	=
	\mathcal{J}\!\left( \boldsymbol\xi_{*} \right)^{\!-1} \!\left( \boldsymbol\epsilon + \mathcal{J}\!\left( \boldsymbol\xi_{*} \right) \boldsymbol\xi_{*} \right)\nonumber
\end{equation}
Substituting this in Equation~\ref{eqn:se3_normalised_prod_rev} and rearranging terms, we can rewrite the mean and covariance in Equation~\ref{eqn:se3_normalised_prod_params} as:
\begin{equation}
\label{eqn:se3_norm_prod_params_rev}
	\begin{gathered}
		\boldsymbol\mu_{*}^{\prime} = -\mathcal{J}\left( \boldsymbol\xi_{*} \right) \boldsymbol\xi_{*},\ \ \ \ 
		\boldsymbol{\mathrm{\Sigma}}_{*}^{\prime}
		\, = \,
		\mathcal{J} \left( \boldsymbol\xi_{*} \right)
		\boldsymbol{\mathrm{\Sigma}}_{*}
		\mathcal{J} \left( \boldsymbol\xi_{*} \right)^{\top}.
    \end{gathered}
\end{equation}
These are established by reversing the steps used in the initial change of variables, by substituting analogous expressions to those of Equations~\ref{eqn:se3_norm_prod_approx_factor_params_0},\ref{eqn:se3_norm_prod_approx_factor_params}.

Now, let us consider the $SE(3)$ data fusion algorithm, which we construct in the form of a fixed-point iterative method because we cannot compute the deterministic mean $\bar{\boldsymbol{\mathrm{X}}}_{*}$ in closed form, since it would require solving a non-linear equation in $\boldsymbol\xi_{*}$ and $\boldsymbol\mu_{*}^{\prime}$ (Equation~\ref{eqn:se3_norm_prod_params_rev}). Therefore, we approximate the Jacobian in this equation as $\mathcal{J} \left( \boldsymbol\xi_{*} \right) \approx \boldsymbol{\mathrm{1}}$ (which becomes more accurate as $\boldsymbol\xi_{*}$ becomes smaller) and then substitute for $\boldsymbol\xi_{*}$ in Equation~\ref{eqn:se3_norm_prod_rev_params} to obtain an approximate solution:
\begin{equation}
\label{eqn:se3_norm_prod_rev_params_approx}
	\begin{gathered}
		\bar{\boldsymbol{\mathrm{X}}}_{*}
		\, \approx \,
		\exp \left( \boldsymbol\mu_{*}^{\prime \wedge} \right) \bar{\boldsymbol{\mathrm{X}}},\ \ \ \ 
		\boldsymbol{\mathrm{\Sigma}}_{*}
		\, \approx \,
		\boldsymbol{\mathrm{\Sigma}}_{*}^{\prime}.
	\end{gathered}
\end{equation}
We solve for $\boldsymbol\mu_{*}^{\prime}$ by updating the operating point $\bar{\boldsymbol{\mathrm{X}}}$ to equal our approximation $\bar{\boldsymbol{\mathrm{X}}}_{*}$, and then iterate until convergence:
\begin{equation}
	\label{eqn:se3_normalised_prod_update}
	\bar{\boldsymbol{\mathrm{X}}}
	\, \leftarrow \,
	\exp \left( \boldsymbol\mu_{*}^{\prime \wedge} \right) \bar{\boldsymbol{\mathrm{X}}}.
\end{equation}
At convergence, $\boldsymbol\mu_{*}^{\prime} \rightarrow \boldsymbol{\mathrm{0}}$ and we obtain the mean and covariance parameters of the normalised product of $SE(3)$ PDFs from Equation~\ref{eqn:se3_norm_prod_rev_params_approx} as $\bar{\boldsymbol{\mathrm{X}}}_{*} \, = \,
 \bar{\boldsymbol{\mathrm{X}}}$ and $\boldsymbol{\mathrm{\Sigma}}_{*} \, = \, \boldsymbol{\mathrm{\Sigma}}_{*}^{\prime}$, which we found to typically converge in 3-4 steps, consistent with the findings presented in \cite{barfoot2014associating}. 
Since we are operating in a real-time context where we want our code to run with predictable timing, we use 5 iterations. 

To simplify the implementation, we expanded the solution in Equation~\ref{eqn:se3_normalised_prod_params} in terms of the original factor PDF means and covariances, and in terms of the inverse Jacobian. We computed the inverse Jacobian up to second order in its series expansion (Equation~\ref{eqn:inv_jacobian_def}), which we found was sufficiently accurate (once again, this is consistent with findings presented in \cite{barfoot2014associating}). We also initialised the operating point $\bar{\boldsymbol{\mathrm{X}}}$ to equal the mean of the left-hand factor PDF of the product. Taken together, this results in Algorithm~\ref{alg:se3_data_fusion} of the methodology (Section~\ref{sec:discriminative_bayesian_filtering}).

While we recognise that our data fusion algorithm is essentially identical to the algorithm proposed in \cite{barfoot2014associating}, our method of deriving it is very different. In our case, we follow an algebraic approach, where the iteration arises in finding an approximate solution to a non-linear equation. This contrasts with the original derivation, which is based on iterative Gauss-Newton minimization of a Mahalanobis cost function. We believe that our derivation offers some new insights that were not present in the original derivation. For example, in \cite{barfoot2014associating}, the authors refer to their method qualitatively as a ``data fusion" method, but do not show that it finds a solution to the normalised product of $SE(3)$ PDFs. However, finding this product is essential for implementing the correction step of our $SE(3)$ discriminative Bayesian filter (Equation~\ref{eqn:bayes_filter_prob_fusion_approx}). 

\vspace{1em}
\section{Controller parameters}
\label{sec:controller_parameters}
\begin{table}[h!]
	\small\sf\centering
	\caption{Controller parameters (gain matrices, clipping ranges, feedback reference pose and feedforward velocity) used in tactile servoing controller for follower robot arm in object tracking experiments. The feedback reference pose is expressed as a 6-vector, where the first three elements denote the 3D position and the last three elements denote the 3D rotation in extrinsic-$xyz$ Euler format.}
	\label{tab:object_tracking_control_params}
	\begin{tabular}{cc}
	\toprule
	Parameter & Value\\
	\midrule
	$\boldsymbol{\mathrm{K}}_{p}$ & $\mathrm{diag}(5,5,5,2,2,0)$\\	
	$\boldsymbol{\mathrm{K}}_{i}$ & $\mathrm{diag}(0.5,0.5,0.5,0.2,0.2,0.2)$\\	
	$\boldsymbol{\mathrm{K}}_{d}$ & $\mathrm{diag}(0.5,0.5,0.5,0.2,0.2,0.2)$\\
	Integral error clipping & Not used\\
	Feedback ref pose & $(0,0,6,0,0,0)$\\
	Feedforward ref velocity & $(0,0,0,0,0,0)$\\
	\bottomrule
	\end{tabular}
 \vspace{3em}
	\small\sf\centering
	\caption{Controller parameters used in tactile servoing controller for surface following experiments. The feedback reference pose is expressed as a 6-vector, as in Table~\ref{tab:object_tracking_control_params}.
	\label{tab:surface_follow_control_params}}
	\begin{tabular}{cc}
	\toprule
	Parameter & Value\\
	\midrule
	$\boldsymbol{\mathrm{K}}_{p}$ & $\mathrm{diag}(0,0,2,2,2,0)$\\	
	$\boldsymbol{\mathrm{K}}_{i}$ & $\mathrm{diag}(0,0,0.1,0.1,0.1,0)$\\	
	$\boldsymbol{\mathrm{K}}_{d}$ & $\mathrm{diag}(0,0,0.05,0.05,0.05,0)$\\
	Integral error clipping & $\left[ -25, 25 \right]$ all components\\
	Feedback ref pose & $(0,0,3,0,0,0)$\\
	Feedforward ref velocity & Task-dependent\\
	\bottomrule
	\end{tabular}
\end{table}
\vfill

\begin{table}[t!]
	\vspace{-3em}
	\small\sf\centering
	\caption{Pushing controller parameters used for single-arm and dual-arm (leader) object pushing experiments. The feedback reference pose is expressed as a 6-vector, as in Table~\ref{tab:object_tracking_control_params}.
	\label{tab:pushing_control_params}}
	\begin{tabular}{cc}
	\toprule
	Parameter & Value\\
	\midrule
	PID 1 (MIMO)\\
	\midrule
	$\boldsymbol{\mathrm{K}}_{p}$ & $\mathrm{diag}(1,0,0,1,0,0)$\\	
	$\boldsymbol{\mathrm{K}}_{i}$ & $\mathrm{diag}(0.1,0,0,0.1,0,0)$\\	
	$\boldsymbol{\mathrm{K}}_{d}$ & $\mathrm{diag}(0.1,0,0,0.1,0,0)$\\
	Integral error clipping & $\left[ -25, 25 \right]$ all components\\
	Feedback ref pose & $(0,0,0,0,0,0)$\\
	Feedforward ref velocity & $(0,0,10,0,0,0)$\\
	\midrule
	PID 2 (SISO)\\
	\midrule
	$K_{p}$ & $0.9$\\	
	$K_{i}$ & $0.3$ (single-arm) / $0.5$ (dual-arm)\\	
	$K_{d}$ & $0.9$\\
	Integral error clipping & $\left[ -10, 10 \right]$\\
	Controller output clipping & $\left[ -15, 15 \right]$\\
	Ref bearing & $0$\\
	\bottomrule
	\end{tabular}
 \vspace{3em}
	\small\sf\centering
	\caption{Tracking (stabilising) controller parameters used for dual-arm (follower) object pushing experiments. The feedback reference pose is expressed as a 6-vector, as in Table~\ref{tab:object_tracking_control_params}.
	\label{tab:stabiliser_control_params}}
	\begin{tabular}{cc}
	\toprule
	Parameter & Value\\
	\midrule
	$\boldsymbol{\mathrm{K}}_{p}$ & $\mathrm{diag}(5,0,5,1,0,0)$\\	
	$\boldsymbol{\mathrm{K}}_{i}$ & $\mathrm{diag}(0.5,0,0.5,0.1,0,0)$\\	
	$\boldsymbol{\mathrm{K}}_{d}$ & $\mathrm{diag}(0.5,0,0.5,0.1,0,0)$\\
	Integral error clipping & $\left[ -200, 200 \right]$ all components\\
	Feedback ref pose & $(0,0,3,0,0,0)$\\
	Feedforward ref velocity & $(0,0,0,0,0,0)$\\
	\bottomrule
	\end{tabular}
\vspace{38.9em}
\end{table}

\vfill

\end{appendices}

\end{document}